\pdfoutput=1
\documentclass[lettersize,journal]{IEEEtran}
\usepackage[switch]{lineno}
\usepackage{amsmath,amsfonts}
\usepackage{algorithmic}
\usepackage{array}
\usepackage[caption=false,font=normalsize,labelfont=sf,textfont=sf]{subfig}
\usepackage{textcomp}
\usepackage{stfloats}
\usepackage{url}
\usepackage{verbatim}
\usepackage{graphicx}
\usepackage{balance}

\usepackage{dsbda-style}
\usepackage{hyperref}

\usepackage{booktabs}
\usepackage{multirow}
\usepackage{multicol}
\usepackage{xspace}
\usepackage{threeparttable}
\usepackage{longtable}

\newcommand{\mlp}{WideMLP\xspace}

\newcommand{\mytextsubscript}[1]{{\color{black}\textsubscript{#1}}}
\newcommand{\myheader}{}

\newcommand{\GenLMs}{generative language models\xspace}

\newcommand{\SLM}{SLM\xspace}
\newcommand{\SLMs}{SLMs\xspace}
\newcommand{\LLM}{LLM\xspace}
\newcommand{\LLMs}{LLMs\xspace}

\let\mycite\cite
\newcommand\myflag[0]{our experiment}
\let\mytablefontsize\small

\newcommand{\best}[1]{\textbf{#1}}
\newcommand{\secondbest}[1]{\underline{#1}}

\usepackage{cite}

\begin{document}

\title{Are We Really Making Much Progress in Text Classification? A Comparative Review}

\author{Ansgar Scherp$^\star$, 
Andor Diera, Fabian Karl, Bao Xin Lin$^\dagger$, Bhakti Khera,

Tim Meuser, Tushar Singhal, and Lukas Galke Poech$^\star$ 
\thanks{L. Galke Poech is with the University of Southern Denmark. A. Scherp, A. Diera, F. Karl, B. Lin, B. Khera, T. Meuser, and T. Singhal are with the University of Ulm, Germany.
$^\star$L. Galke Poech and A. Scherp contributed equally. $^\dagger$Deceased.}

}

\maketitle

\begin{abstract}
We survey the literature on single-label, multi-label, and hierarchical text classification and provide a quantitative comparison of methods categorized into bag-of-words, sequence-based, and graph- or hierarchy-based approaches.
Despite a recent surge in graph-based methods, they do not provide an improvement over fine-tuned transformer models on most evaluated datasets. 
Decoder-only generative language models show promise in few-shot in-context learning, but appear to lag behind fine-tuned language models when sufficient training data is available. 
The amount of training data needed for a fine-tuned language model to exceed the performance of a generative model is task-dependent.
We further highlight the variance in reported numbers across the literature when applying the same model to the same dataset, which can be traced to the use of different hyperparameter values, such as the fine-tuning learning rate.
For practitioners, we recommend using a fine-tuned language model when sufficient training data is available.
Otherwise, a frozen generative model, enhanced by few-shot in-context learning or reasoning, is preferable.

\noindent \textbf{Data availability}: 
The source code and further information are available at \url{https://github.com/ascherp/text-classification-survey}.

\end{abstract}


\section{Introduction}
\subsection{Motivation and Background}

Text classification is the task of assigning a categorical label or multiple of such labels to a given text unit~\cite{DBLP:journals/csur/Sebastiani02,DBLP:journals/tkde/MoreoES20}. 
It is a central task in natural language processing (NLP), with numerous practical applications, such as classifying scholarly documents, social media posts, news articles, or email spam.
Unsurprisingly, to this day, text classification remains a very active research field, with new methods appearing every week, as reflected by recent surveys~\cite{
DBLP:journals/tkde/HuLZHNL24,
fieldsSurveyTextClassification2024,
electronics13071199,
DBLP:journals/tist/LiPLXYSYH22,
DBLP:journals/csur/MinaeeKCNCG21,
DBLP:journals/corr/abs-2107-03158,
DBLP:journals/wias/ZhouGLVTBBK20, 
DBLP:journals/information/KowsariMHMBB19,
DBLP:journals/air/Kadhim19}. 

Besides the rapid pace of research in text classification, conceptually new approaches have also emerged that are not yet sufficiently covered in existing surveys. Those include the use of graph neural networks (GNNs) to process text, which attracts increasing attention from researchers, and increasingly prevalent large language models (LLMs), which are often assumed to be the state-of-the-art in all natural language processing tasks. 
Although there are surveys with good coverage of new methods, \eg focusing on LLMs~\cite{DBLP:journals/tkde/HuLZHNL24} or GNNs~\cite{gnns-for-nlp-survey}, those then lack a quantitative comparison. 
Other surveys perform a quantitative comparison but focus on single-label classification and use their own splits instead of the established splits per dataset~\cite{DBLP:journals/eswa/ReusensSTSVBB24}.
This survey covers both of these new families alongside classical approaches and provides a quantitative comparison across multi-class (or single-label), multi-label, and hierarchical text classification.

Many new methods for text classification are based on graph neural networks (GNNs)~\cite{book:hamilton:grl}.
Common to these GNN-based approaches is that they first generate a synthetic graph from the corpus that contains text-augmented vertices and edges with information on word and document co-occurrences, \eg~\cite{DBLP:conf/aaai/YaoM019,texting_acl2020}. 
Second, the GNN is trained on this graph to carry out the text classification task.
In hierarchical text classification, where classes are organized along a thematic hierarchical thesaurus~\cite{DBLP:journals/csur/Sebastiani02}, GNNs are often used to encode the label hierarchy, \eg~\cite{hbgl,DBLP:conf/acl/WangWH0W22}.

Another group of text classification methods is based on language models, which can be organized in encoder-only, encoder-decoder, and decoder-only models~\cite{yang2023harnessing}.
Encoder-only pre-trained language models such as BERT~\cite{DBLP:conf/naacl/DevlinCLT19} took big strides in many natural language processing tasks, including categorical text classification~\cite{galkescherp-acl2022}. 
The encoder-only transformer language models were followed by encoder-decoder variants like T5~\cite{T5} and decoder-only generative \LLMs such as the GPT models~\cite{DBLP:journals/corr/abs-2303-08774,DBLP:conf/nips/BrownMRSKDNSSAA20}.
Decoder-only transformer language models focus on text generation with remarkable in-context learning abilities.
This makes them strong zero-shot and few-shot models~\cite{carp,DBLP:conf/iclr/PatelLRCRC23,wei2022finetuned,DBLP:conf/nips/BrownMRSKDNSSAA20}. 
Yet, whether in-context learning LLMs are generally superior to fine-tuned small language models remains an open question~\cite{
DBLP:conf/emnlp/PecherSB25-100-samples,
DBLP:conf/acl/QoribMN24,
bucher2024finetunedsmallllmsstill,
lepagnol2024smalllanguagemodelsgood,
liu2024llmembedrethinkinglightweightllms,
edwards2024language,
li2023label}.

Decoder-only language models are usually much larger than masked language models~\cite{DBLP:conf/nips/BrownMRSKDNSSAA20}, which is because the causal language model objective architecture (left-to-right prediction) facilitates scalability through diagonal masking of the attention matrices -- enabling the model to get an error signal for each token in a batch.
Thus, we increasingly find the distinction between \SLMs (small language models) and \LLMs, \eg~\cite{DBLP:conf/aaai/Hu0CSLW024,DBLP:journals/corr/abs-2406-17534-retrieval-style,DBLP:journals/corr/abs-2402-16844}, which is based on an arbitrarily chosen parameter count.

We argue that a key distinction between \SLMs and \LLMs, particularly for classification tasks, lies in whether they are amenable to fine-tuning.
The encoder-only and encoder-decoder \SLMs (\eg BERT and T5) are based on masked language modeling (or masked span prediction for T5 respectively) and typically undergo subsequent task-specific fine-tuning. Decoder-only \LLMs (\eg GPTs) are pre-trained using a causal language modeling objective.
Sufficiently powerful decoder-only models are capable of tackling new tasks based on instructions in their context window at inference time~\cite{DBLP:conf/nips/BrownMRSKDNSSAA20}. That is, while still possible, fine-tuning is not strictly needed for decoder-only models to facilitate downstream tasks like text classification. 
Task-specific fine-tuning with \LLMs is typically more prohibitive there due to the higher number of parameters.
Thus, most current text classification approaches using \LLMs rely on few-shot in-context learning and dedicated prompting schemes~\cite{carp}.

\subsection{What is the Problem? Why is our Study Needed?}
\label{sec:questions}

Encoder-only language models like BERT have led to major improvements to the state of the art on many NLP tasks, including categorical text classification~\cite{galkescherp-acl2022}.
We ask whether the numerous newly proposed methods provide substantial improvements over encoder-only language models. 
We pinpoint this question to three aspects analyzed in this survey:

\begin{itemize}
    \item 
    \textit{Use of Graphs:}
The use of GNNs has been specialized for text classification in two ways: First, using synthetic graphs induced from the text corpus, \eg TextGCN~\cite{DBLP:conf/aaai/YaoM019}.
Those synthetic graphs may not provide additional information beyond what a neural network can directly extract from the text~\cite{galkescherp-acl2022}.
Second, graphs are provided as additional input for hierarchical classification and processed by explicit graph-encoders, \eg HiAGM~\cite{hiagm}.
Graph encoders may not provide added benefits to the task of hierarchical classification~\cite{hydra,radar}.

\item \textit{Large Language Models:}
The literature suggests that generative \LLMs do not generally improve over encoder-only \SLMs for text classification tasks~\cite{DBLP:conf/emnlp/PecherSB25-100-samples,
pushing-the-limit-11346014-rgpt,
carp,
yuan2023revisiting,
li2023chatgpt,
yu2023open}.
This comes despite the advantages of generative \LLMs of high capacity, prompt engineering techniques~\cite{carp,chain-of-thought}, 
instruction fine-tuning~\cite{wei2022finetuned}, and prompt tuning~\cite{liu-etal-2022-p,lester-etal-2021-power}.

\item \textit{Comparability of Methods:}
A general challenge when assessing the performance of methods is the comparability of the results.
Besides properly reporting the used datasets, splits, preprocessing, etc. an important question is if baselines are properly optimized~\cite{DBLP:conf/recsys/DacremaCJ19,leech2024questionablepracticesmachinelearning}. 
It is especially challenging to properly compare methods for text classification when so many new papers appear using GNNs and language models.
\end{itemize}

\subsection{Methodology}

We extensively review literature on modern and classical machine learning methods for single-label, multi-label, and hierarchical text classification.
Based on the literature search, we derive the families of methods for text classification. 

\begin{itemize}
\item Methods based on Bag of Words (BoW) using, \eg a support vector machine or a multi-layer perceptron~\cite{galkescherp-acl2022}.

\item Methods considering text as a sequence of tokens, \eg the encoder-only BERT~\cite{DBLP:conf/naacl/DevlinCLT19}, encoder-decoder T5~\cite{T5}, or decoder-only GPT~\cite{DBLP:conf/nips/BrownMRSKDNSSAA20}, 

\item Graph-based methods that employ graph neural networks on synthetic graphs like TextGCN~\cite{DBLP:conf/aaai/YaoM019} and hierarchy-based text classification methods like HGCLR~\cite{DBLP:conf/acl/WangWH0W22}.
\end{itemize}

We identify established single-label and multi-label benchmark datasets for our comparison and identify the top-performing methods.
We carefully assess the validity of the results for each method and each paper in the literature.
We check, among others, the train-test split used (whether they deviate from established benchmark splits), 
number of classes considered, 
hyperparameter values (are they provided and comparable), 
whether the baselines are optimized, 
metrics applied, 
and if there is any unusual preprocessing of the datasets that may have influenced the results.  

We aggregate results from the literature, complement them with experiments where relevant, and provide a systematic comparison of methods across different method families.

\subsection{Key Results}
Our comparative survey shows that fine-tuned transformer models are generally strong models for text classification. 
Despite their sheer number of parameters and training, \LLMs based on in-context learning generally do not outperform fine-tuned \SLMs in text classification.
Even when pushing the limit of using such \GenLMs by applying tricks such as advanced prompting techniques~\cite{carp} or using ensembles~\cite{pushing-the-limit-11346014-rgpt}, the performance does not, or only marginally on individual datasets, outperform those of encoder-only \SLMs. 
For example, it requires an ensemble of fine-tuned Llama models to reach or slightly outperform an encoder-only \SLM on two benchmark datasets~\cite{pushing-the-limit-11346014-rgpt}.

We attribute these observations to two main factors.
First, task-specific fine-tuning is still an important distinctive criterion.
The encoder-only models like BERT are pre-trained with the masked language modeling objective and fine-tuned with a classifier head for specific datasets on the text classification task.
Second, the left-to-right attention mask seems to be suboptimal for text classification tasks.
This again relates well to the distinction between discriminative and generative approaches~\cite{DBLP:conf/sigir/Nallapati04} -- the former 
focusing on the decision boundary directly, \ie $p(y|x)$, which we interpret here as task-specific fine-tuning, 
and the latter 
aiming to learn the full joint distribution $p(x,y)$, which we interpret here as in-context learning.

For the GNN-based methods, despite the huge number of methods developed in recent years, the idea of exploiting a synthetically induced graph to improve text classification has not lived up to its promise.
Many graph-based methods such as \cite{DBLP:conf/aaai/YaoM019,DBLP:conf/aaai/LiuYZWL20,DBLP:conf/wsdm/RageshSIBL21} are not only outperformed by simply applying BERT but already fall below simple baselines such as a logistic regression or a simple classifier such as a multi-layer perception on a bag-of-words representation~\cite{galkescherp-acl2022,DBLP:conf/wsdm/RageshSIBL21}.
Such baseline models already extract relevant information for text classification from the raw text~\cite{DBLP:journals/csur/Sebastiani02}.
Adding a synthetically generated graph from the corpus generally does not provide additional information that a neural network can exploit~\cite{galkescherp-acl2022}.

A worrying observation is the lack of strong baselines and/or not properly tuning them appropriately.
Like Dacrema, Cremonesi, and Jannach~\cite{DBLP:conf/recsys/DacremaCJ19} observed for neural recommender systems that ``works can be outperformed at least on some datasets by conceptually and computationally simpler algorithms'', the reason is also that baselines are not properly optimized~\cite{DBLP:conf/recsys/ShehzadJ23} (``everyone's a winner'').
Classical machine learning methods such as support vector machines, logistic regression, and multi-layer perceptrons are largely ignored in papers published in the last years, despite showing consistently strong results across various datasets~\cite{galke2023really,galkescherp-acl2022,DBLP:conf/wsdm/RageshSIBL21}.

There are cases where even the baseline's most basic hyperparameter, the learning rate (or fine-tuning learning rate, in the case of BERT models), is not properly considered.
We show this in the example of BERT and its variants. 
Despite running the same BERT model on the same datasets and splits, the numbers from the literature show a discrepancy in BERT's performance up to a $13$ points in accuracy.
Generally, such discrepancies may be attributed to the learning rate, preprocessing, batch size, implementation details, random seeds, and tuning efforts. 
We analyze the effect of the learning rate in more detail.

Following studies in other areas of machine learning, we conclude that strong baselines must be used and their hyperparameters properly selected as it is an important means to argue about \textit{true} scientific advancement~\cite{DBLP:conf/acl/HenaoLCSSWWMZ18,DBLP:conf/recsys/DacremaCJ19}, and also their usefulness in practical settings.

\subsection{Remainder}

Below, we introduce our survey methodology.
The experimental apparatus, including the considered datasets, and the procedure for running our experiments are described in \Secref{sec:apparatus}.
Sections \ref{sec:bow}, \ref{sec:sequence}, and \ref{sec:graph} provide an overview of the literature in bag-of-words approaches, sequence-based approaches, and graph-based approaches, respectively.
The results of our quantitative comparison are presented in \Secref{sec:results}.
Subsequently, we discuss 
the findings in \Secref{sec:findings},
limitations in \Secref{sec:limitations},  and provide future directions in \Secref{sec:future}, before we conclude.

\section{Methodology}
\label{sec:methodology}

This survey covers single-label text classification, multi-label text classification, and hierarchical text classification -- covering published methods up to May 2026.
We do not consider \emph{extreme} multi-label classification, where the focus is on dealing with a very large label space, which requires a different set of specialized techniques~\cite{xml0,xml4,xml3,xml2} and different metrics (ranking metrics rather than classification metrics).
We leave this comparison for future work. 

In a first step, we have collected and analyzed recent surveys on single-label, multi-label, and hierarchical text classification and searched for research papers that include comparison studies~\cite{
wang2023graph,
duarte2023review,
buguenoConnectingDotsWhat2023,
DBLP:journals/tist/LiPLXYSYH22,
DBLP:journals/csur/MinaeeKCNCG21,
DBLP:journals/pr/TarekegnGM21,
DBLP:journals/corr/abs-2107-03158,
raihan2021-survey,
DBLP:journals/wias/ZhouGLVTBBK20,
DBLP:conf/esann/QaraeiKB20,
DBLP:journals/corr/abs-2011-11197,
DBLP:journals/information/KowsariMHMBB19,
DBLP:journals/air/Kadhim19,
galkescherp-acl2022,
DBLP:conf/jcdl/MaiGS18,
DBLP:conf/kcap/GalkeMSBS17,
DBLP:conf/sigir/ZhangWYWZ16}.
These cover the range from classical machine learning models to deep classification models.
Second, we have screened for literature in key NLP and artificial intelligence venues.

Based on this input, we have determined the families of methods and benchmark datasets used (see Tables~\ref{tab:datasets} and \ref{tab:multilabeldatasets}).
The categorization of the methods is mainly based on the method signature, distinguishing whether methods operate on BoW, sequence, or graph-structured input~\cite{galkescherp-acl2022}. 
We extend the categorization by further distinguishing 
sequence-based methods into classical recurrent and convolution models, encoder-only (language) models, encoder-decoder models, and decoder-only models.
Furthermore, we distinguish
between two types of graph-based methods: 
First, approaches that employ a graph structure derived from a corpus of text documents, which we call synthetic text-graph approaches. 
Second, hierarchy-based methods that use a graph model to encode the hierarchical structure among classes. 

We focus our analysis on methods that show strong performance and include them in our study.
Comparing results from different studies comes with limitations regarding comparability of results.
To ensure comparability to the extent possible, we have verified that the same train-test split of the datasets is used for all methods and that the same metrics are used.
We have also checked whether the datasets have been modified (\eg use only a subset of the classes or have removed samples) to avoid bias and mistakenly give advantages.

We do not consider papers that do not allow for a fair comparison with the state of the art.
Reasons include that they 
\begin{itemize}
\item used different benchmark datasets only like \cite{
sharma2025readreinforcementbasedadversariallearning,
petridis2024textclassificationneuralnetworks,
peng2025text,
DBLP:journals/nca/JiaJDZLXC23-mhgat,
10748883-boing-aviation-paper,
LMTCSG-10705790,
AGBESI2024e38515,
10.1007/978-3-031-72350-6_20,
DBLP:journals/corr/abs-2408-15650,
Wei_Sun_2024,
DBLP:conf/kdd/Yu00S24,
DBLP:conf/sigir/DaiYCX24, 
JAMSHIDI2024102306,
DBLP:journals/access/ThaminkaewLV24,
DBLP:journals/tkde/HuLZHNL24,
DBLP:conf/coling/LyKCK24,
DBLP:journals/corr/abs-2401-01667-MLP-Compass,
thaminkaew2024ieeeaccess,
li2023chatgpt,yu2023open,yuan2023revisiting,
zhu2023,
chae_davidson_2023,
zhang2023pretraining,
Yang-ClimateChangeClassifiert-2023,
DBLP:journals/apin/TanRW23,
bgnn-xml,
DBLP:conf/ijcnn/TranSZPB22,
LiuEtAl2022-LongText,
DBLP:conf/aiia/BreazzanoC021a,
DBLP:journals/corr/abs-2206-07253,
DBLP:journals/corr/abs-2112-11389,
chalkidis-etal-2020-empirical,
DBLP:conf/eacl/SchwenkBCL17,gururangan-etal-2019-variational},

\item modified the datasets to use a different number of classes as done in \cite{DBLP:conf/coling/ZhouQZXBX16,DBLP:conf/aaai/LaiXLZ15,MTAPparlakNovelFeatureClassbased2023},

\item employed different train-test splits like
\cite{
donabauer2024tokenlevelgraphsshorttext,
DBLP:journals/eswa/ReusensSTSVBB24,
edwards2024language,
StylianouEtAl2023,
DBLP:journals/corr/abs-2211-02563,
DBLP:conf/ijcnn/SunHCYSM22,
raihan2021-survey,
DBLP:conf/icaart/PalSS20,
moreo2020,
canutoThoroughEvaluationDistanceBased2018}, 

\item conduct their own sampling of the train-test splits instead of using standardized splits~\cite{PAAM-HiA-T5}, 

\item perform data augmentation from external knowledge bases~\cite{zhu2023},

\item train-test splits are not reported~\cite{moreo2020,sf-cnn-iasc.2023.027429,YuanEtAl-MSVM-kNN-2008},

\item used fewer training examples~\cite{
DBLP:conf/acl/LiangZ0Z24,
DBLP:conf/aaai/LiuZ0ZWMCYZ24,
Xie2024DataLT,
zhu2023,
duarte2023review,
DBLP:conf/emnlp/ZhengWYD22,
DBLP:conf/ijcnn/SunHCYSM22,DBLP:conf/emnlp/WangWYD21}, 
or

\item used different evaluation measures~\cite{tan2022,DBLP:conf/emnlp/HuangGKOO21}.

\end{itemize}

The rationales for why certain changes are made were not always clear in the literature.
However, it reflects a general problem of comparability in machine learning research~\cite{leech2024questionablepracticesmachinelearning}.

\section{BoW-based Methods}\label{sec:bow}
Under pure BoW-based (Bag of Words) text classification, we denote approaches that operate only on a multiset of words (or tokens) from the input document. 
Given paired training examples $(\vx,y) \in \train$, each consisting of a vector that holds the frequency of words in its components $\vx \in \sR^{n_\mathrm{vocab}}$, which is commonly referred to as a bag (multiset) of words, and a class label $y \in \sY$. 
The goal is to learn a generalizable function $\hat\vy = f_\theta^{(\mathrm{BoW})}(\vx)$ with parameters $\theta$ such that $\operatorname{arg\,max}(\hat\vy)$ is the true label $y$ for input $\vx$.
For multi-label classification, the BoW-based model considers multiple class labels. 
Instead of using $\operatorname{arg\,max}(\hat\vy)$ to decide on a label, a binary sigmoid output is commonly used per label, along with a threshold $\lambda$ that determines whether the corresponding class will be assigned. 
The multi-label model is trained with a binary cross-entropy loss instead of a categorical cross-entropy.
As output, the multi-label classifier can produce between $0$ and $|\sY|$ many labels~\cite{DBLP:journals/csur/Sebastiani02}, \ie it is possible that no label is predicted.

\subsection{Classical BoW Methods}

Classical machine learning methods that operate on a BoW-based input are extensively discussed in surveys and comparison studies~\cite{
DBLP:conf/wsdm/RageshSIBL21,
DBLP:journals/information/KowsariMHMBB19,
DBLP:journals/air/Kadhim19,
DBLP:conf/kcap/GalkeMSBS17}.
These studies show that the best-performing classical models are Support Vector Machines (SVM) and logistic regression (LR).
Especially, the strong performance of logistic regression is astonishing.
For instance, Ragesh~\etal\cite{DBLP:conf/wsdm/RageshSIBL21} have shown that logistic regression outperforms the advanced graph-based TextGCN~\cite{DBLP:conf/aaai/YaoM019} method. 
The CFE-IterativeAdditive method uses a linear SVM with term-based substring matching (from the documents) to the class names~\cite{ATTIEH2023110215}.
It uses this label matching of the terms to adapt the global IDF weights iteratively, denoted as TF-ICF.

\subsection{Deep BoW Methods}

With more advanced BoW methods, Galke~\etal\cite{galkescherp-acl2022} have found that an MLP on a bag-of-words representation of the text outperforms many graph-based approaches.
Earlier approaches are mainly based on pre-trained word embeddings~\cite{DBLP:conf/nips/MikolovSCCD13,DBLP:conf/emnlp/PenningtonSM14}.
For instance, Iyyer~\etal\cite{DBLP:conf/acl/IyyerMBD15} proposed Deep Averaging Networks (DAN), a combination of word embeddings and deep feedforward networks.
DAN is an MLP with one to six hidden layers, non-linear activation, dropout, and AdaGrad as an optimization method.
The results suggest that pre-trained embeddings such as GloVe~\cite{DBLP:conf/emnlp/PenningtonSM14} would be preferable over randomly initialized neural bag-of-words~\cite{DBLP:conf/acl/KalchbrennerGB14}.
In fastText~\cite{DBLP:journals/tacl/BojanowskiGJM17,DBLP:conf/eacl/GraveMJB17}, a linear layer is used on top of pre-trained embeddings for classification.
Furthermore, Shen~\etal\cite{DBLP:conf/acl/HenaoLCSSWWMZ18} explore different pooling variants for the input word embeddings and find that their Simple Word Embedding Models (SWEM) can rival approaches based on recurrent (RNN) and convolutional neural networks (CNN).
Note that approaches such as fastText and SWEM that apply a logistic regression on top of pre-trained word embeddings share a similar architecture as an MLP with one hidden layer. 
However, the standard training protocol involves pre-training the word embedding on large amounts of unlabeled text and then freezing the word embeddings while training the logistic regression~\cite{DBLP:conf/nips/MikolovSCCD13}.

\section{Sequence-based Methods}\label{sec:sequence}
As sequence-based methods, we consider recurrent and convolutional neural networks, \SLMs (\eg BERT and T5) and \LLMs (\eg GPT-3.5). 
These approaches aim for \emph{contextualized} word representations, for which nearby words and word order are taken into account.
The model signature for sequence-based methods is
$\hat{y} = f_\theta^{(\mathrm{sequence})} ( \langle x_1, x_2, \ldots, x_k \rangle )$,
where $k$ is the (maximum) sequence length.

\subsection{Recurrent and Convolutional Neural Networks}

Before the era of pre-trained language models, recurrent neural networks (RNNs) were a natural choice for any NLP task.
In particular, LSTMs~\cite{lstm} enhanced with an attention mechanism~\cite{bahdanau2014neural} and bidirectional processing~\cite{liu2016recurrent}, were considered the state of the art for NLP tasks before transformers.
For instance, SGM~\cite{sgm} is a model for multi-label classification based on a bidirectional LSTM~\cite{lstm} with an attention mechanism, which has later been extended to use T5 as a backbone language model~\cite{seq2tree}. 
Word embeddings, such as GloVe~\cite{DBLP:conf/emnlp/PenningtonSM14} have been used with CNNs~\cite{DBLP:conf/coling/HuangCC22} as well as Bi-LSTM~\cite{zhao2021sequential}.
Lyu and Liu~\cite{DBLP:journals/corr/abs-2006-15795} combine a 2D-CNN with a bidirectional RNN.

\subsection{Encoder-only Small Language Models}
The transformer~\cite{DBLP:conf/nips/VaswaniSPUJGKP17}, originally developed for machine translation, introduced a key-value self-attention mechanism and had an immense impact on language modeling.
Transformer-based language models are pre-trained with a self-supervised training objective on large text corpora and can be subsequently fine-tuned with a supervised objective on a specific task.
Generally, transformer-based language models can be categorized into encoder-only models, encoder-decoder models, and decoder-only models~\cite{yang2023harnessing}.
BERT and its follow-up approaches are encoder-only language models and are trained with the masked-language modeling objective~\cite{DBLP:conf/naacl/DevlinCLT19}.
Their main purpose is encoding the text and performing a classification or regression task with a dedicated module (\eg a linear output layer) on top of the encoder.
The output layer is trained from scratch (\ie from random initialization) during fine-tuning.

Popular follow-up works of BERT are
RoBERTa~\cite{liu_roberta:_2019}, DistilBERT~\cite{distilbert}, ALBERT~\cite{lan_albert_2020}, DeBERTa~\cite{he_deberta_2021}, and
ERNIE~2.0~\cite{sun_ernie_2019} \emph{inter alia}.
RoBERTa improves the pre-training procedure of BERT and removes the next sentence prediction objective. DeBERTa improves upon BERT through a disentangled attention mechanism and corresponding masking techniques.
DeBERTaV3~\cite{DBLP:conf/iclr/HeGC23} is a modified DeBERTa with an ELECTRA-style pre-training~\cite{DBLP:conf/iclr/ClarkLLM20-electra}.
DistilBERT~\cite{distilbert} is a distilled version of BERT with 40\% reduced parameters and 60\% faster inference time that retains 97\% of BERT's performance on the GLUE benchmark~\cite{DBLP:conf/iclr/WangSMHLB19}.
ALBERT reduces the memory footprint and increases the training speed of BERT for improved scalability.
DeBERTa introduces a disentangled attention mechanism, \ie keeping word position and content embeddings separate, and an enhanced mask decoder, which introduces absolute position encodings to the final softmax.
ERNIE 2.0 employs a continual multi-task learning strategy. 
Whenever a new task is introduced, the previous model parameters are used for initialization, and the new task is added to the multi-task learning objective.
The AM-BERT and AM-RoBERTa models augment transformer fine-tuning for text classification with two attention matrices based on keyphrases and on the input's dependency graph~\cite{am-bert}.
Finally, ModernBERT is a refurbished variant of the BERT model integrating various optimizations developed over the years and providing an input length of $8,192$ tokens~\cite{warner2024smarterbetterfasterlonger}.

Pre-trained encoder-only language models have also found their way into hierarchical text classification.
For instance, HBGL~\cite{hbgl} uses BERT to represent the text and represent the hierarchically organized classes.
For the classes, HBGL first learns the label embeddings from the global hierarchy, \ie the taxonomy.
Specifically, the label embeddings are learned by employing a masked language modeling objective on sequences that correspond to paths in the label hierarchy.
Subsequently, it learns to predict the document labels one by one in the order of the local hierarchy, \ie level-wise from the root to the most specific label in the taxonomy.
By this, HBGL exploits the hierarchy of the labels as defined in the taxonomy and treats the label generation as a multi-label text classification task in a similar way as Nam et al.~ \cite{namMaximizingSubsetAccuracy2017} did with sequence-to-sequence models trained from scratch.
Thus, HBGL is essentially a \textit{hierarchy-aware} sequence-based transformer model. 
However, as described above, the key ingredient of HBGL is making use of the sequence-based model BERT.
HBGL \textit{does not use an external graph encoder} for representing the taxonomy.
In contrast, the graph-based methods (described in Section~\ref{sec:graph}) always use an \textit{explicit} graph encoder for representing the taxonomy, most commonly a graph neural network, independent of what other model is being used (\eg a CNN, BERT, or other).
Since HBGL's core is heavily based on BERT and does not have an explicit graph encoder, we consider it a sequence-based transformer method.

\subsection{Encoder-decoder Small Language Models}
\label{sec:finetune-encdec-slms}

Examples of encoder-decoder language models include BART~\cite{lewis2020bart} and T5~\cite{T5}, which combine ideas from masked language models and generative transformers. 
The idea is to cast downstream tasks, including classification, into a text-to-text framework with task-specific prompt prefixes. 
Although these models generate text, they can also be used for classification tasks by interpreting generated tokens as class labels (generative variant, as in \cite{radar}) or through fine-tuning a dedicated new classifier head (discriminative variant).
In multi-label classification, text-to-text (= sequence-to-sequence) models have been used even before the era of large language models~\cite{namMaximizingSubsetAccuracy2017} to facilitate the prediction of multiple labels.

The encoder-decoder model Seq2Tree adopts T5 to perform hierarchical text classification~\cite{seq2tree}.
It uses a depth-first search approach to linearize the label hierarchy in order to encode it as a sequence in the model.
RADAr is a sequence-to-sequence model that uses RoBERTa as a text encoder and a custom two-layer Transformer-based decoder for hierarchical text classification~\cite{radar}.
In contrast to hierarchical text classification methods, see Section~\ref{sec:rw:hierarcy-based-methods}, RADAr does not operate on a given dataset taxonomy.

\subsection{Large Language Models}

Decoder-only language models such as GPT-3~\cite{DBLP:conf/nips/BrownMRSKDNSSAA20} and Llama3~\cite{llama3}
are trained with a left-to-right, \ie causal language modeling, objective. This makes decoder-only models most suitable for text generation. 
However, decoder-only models are very flexible and can also be used to carry out other downstream tasks, including text classification~\cite{DBLP:conf/nips/BrownMRSKDNSSAA20,radfordLanguageModelsAre}. 
This is done by specifying the task in natural language and providing a few examples in the prompt~\cite{DBLP:conf/nips/BrownMRSKDNSSAA20}. 
This practice, known as in-context learning, does not require updating the model. 
In-context learning has led to increased interest in the design of and working with prompts, such as Chain-of-Thought (CoT) prompting~\cite{chain-of-thought} \emph{inter alia}.

Sun~\etal~\cite{carp} evaluated GPT-3.5+CoT on single-label text classification. They also introduced the prompting strategy Clue and Reasoning Prompting (CARP), where the instructions consist of first finding relevant clues in the text input and then providing a classification result based on the clues along with an explanation. 
The authors evaluate GPT-3.5, GPT-3.5+CoT, and GPT-3.5+CARP with two different samplers for selecting the in-context examples.
The samplers are a uniform sampler and a RoBERTa model fine-tuned on a generic natural language inference task.
Using RoBERTa representations of the training documents, the sampler employs a $k$NN search on the examples to sample more representative documents per class to be included in the prompt.
The best-performing CARP variant employs a 16-shot RoBERTa sampler with a majority vote over multiple runs of prompting GPT-3.5, which we denote as GPT-3.5+CARP+vote.

Prompt boosting has shown promising results for single-label classification
\cite{pmlr-v202-hou23b}, where differently prompted LLMs form an ensemble with an adaptive boosting algorithm. 
However, the few-shot performance of large language models without any fine-tuning is still lower than that of fine-tuned small language models~\cite{edwards2024language}.
Beyond prompting techniques,
Zhang et al.~\cite{pushing-the-limit-11346014-rgpt} have experimented with fine-tuning an ensemble of Llama-2~\cite{touvronLlamaOpenFoundation2023} models for text classification, leading to competitive scores.
Li et al.~\cite{li2023label} fine-tune a Llama-2 model~\cite{touvronLlamaOpenFoundation2023} without causal masking such that both left and right context can be considered during self-attention. 

In the end, it is currently unknown whether in-context learning with a large language model is sufficient for a given task or whether fine-tuning is needed. 

\section{Graph-based Methods}\label{sec:graph}

Graphs can serve several purposes in text classification: One is to consider the input data as a graph (\ie the documents, their words), which we call synthetic text-graph approaches to distinguish them from graphs in which the structure has a natural interpretation (\eg citations graphs). Another purpose is considering an additional label hierarchy as input to the model~\cite{DBLP:journals/csur/Sebastiani02} (\ie classes organized in a hierarchy), which we call hierarchy-based methods.

\subsection{Synthetic Text-Graph Methods}

Using graphs induced from text has a long history in text classification. 
An early work is the term co-occurrence graph of the KeyGraph algorithm~\cite{DBLP:conf/adl/OhsawaBY98}.
The graph is split into segments, representing the key concepts in the document.
Co-occurrence graphs have also been used for automatic keyword extraction~\cite{Rose2010} and classification~\cite{DBLP:conf/emnlp/ZhangDXLZ21}.
Modern methods exploit this idea of a graph induced from the text.
The text corpus is first transformed into a graph, which is then fed as input into a graph neural network (GNN)~\cite{book:hamilton:grl}.

The synthetic text-graph approaches to text classification first set up a \emph{synthetic} graph based on the text corpus $\train$ such that an adjacency matrix is created from a document corpus $\hat\mA := \operatorname{make-graph}(\train)$.
The graph is composed of word nodes and document nodes, each receiving its own embedding (by setting $\mX =\mI$).
For example, in TextGCN, the graph is created from word-word edges (modeled by pointwise mutual information) and word-document edges (based on TF-IDF scores). 
Then, a parameterized function $f_\theta^{(\mathrm{graph})}(\mX, \hat\mA)$ is learned that uses the graph as input, where $\mX$ are the node features. 
In the case of TextGCN, the function is a standard graph convolutional network~\cite{DBLP:conf/iclr/KipfW17}. 
Note that graph-based approaches such as TextGCN disregard word order, similar to the BoW-based models described above.

Among others, methods that follow this synthetic text-graph approach include 
TextGCN~\cite{DBLP:conf/aaai/YaoM019},
TensorGCN~\cite{DBLP:conf/aaai/LiuYZWL20},
Hete\-GCN~\cite{DBLP:conf/wsdm/RageshSIBL21},
HyperGAT~\cite{DBLP:conf/emnlp/DingWLLL20}, 
HGAT~\cite{DBLP:journals/tois/YangHSJLN21},
DADGNN~\cite{DBLP:conf/emnlp/LiuGG0F21}, 
STGCN~\cite{stgcn},
SHINE~\cite{DBLP:conf/emnlp/WangWYD21}, and
AGGNN~\cite{aggnn}.
While many of these methods are strictly transductive such as HeteGCN and TensorGCN, other methods are 
inductive~\cite{DBLP:conf/coling/HuangCC22,induct-gcn,texting_acl2020,DBLP:conf/emnlp/HuangMLZW19} or can be adapted to become inductive~\cite{DBLP:conf/wsdm/RageshSIBL21}.
Transductive models need access to the unlabeled test documents at training time. 
This requires computing the graph also on the documents of the test set and making this information available during training (but without the labels from the test set).
In contrast, inductive models can be applied to new data. 
Here, the graph induced from the text is computed only on the training set.

Transductive training has inherent drawbacks as the models cannot be applied to new documents.
For example, in TextGCN's original transductive formulation, the entire graph, including the unlabeled test set, must be available for training.
This may be prohibitive in practical applications as each batch of new documents would require retraining the model.
When TextGCN and other graph-based methods are adapted for inductive learning, where the test set is unseen, they achieve notably lower scores~\cite{DBLP:conf/wsdm/RageshSIBL21}. 
Note that all previously described bag-of-words and sequence-based models fall in the inductive category and can be applied to new documents.

We briefly discuss further selected graph-based methods.
HeteGCN combined ideas from Predictive Text Embedding~\cite{DBLP:conf/kdd/TangQM15} and TextGCN and splits the adjacency matrix into its word-document and word-word sub-matrices and fuses the different layers' representations when needed.
TensorGCN explores and combines multiple different ways of converting the text into a graph, such as a semantic graph created with an LSTM, a syntactic graph created by dependency parsing, and a sequential graph based on word co-occurrence. 
HyperGAT combines graph attention~\cite{velickovic2018graph} with the concept of hyperedges based on sequential structure and topic models~\cite{DBLP:journals/jmlr/BleiNJ03}.
AGGNN~\cite{aggnn} focuses on text pooling mechanism along with gated graph sequence neural networks~\cite{DBLP:journals/corr/LiTBZ15}.
DADGNN is a graph-based approach that uses attention diffusion and decoupling techniques for tackling the over-smoothing problem of the GNN and building deeper models. 
Lastly, STGCN tackles short text classification by building upon ideas from TextGCN and adding word-topic and document-topic edges from a topic model, similar to HyperGAT.
The authors also experimented with combining STGCN with a BiLSTM and a BERT model.
In their experiments, the combination STGCN+BERT+BiLSTM gave the best results, while pure STGCN fell behind pure BERT. MHGAT~\cite{mhgat} follows a different approach and captures word order by adding position-specific hyperedges to the graph to be processed by a graph attention network. These position edges are obtained through a sine/cosine transformation, following a similar strategy as in the Vaswani transformer's positional encoding~\cite{DBLP:conf/nips/VaswaniSPUJGKP17}.

Particularly between 2022 and 2024, numerous new graph-based models have been published.
Many of them use BERT in conjunction \textit{with an explicit graph encoder}, usually a graph neural network, such as in BertGCN~\cite{DBLP:conf/acl/LinMSHKLW21}, CTGCN~\cite{ctgcn}, ILGCN~\cite{ilgcn}, TSW-GNN~\cite{tsw-gnn}, and ConTextING~\cite{DBLP:conf/coling/HuangCC22}.
They differ from the other GNN-based methods as the graph is not computed based on word co-occurrences but BERT's subword tokens.
Further graph-based methods are KGAT~\cite{DBLP:conf/nlpcc/WangWYZSJWZ22}, InducT-GCN~\cite{induct-gcn}, TextSSL~\cite{textssl2022}, GLTC~\cite{gltc2023}, and others.
A recent survey on GNNs for text classification was performed by Wang, Ding, and Han~\cite{wang2023graph}.
Moreover, Bugue\~{n}o and de Melo~\cite{buguenoConnectingDotsWhat2023} compare different initial document representations including Word2vec~\cite{DBLP:conf/nips/MikolovSCCD13}, GloVe~\cite{DBLP:conf/emnlp/PenningtonSM14}, and frozen BERT embeddings and use them in conjunction with graph neural networks. 
They employ frozen and fine-tuned BERT models as baselines for the categorical text classification task. 
Their results show that---on most datasets---graph neural networks hardly compete with a fine-tuned BERT.

\subsection{Hierarchy-based Methods}
\label{sec:rw:hierarcy-based-methods}
In addition to using text-induced graphs as described above, the dataset's classes may be organized into a graph structure.
This is the case in hierarchical text classification, where each document should be annotated with a set of labels rather than a single class label~\cite{DBLP:journals/csur/Sebastiani02} and the classes are organized along a taxonomy. 
The taxonomic hierarchy of labels is typically modeled as a tree or a directed acyclic graph~\cite{shen-etal-2021-taxoclass,hiagm,pengHierarchicalTaxonomyAwareAttentional2021}. 
The goal is to predict multiple class labels that correspond to one or more nodes in the hierarchy.

Following the taxonomic hierarchy to the root, the classes become more general (broader), while going towards the leaves, the classes become more specific (narrower).
The documents are typically annotated with some specific classes in the taxonomy.
In hierarchical text classification, this taxonomy is often used to \emph{enrich} the gold standard~\cite{hiagm}. 
This means that all vertices along the entire path from the root to the assigned classes are added as ground truth.
The hierarchy-based text classifier HiAGM~\cite{hiagm} and its follow-up works, such as HGCLR~\cite{DBLP:conf/acl/WangWH0W22} and HBGL~\cite{hbgl}, rely on this enrichment.

The enrichment affects both the training and evaluation of hierarchy-aware methods.
After this enrichment, the dataset consists of a set of documents $\mX$. Each document is annotated with multiple labels, typically modeled as a label indicator matrix $\hat\mY$, and a hierarchy of classes $\mH$.
Then, the goal is to learn a function $\vx, \mH \mapsto \hat\vy$ that maps the current document $\vx$ to a set of enriched labels $\hat\vy$, while also taking into account the label hierarchy $\mH$. 
The hierarchy-based methods make use of an explicit graph encoder to represent the taxonomy and to exploit it in the model architecture to classify the text.

We briefly discuss the selected methods.
HiAGM is a hierarchical text classifier that models the hierarchy as a directed graph along with hierarchy-aware structure encoders~\cite{hiagm}.  
It comes in two variants: HiAGM-LA, which is a multi-label attention model that uses an inductive approach.  
HiAGM-TP is a text feature propagation model that uses a deductive approach to extract hierarchy-aware text features. It uses a GNN-based encoder to obtain a representation for each class and compare it with a Tree-LSTM representation of the text.  
In early 2021, several other hierarchical label-based attention models were published.  For example, HLAN \cite{dong2021explainable}, LA-HCN \cite{DBLP:journals/corr/abs-2009-10938}, RLHR~\cite{liu2021improving}, and the
weakly-supervised Taxo\-Class~\cite{shen-etal-2021-taxoclass}.
Other hierarchy-based methods are
\cite{DBLP:journals/corr/abs-1812-11270,DBLP:conf/www/XiaoLS19,DBLP:journals/corr/abs-1909-00161}.
Further methods that use BERT for text encoding and a graph neural
network for the hierarchy are BERT+HiMATCH~\cite{DBLP:conf/acl/ChenMLY20}, as well as HGCLR.
The latter uses contrastive learning to align text and graph representations.
K-HTC combines BERT as a text encoder with a knowledge graph based on entities relevant to the classification task~\cite{DBLP:conf/acl/LiuZHWZ0C23-k-htc}. 

The hierarchy-aware and label-balanced (HALB) model extends HGCLR by replacing the classification with asymmetric loss and adding another loss for separating samples with similar representation but different labels~\cite{DBLP:journals/kbs/ZhangLSXTH24-halb}.
The Hierarchy-aware Information Lossless contrastive Learning (HILL) model uses BERT as a text encoder and a graph encoder together with a hierarchy-aware contrastive loss~\cite{hill2024}.
HGBL (note, different from HBGL described earlier) is another model based on contrastive learning on the label and text features using a graph-encoder and BERT as text encoder~\cite{DBLP:journals/npl/ZhangDLZ25-hgbl}.
RADAr uses an encoder-decoder architecture for hierarchical text classification without requiring a graph encoder~\cite{radar}.
It uses RoBERTa as a text encoder and a custom autoregressive decoder with two transformer layers to generate the labels in level-wise order.
HYDRA simplifies this architecture and relies solely on RoBERTa as the text encoder, together with level-wise losses to model the taxonomy~\cite{hydra}.  

The methods discussed so far rely on a small encoder-only language model.
Retrieval-style ICL is an approach for hierarchical text classification using a large-language model and few-shot in-context learning~\cite{DBLP:journals/corr/abs-2406-17534-retrieval-style}.

\section{Experimental Apparatus}\label{sec:apparatus}

We introduce the benchmark datasets we identified for single-label, multi-label, and hierarchical text classification.
We provide an overview of the models considered from the different families of text classification approaches. 
While the vast majority of results are taken from the literature, we also conduct our own runs. 
For these, we describe our procedure, choice of hyperparameters and their optimization, and evaluation measures.

\subsection{Datasets}\label{sec:datasets}

Our quantitative comparison focuses on topic classification, while including popular sentiment analysis datasets as control: 
MovieReviews for single-label classification, and GoEmotions for multi-label classification.

\paragraph{Single-label Datasets} 
We use the benchmark datasets 20ng, R8, R52, ohsumed, and MR with their standard train-test splits.
Twenty Newsgroups (20ng)\footnote{\url{http://qwone.com/~jason/20Newsgroups/}} (bydate version) contains long posts categorized into 20 newsgroups.
R8 and R52 are subsets of the R21578~\cite{reuters} news dataset with 8 and 52 classes, respectively.  
It is a collection of documents that appeared on Reuters newswire in 1987. 
Ohsumed\footnote{\url{http://disi.unitn.it/moschitti/corpora.htm}} is a corpus of medical abstracts from the MEDLINE database that are categorized into diseases (one per abstract). 
Movie Reviews (MR)\footnote{\url{https://www.cs.cornell.edu/people/pabo/movie-review-data/}}~\cite{pang-lee-2005-seeing}, split by Tang~\etal~\cite{DBLP:conf/kdd/TangQM15}, is a binary sentiment analysis dataset on sentence level. 
Table~\ref{tab:datasets} shows the dataset characteristics. 

\begin{table}[ht]
    \centering
    \caption{Characteristics of the single-label text classification datasets. 
    We show the number of documents N and
    the standard train-test split.
    \#C is the number of classes.
    Finally, we report the documents' average length and standard deviation.
    }
    \label{tab:datasets}
    \begin{tabular}{lrrrrr}
    \toprule
    \textbf{Dataset} & \textbf{N}       & \textbf{\#Train} & \textbf{\#Test}  & \textbf{\#C} & \textbf{Avg. length}   \\
    \midrule                                                              
    20ng    & 18,846  & 11,314  & 7,532   & 20        & 551 $\pm$ 2,047 \\
    R8      & 7,674   & 5,485   & 2,189   & 8         & 119 $\pm$ 128   \\
    R52     & 9,100   & 6,532   & 2,568   & 52        & 126 $\pm$ 133   \\
    ohsumed & 7,400   & 3,357   & 4,043   & 23        & 285 $\pm$ 123   \\
    MR      & 10,662  & 7,108   & 3,554   & 2         & 25 $\pm$ 11     \\
    \bottomrule
    \end{tabular}
\end{table}

\paragraph{Multi-label and Hierarchical Datasets}
\label{sec:multi-label-datasets}
Table~\ref{tab:multilabeldatasets} shows the characteristics of the multi-label and hierarchical datasets.
Reuters-21578 (R21578)~\cite{reuters} is a popular dataset for multi-label classification.
We use the train-test split from NLTK.\footnote{\url{https://www.nltk.org/book/ch02.html}}
The labels in R21578 are not hierarchically organized.
RCV1-V2 is a newer version of the R21578 dataset containing a much larger amount of hierarchically categorized newswire stories. 
For RCV1-V2, we use the train-test split proposed by Lewis et al.~\cite{rcv1-v2}. 
EconBiz~\cite{DBLP:conf/jcdl/MaiGS18} is a dataset containing scientific papers in economics.
It provides the titles of a metadata export as well as the full text of papers up to 2017.
We use the split provided by the authors for comparability.
GoEmotions is a corpus of comments extracted from Reddit, with human annotations to 27 emotion categories~\cite{demszky2020goemotions}. 
We use the same train-test split as in the original paper. 
GoEmotions does not have a hierarchical label structure. Amazon-531~\cite{hiddenfactors} contains
49,145 product reviews and a three-level class taxonomy consisting of 531 classes. 
DBPedia-298~\cite{Lehmann2015DBpediaA} includes 245,832 Wikipedia articles and a three-level class taxonomy with 298 classes. For Amazon-531 and DBPedia-298, we use the same train-test split as in TaxoClass~\cite{shen-etal-2021-taxoclass}.
NYT AC~\cite{sandhaus} contains New York Times articles from 1987 to 2007. We use the train-validation-test split from HiAGM~\cite{hiagm}.
The Web of Science (WOS) dataset contains 46,985 scientific papers~\cite{DBLP:conf/icmla/KowsariBHMGB17}, each with exactly one parent and one child label.
In the hierarchical datasets (Hier.=Y), each label set includes the more general labels along the path up to the root of the hierarchy.

\begin{table*}[ht]
    \centering
    \caption{Characteristics of the multi-label and hierarchical classification datasets. 
    We show the same statistics for the single-label datasets.
    In addition, we report the average number of class labels per document and whether the dataset comes with a hierarchy (Hier.).
    }
    \label{tab:multilabeldatasets}
    \begin{tabular}{lrrrrrrr}
    \toprule
    \textbf{Dataset} & \textbf{Hier.} & \textbf{N} & \textbf{\#Train} & \textbf{\#Test} & \textbf{\#C}  & \textbf{Labels per doc.} \\    
    \midrule                                                               
    R21578   & N & 10,788  & 7,769  & 3,019 & 90  &  1.24 $\pm$ 0.75   \\ 
    RCV1-V2  & Y   &  804,414  & 23,149   & 781,265 & 103  & 3.24 $\pm$ 1.40  \\ 
    EconBiz  &  Y & 1,064,634  & 994,015  & 70,619 & 5,661  & 4.36 $\pm$ 1.90  \\ 
    GoEmotions  & N  & 48,837  & 43,410  & 5,427 & 28   & 1.18 $\pm$ 0.42   \\ 
    Amazon-531  &  Y   & 49,145  &  29,487  &  19,658  & 531 & 2.93 $\pm$ 0.26  \\  
    DBPedia-298  &  Y   &  245,832  &  196,665  & 49,197 & 298   & 3.00 $\pm$ 0.00\\ 
    NYT AC  & Y   &  36,471  & 29,179    & 7,292 & 166 &  7.59 $\pm$ 5.61  \\   
    WOS & Y & 46,985 & 37,588 & 9,397 & 141 & 2.00 $\pm$ 0.00\\
   \bottomrule
    \end{tabular}
\end{table*}

\subsection{Methods and Complementing Experiments}

We build our quantitative comparison on existing studies such as Ding~\etal\cite{DBLP:conf/emnlp/DingWLLL20}, Ragesh~\etal\cite{DBLP:conf/wsdm/RageshSIBL21}, Li~\etal\cite{DBLP:journals/corr/abs-2405-11524}, and Galke~\etal\cite{galkescherp-acl2022}.

\paragraph{BoW-based Methods}
For the methods based on Bag of Words (BoW), we rely on the study by Galke~\etal\cite{galkescherp-acl2022}, who have evaluated various variants of an MLP operating on a bag-of-words.
These include an MLP with one wide hidden layer
 and two hidden layers and different text representations as input, such as TF-IDF weighting, and using pre-trained word embeddings such as GloVe.
We further list the numbers for fastText, SWEM, and logistic regression from Ding \etal~\cite{DBLP:conf/emnlp/DingWLLL20}.
Motivated by \cite{DBLP:journals/corr/abs-2211-02563}, we complement these numbers by running SVMs on unigram and trigram features to include strong classical machine learning methods~\cite{DBLP:conf/kcap/GalkeMSBS17}.

\paragraph{Recurrent and Convolutional Neural Networks} We include scores of sequence-based models such as LSTMs~\cite{DBLP:conf/emnlp/DingWLLL20,zhao2021sequential} and CNNs~\cite{DBLP:conf/ijcnlp/ZhangW17,DBLP:conf/emnlp/Kim14}.
In contrast to the BoW-based models, these consider the sequence of the textual input and exploit this to train the classifier.

\paragraph{Small Language Models}
We run experiments with various pre-trained language models.
The numbers reported in the literature for BERT applied to the same datasets vary widely across papers. 
We fine-tune BERT and some of the most popular encoder-only language models, including DistilBERT, RoBERTa, DeBERTa,  ALBERTv2, and ERNIE 2.0 ourselves.
Furthermore, we conducted runs with the encoder-decoder models BART and T5, as there were no numbers available in the literature.

\paragraph{Large Language Models}
We found numbers for \LLMs including GPT-3.5 (\texttt{text-davinci-003}) with the CARP prompting strategy~\cite{carp}, a sophisticated few-shot learning technique that relies on sampling examples based on RoBERTa-large embeddings (R.S. variant) and voting among multiple runs. 
Note that in-context learning evaluations of LLMs, such as CARP, do not make use of the full training set but only a few examples. 
A different strategy we include here is to fine-tune the parameters of LLMs for text classification, as done in RGPT~\cite{pushing-the-limit-11346014-rgpt}. Specifically, RGPT consists of an ensemble of fine-tuned Llama-2 models~\cite{touvronLlamaOpenFoundation2023}.

\paragraph{Synthetic Text-Graph Methods}

We review both transductive and inductive graph-based methods for text classification.
These include TextGCN along with its successors HeteGCN, TensorGCN, HyperGAT,
DADGNN, Simplified GCN (SGC)~\cite{DBLP:conf/icml/WuSZFYW19}, and many others.
Ragesh~\etal\cite{DBLP:conf/wsdm/RageshSIBL21} evaluated a variant of TextGCN that is capable of inductive learning, which we include in our results. 

\paragraph{Hierarchy-based Methods}
\label{sec:hiagm}
\label{sec:Hierarchy-based-Text-Classification}
For hierarchy-based text classification, we report results for models that use graph encoders such as BERT+HiMatch, HiAGM, HGCLR, and HBGL.
We report models that do not use a graph encoder, \eg the encoder-decoder-based Seq2Tree and RADAr, as well as HYDRA, which uses layer-wise losses on RoBERTa.
For comparison, we run experiments with the BoW-based method WideMLP~\cite{galkescherp-acl2022}, which considers the classes as a set.
More training is needed for transformer-based models on hierarchical datasets than on multi-label datasets~\cite{hydra}.

\subsection{Procedure}

We distinguish between single-label and multi-label text classification settings.
We apply standard train-test splits unless there is no default split provided (see Section~\ref{sec:datasets}).

We consider an inductive setting, where the test documents remain unseen until test time, \ie they are not available for training or preprocessing. The distinction only matters for graph-based approaches because BoW-based and sequence-based models are usually inductive.

To avoid bias in the comparability of the results, we carefully checked all relevant parameters, such as the train/test splits, the number of classes in the datasets, whether datasets have been pre-processed to make the task substantially easier, and the evaluation metrics (see methodology in Section~\ref{sec:methodology}). 

We repeat all of our own experiments five times with a different random initialization of the parameters and report the mean and standard deviation of these five runs.
To tune the hyperparameters of the multi-label classification models, we choose randomly 20\% of the train set as a validation set. 
Below, we provide a detailed description of the hyperparameter optimization and evaluation procedures for the models that we have run ourselves.

\subsection{Hyperparameters}

Generally, we rely on the hyperparameter values and hyperparameter optimization strategies reported by the original works.
In cases where we run experiments, we describe below how we set the hyperparameter values for the single-label, multi-label, and hierarchical text classification tasks.
For each case, we follow the families of classification approaches.

\paragraph{Single-label Case}
For the unigram and word-trigram SVM models, we first transform the features via TF-IDF and then employ a linear support vector classifier with hinge loss and default hyperparameters (l2 penalty, regularization strength $C=1$).

For \textit{fine-tuning the small language models} such as 
BERT-base, DistilBERT, 
RoBERTa-base, 
DeBERTa-base, ERNIE, and ALBERT, we adopt the fine-tuning strategy of Galke and Scherp~\cite{galkescherp-acl2022}:
We fine-tune each model for 10 epochs with a linearly decaying learning rate. We conduct a manual search among learning rates between $10^{-5}$ and $5 \cdot 10^{-5}$, as common for fine-tuning pre-trained language models~\cite{DBLP:conf/naacl/DevlinCLT19,T5}. Our search range for batch sizes was between $4$ and $128$ -- which in turn control the number of optimizer steps. 
Specifically, we use $\mathrm{lr}=4.5\cdot 10^{-5}$ for DistilBERT, $4 \cdot 10^{-5}$ and for RoBERTa, $2 \cdot 10^{-5}$ for DeBERTa, $2.5 \cdot 10^{-5}$ for ERNIE, and $1 \cdot 10^{-5}$ for ALBERT and BERT-large. 
The effective batch sizes (i.e., taking gradient accumulation into account) are 128 for DistilBERT, 16 for DeBERTa and BERT-large, and 32 for RoBERTa, ERNIE, and ALBERT.
The transformers' output sequence is pooled according to the respective model's suggested strategy.
For example, for BERT and most BERT-like models, the final representation of the first token is used (the inserted special CLS token) and fed into an MLP module. 
We use the uncased versions where available, \eg BERT, and the cased variants of later models, \eg RoBERTa, which no longer provide a separate uncased variant.
We fine-tune BART and T5 for $10$ epochs with a learning rate of $5\cdot
10^{-5}$. 
For this, we use the discriminative variant, see Section~\ref{sec:finetune-encdec-slms}, \ie attach a newly initialized classifier head that processes the last token's representation and fine-tune all model parameters against a categorical cross-entropy loss.

\paragraph{Multi-label Case}
For \textit{training the \mlp{}} in the multi-label case, we tune the hyperparameters using a manual search on the R21578 dataset with learning rates between $10^{-3}$ and $10^{-1}$ and batch sizes between $8$ and $256$.
We noticed that higher learning rates are beneficial combined with a TF-IDF input representation and $100$ epochs of training. We thus train the MLPs consistently with a learning rate of $10^{-1}$ for all datasets.  We increase the batch size according to the dataset size in order to limit the overall training time: For the smaller datasets (R21578, GoEmotions, Amazon-531, NYT AC, RCV1-V2), we use a batch size of 8. 
For DBPedia-298, the batch size is $32$, and for EconBiz, it is $256$.

All models are trained with Adam~\cite{DBLP:journals/corr/KingmaB14} to minimize binary cross-entropy loss. 
At test time, class labels are assigned depending on a threshold on the class-specific output.
It is common to threshold the sigmoidal output units at values of $\lambda = 0.5$~\cite{zhang,Tsoumakas}. 
However, Galke~\etal\cite{DBLP:conf/kcap/GalkeMSBS17} had found that a smaller threshold such as $\lambda=0.2$ can be advantageous, especially in setups with an imbalanced label distribution.
In pre-experiments with \mlp, we tested different thresholds from $\lambda=0.5$ to $\lambda=0.1$ ($0.1$ steps) and experienced similar results as reported in Galke~\etal\cite{DBLP:conf/kcap/GalkeMSBS17}.

For the \emph{encoder-only models}, \ie the variants of BERT, we use a manual search to find the best hyperparameters using a search range in learning rate between $10^{-5}$ and $5 \cdot 10^{-5}$, batch sizes between $4$ and $128$, and epochs in $\{5,10,15\}$. 
We again fine-tune all parameters of the model, as we do for single-label classification, with Adam to
minimize binary cross-entropy loss. 
In the pre-experiments with R21578, we found that small batch sizes are preferable for fine-tuning these models.
We used a learning rate of $5 \cdot 10^{-5}$ 
with a batch size of $4$ for all data sets.
We fine-tune the models on the datasets for either $5$ or $15$ epochs for multi-label training, following~\cite{galkescherp-acl2022}.
DBPedia-298 and GoEmotions had the best results with $5$ epochs as the validation loss increased in subsequent iterations.
The encoder models employ a threshold of $\lambda=0.5$ after pre-experiments with $0.2$ and $0.5$. 

For the encoder-decoder models BART and T5, we use Adam to optimize all parameters (incl. classifier head) against binary cross-entropy with a learning rate of $5 \cdot 10^{-5}$. We use the best identified number of epochs and batch sizes as determined for the encoder-only models, and we again test thresholds for class assignment $\lambda = 0.2$ and $0.5$, out of which the $0.2$ threshold led to better results. 

\paragraph{Hierarchical Case}
The runs on the datasets NYT and RCV1-v2 use the same hyperparameters as described for the multi-label case above. For WOS, we train WideMLP for 100 epochs with a learning rate of $3 \cdot 10^{-1}$ and a batch size of 128.
The discriminative variants of BART and T5 on WOS were trained with a linearly decaying learning rate of $2 \cdot 10^{-5}$, weight decay of $0.1$, and gradient clipping to $1.0$.
For other hierarchy-based methods, we rely on numbers from the literature.

\subsection{Measures}
We report accuracy as the evaluation metric for single-label datasets.
Note that the accuracy is equivalent to Micro-F1 in single-label classification~\cite{galkescherp-acl2022}.
We report the mean accuracy and standard deviation (SD) over five runs for neural network methods, which rely on random initialization and other noise sources during training, such as dropout. 
For those models where we rely on numbers from the literature, we check if multiple runs are reported and include the corresponding information in our report. 
Note that the exact number of runs may differ from paper to paper. Most papers report five runs (if they have multiple runs), but others report ten runs.

For the multi-label datasets, we follow Galke~\etal\cite{DBLP:conf/kcap/GalkeMSBS17} and report the sample-based F1 measure. 
We chose this sample-based evaluation measure because it reflects the classification quality of each document separately~\cite{DBLP:conf/jcdl/MaiGS18}. 
The sample-based F1 measure is calculated by the harmonic mean of precision and recall for each example individually, and then these scores are averaged.
For hierarchical text classification, we report globally averaged Micro-F1 and class-averaged Macro-F1, as common in the hierarchical text classification literature.

\section{Quantitative Comparison}\label{sec:results}

We present the results of our quantitative comparison, starting with single-label text classification.
Subsequently, we present our results of a sensitivity analysis of transformer models regarding the fine-tuning learning rate.
This is followed by the results on multi-label and hierarchical text classification.
We bold the best numbers and underline the second best per model family.

\subsection{Single-label Text Classification}

\renewcommand{\mytextsubscript}[1]{{\color{black}~\textsubscript{#1}}}
\newcommand\mycaption{Results for the single-label
text classification datasets. For our experiments, we report the mean accuracy
and standard deviation (SD) over five runs. For numbers from the literature, we
report the SD if available. GPT-3.5 methods use 16 examples per class for
in-context learning, and the (R.S)-variant uses a Sup-SimCSE-RoBERTa-Large sampler to select
these examples. Column ``Provenance'' reports the source. d.\,f. is short for the use of a different variant of the dataset or a different split, and thus, the number is omitted to ensure comparability.}
\newcommand\mylabel{\label{tab:results_textclf_micro}}
\begin{table*}
\mytablefontsize
\centering
\caption{\mycaption{}}\mylabel{}
\begin{tabular}{llllllr}
\toprule
\textbf{Inductive Setting} & \textbf{20ng} & \textbf{R8} & \textbf{R52} & \textbf{ohsumed} & \textbf{MR} & \textbf{Provenance}\\
\midrule

\textit{BoW-based Methods} & & & & & & \\

    Logistic regression with TF-IDF & 83.70 & 93.33 & 90.65 & 61.14 & 76.28 & \mycite{DBLP:conf/wsdm/RageshSIBL21}\\

Unigram SVM with TF-IDF & 83.44 & 97.49 & 94.70 & 67.40 & 76.36 & \myflag{} \\
    Trigram SVM with TF-IDF & 83.39 & 97.21 & 93.85 & \best{69.30} & \best{77.35} & \myflag{} \\

    XGBoost with TF-IDF & 73.29 & 94.75 & 88.82 & 58.27 & 64.46 & \myflag{}\\
    
     \mlp with TF-IDF & 84.20\mytextsubscript{0.16} & 97.08\mytextsubscript{0.16} & 93.67\mytextsubscript{0.23} & 66.06\mytextsubscript{0.29} & 76.32\mytextsubscript{0.17} & \cite{galkescherp-acl2022}\\
    \mlp & 83.31\mytextsubscript{0.22} & \secondbest{97.27}\mytextsubscript{0.12} & 93.89\mytextsubscript{0.16} & 63.95\mytextsubscript{0.13} & \secondbest{76.72}\mytextsubscript{0.26} & \cite{galkescherp-acl2022}\\ 
    \mlp-2  & 81.02\mytextsubscript{0.23} & 96.61\mytextsubscript{1.22} & \secondbest{93.98}\mytextsubscript{0.23} & 61.71\mytextsubscript{0.33} & 75.91\mytextsubscript{0.51} & \cite{galkescherp-acl2022}\\
    GloVe+\mlp  & 76.80\mytextsubscript{0.11} & 96.44\mytextsubscript{0.08} & 93.58\mytextsubscript{0.06} & 61.36\mytextsubscript{0.22} & 75.96\mytextsubscript{0.17} & \cite{galkescherp-acl2022}\\ 
    GloVe+\mlp-2  & 76.33\mytextsubscript{0.18} & 96.50\mytextsubscript{0.14} & 93.19\mytextsubscript{0.11} & 61.65\mytextsubscript{0.27} & 75.72\mytextsubscript{0.45}& \cite{galkescherp-acl2022}\\

SWEM & \secondbest{85.16}\mytextsubscript{0.29} & 95.32\mytextsubscript{0.26} & 92.94\mytextsubscript{0.24} &  63.12\mytextsubscript{0.55} & 76.65\mytextsubscript{0.63} & \mycite{DBLP:conf/emnlp/DingWLLL20}\\
    fastText & 79.38\mytextsubscript{0.30} & 96.13\mytextsubscript{0.21} &  92.81\mytextsubscript{0.09} &  57.70\mytextsubscript{0.49} & 75.14\mytextsubscript{0.20} & \mycite{DBLP:conf/emnlp/DingWLLL20}\\

CFE-IterativeAdditive & 
\best{85.51}\mytextsubscript{0.04} & 
\best{97.94}\mytextsubscript{0.02} & 
\best{95.13}\mytextsubscript{0.04} & 
\secondbest{68.90}\mytextsubscript{0.02} & 
--- & 
\mycite{ATTIEH2023110215} \\

\midrule
     
    \textit{Sequence-based Methods} & & & & & & \\
    CNN+GloVe & 82.15 & 95.71 & 87.59 & 58.44 & 77.75 & \mycite{DBLP:conf/coling/HuangCC22} \\
    CNN-non-static & --- & --- & --- & --- & 81.5 & \mycite{DBLP:conf/emnlp/Kim14} \\

Word2Vec+CNN & --- & --- & --- & --- & 81.24 & \mycite{DBLP:conf/ijcnlp/ZhangW17} \\
    GloVe+CNN & --- & --- & --- & --- & 81.03 & \mycite{DBLP:conf/ijcnlp/ZhangW17} \\

LSTM w/ pre-training & 75.43\mytextsubscript{1.72} & 96.09\mytextsubscript{0.19} &  90.48\mytextsubscript{0.86} &  51.10\mytextsubscript{1.50} & 77.33\mytextsubscript{0.89} & \mycite{DBLP:conf/emnlp/DingWLLL20}\\
 Bi-LSTM\mytextsubscript{GloVe embeddings}  & --- & 96.31  &  90.54 & ---  & 77.68 & \mycite{zhao2021sequential}    \\

GPT-3.5\mytextsubscript{finetuning via OpenAI API}
& --- & --- & 95.27\mytextsubscript{0.55} & 51.84\mytextsubscript{0.45} & --- & \cite{DBLP:journals/corr/abs-2405-11524} \\

Bloom-7.1B\mytextsubscript{4-bit LoRA finetuning} & 
--- & --- & --- & 
67.54\mytextsubscript{0.6} & --- & \cite{DBLP:journals/corr/abs-2405-11524} \\

Llama2-7B\mytextsubscript{4-bit LoRA finetuning} & 
--- & --- & --- & 
67.66\mytextsubscript{0.72} & --- & \cite{DBLP:journals/corr/abs-2405-11524} \\

Llama3-8B\mytextsubscript{4-bit LoRA finetuning} & 
--- & --- & --- & 
68.02\mytextsubscript{0.26} & --- & \cite{DBLP:journals/corr/abs-2405-11524} \\ 

GPT-3.5\mytextsubscript{16-shot}  & --- &  91.58 & 91.56 & --- & 89.15  & \mycite{carp} \\
GPT-3.5+CoT\mytextsubscript{16-shot} & --- & 92.49 & 92.03  & --- &  89.91 & \mycite{carp} \\
GPT-3.5+CARP\mytextsubscript{16-shot}& --- & 97.60  & 96.19  & --- &  90.03& \mycite{carp} \\

  GPT-3.5 (R.S.)\mytextsubscript{16-shot}& --- & 95.57 & 95.79 & --- & 90.90  & \mycite{carp} \\
  GPT-3.5+CoT (R.S.)\mytextsubscript{16-shot}& --- & 95.59 & 95.89 & --- & \secondbest{90.17} & \mycite{carp} \\
  GPT-3.5+CARP+vote (R.S.)\mytextsubscript{16-shot}& --- & 98.78  & 96.95  & --- & \best{92.39} & \mycite{carp}\\

RGPT w. RoBERTa-base              & --- & --- & --- & 74.70 & d.\,f. & \cite{pushing-the-limit-11346014-rgpt}\\

RGPT w. $k$=7 Llama 2 + LoRA & --- & --- & --- & \best{77.41} & d.\,f. & \cite{pushing-the-limit-11346014-rgpt}\\

QLFRw. LLaMA2-13B + LoRA\mytextsubscript{20-shot}  & --- & --- & --- &  61.10 & 81.70 & \cite{DBLP:journals/corr/abs-2401-03158}  \\

LLMEmbed\mytextsubscript{BERT+RoBERTa+Llama embeddings} 
& --- & 98.22 & 95.68 & --- & d.\,f. & \cite{liu2024llmembedrethinkinglightweightllms} \\

BERT-base 
    & 87.21\mytextsubscript{0.18} & 98.03\mytextsubscript{0.24} & 96.17\mytextsubscript{0.33} & 71.46\mytextsubscript{0.54} & 86.61\mytextsubscript{0.38} & \mycite{galkescherp-acl2022}\\

BERT-large & 85.83\mytextsubscript{0.64} & 97.98\mytextsubscript{0.29} & 96.41\mytextsubscript{0.28} & 72.69\mytextsubscript{0.63} & 88.22\mytextsubscript{0.21} & \myflag{} \\
   
    RoBERTa-base 
    & 86.80\mytextsubscript{0.51} & 98.19\mytextsubscript{0.18} & 97.13\mytextsubscript{0.10} & 75.08\mytextsubscript{0.42} & 88.68\mytextsubscript{0.29} & \myflag{} \\

    DeBERTa-base   
    & 87.60\mytextsubscript{0.45} & 98.30\mytextsubscript{0.20} 
      & 97.10\mytextsubscript{0.13} & \secondbest{75.94}\mytextsubscript{0.33} & 89.98\mytextsubscript{0.26} 
     & \myflag{} \\
    
    DistilBERT-base & 86.90\mytextsubscript{0.04} & 97.93\mytextsubscript{0.11} & 96.89\mytextsubscript{0.12} & 71.65\mytextsubscript{0.38} & 85.11\mytextsubscript{0.25} & \myflag{}\\

     ERNIE 2.0 
     & 87.79\mytextsubscript{0.29} & 97.95\mytextsubscript{0.16} & 96.96\mytextsubscript{0.23} & 73.33\mytextsubscript{0.30} & 89.19\mytextsubscript{0.24} & \myflag{} \\
    
     ALBERTv2 
     & 82.08\mytextsubscript{0.30} & 97.88\mytextsubscript{0.22} & 94.95\mytextsubscript{0.20} & 62.31\mytextsubscript{2.11} & 86.28\mytextsubscript{0.21} & \myflag{} \\
    
    AM-BERT\mytextsubscript{BERT-base} & \secondbest{89.03} & \secondbest{98.43} & \secondbest{97.17} & 73.47 & 86.83 & \mycite{am-bert} \\
    AM-RoBERTa\mytextsubscript{RoBERTa-base} & \best{90.32} & \best{98.97} & \best{98.12} & 73.89 & 89.75 & \mycite{am-bert} \\
    
BART-base\mytextsubscript{discriminative}
& 87.24\mytextsubscript{0.17} & 98.14\mytextsubscript{0.15} & 96.99\mytextsubscript{0.15} & 75.38\mytextsubscript{0.21} & 87.41\mytextsubscript{0.45} & \myflag{} \\

T5-base\mytextsubscript{discriminative}  
& 85.90\mytextsubscript{0.23} & 97.98\mytextsubscript{0.05} & 95.16\mytextsubscript{0.13} & 72.14\mytextsubscript{0.29} & 90.10\mytextsubscript{0.38} & \myflag\\

LFTC (compression w. $1$-NN) \new & 81.4 & 96.5 & 90.6 & 43.5 & --- & 
\cite{mao2024lowresourcefasttextclassification} \\

\midrule

    \textit{Graph-based Methods} & & & & & & \\
Text-level GNN & --- & 97.8\mytextsubscript{0.2} &  94.6\mytextsubscript{0.3} & 69.4\mytextsubscript{0.6} & --- & 
\mycite{DBLP:conf/emnlp/HuangMLZW19}
\\   

            TextING-M & --- & 98.13\mytextsubscript{0.31} & 95.68\mytextsubscript{0.35} &
   70.84\mytextsubscript{0.52} &
   80.19\mytextsubscript{0.31} &
   \mycite{texting_acl2020}
   \\

TextGCN & 80.88\mytextsubscript{0.54} & 94.00\mytextsubscript{0.40} & 89.39\mytextsubscript{0.38} & 56.32\mytextsubscript{1.36} & 74.60\mytextsubscript{0.43} & \mycite{DBLP:conf/wsdm/RageshSIBL21}\\
     HeteGCN & 84.59\mytextsubscript{0.14} & 97.17\mytextsubscript{0.33} & 93.89\mytextsubscript{0.45} & 63.79\mytextsubscript{0.80} & 75.62\mytextsubscript{0.26} & \mycite{DBLP:conf/wsdm/RageshSIBL21}\\
     HyperGAT-ind & 84.63 & 97.03 & 94.55 & 67.33 & 77.08\mytextsubscript{0.27} &\mycite{DBLP:conf/coling/HuangCC22}\\
    DADGNN & --- & 98.15\mytextsubscript{0.16} & 95.16\mytextsubscript{0.22} & --- & 78.64\mytextsubscript{0.29} & \mycite{DBLP:conf/emnlp/LiuGG0F21}\\

SGNN & ---  & 98.09 & 95.46 & --- & 80.58  & \mycite{zhao2021sequential}    \\
    ESGNN & ---  & \secondbest{98.23} & 95.72 & --- & 80.93 & \mycite{zhao2021sequential}    \\

    C-BERT (ESGNN+BERT)\mytextsubscript{BERT-base}  & --- & \best{98.28} & \best{96.52} & --- & \secondbest{86.06} &         \mycite{zhao2021sequential}    \\

ConTextING-RoBERTa\mytextsubscript{RoBERTa-base}  & 85.00 & 98.13 & \secondbest{96.40} & \secondbest{72.53} & \best{89.43} &
    \mycite{DBLP:conf/coling/HuangCC22} 
    \\

TextSSL & \secondbest{85.26}\mytextsubscript{0.28} & 97.81\mytextsubscript{0.14} & 95.48\mytextsubscript{0.26} & 70.59\mytextsubscript{0.38} & 79.74\mytextsubscript{0.19} & \mycite{textssl2022}\\

GLTC & --- & 98.17 & 95.77 & 71.82 & 80.29 & \mycite{gltc2023} \\

InducT-GCN & --- & 96.53\mytextsubscript{0.17} & 93.23\mytextsubscript{0.15} & --- & --- & \mycite{induct-gcn} \\

MHGAT  &
\best{92.68}\mytextsubscript{0.30} &
97.65\mytextsubscript{0.47} &
94.78\mytextsubscript{0.37} &
\best{72.88}\mytextsubscript{0.84} &
78.09\mytextsubscript{0.73} &

\cite{mhgat}
\\ 

\bottomrule

\end{tabular}
\end{table*}

Table~\ref{tab:results_textclf_micro} shows the results of the single-label text classification on the five datasets.
We see that sequence-based transformers are generally the best methods, followed by the graph-based methods and BoW methods.

Among the BoW-based methods, the SVM variants are the strongest on ohsumed and MR, followed by WideMLP and SWEM.
The CFE-IterativeAdditive with its use of label matching to compute adapted global IDF weights TF-ICF is the best BoW model on 20ng, R8, and R52.
The strength of TF-IDF weighting is also shown by the Trigram SVM with TF-IDF and WideMLP with TF-IDF.
We further observe that pure BoW and TF-IDF-weighted BoW yield better results than approaches that exploit pre-trained word embeddings such as GloVe-MLP, fastText, and SWEM.

Regarding the sequence-based methods, a GPT~3.5 with 16 shots and CARP and voting achieves the best results on MR.
It requires a fine-tuned RoBERTa, which is used to select the most relevant shots.
RGPT achieves the highest score on ohsumed with a sequence of seven fine-tuned Llama 2 models and low-rank adaptation.
AM-RoBERTa, using two additional attention matrices for keywords and term dependencies, achieves the best results on 20ng, R8, and R52. 
DeBERTa-base attains the highest scores among the encoder-only models. 
The margin to a BERT-base model is most notable on ohsumed ($75.9$ vs. $71.5$) and on MR ($90.0$ vs. $86.6$).
 BERT-large improves the scores over BERT-base by a small 1 point for the ohsumed and MR datasets, but the inverse of a performance decrease of 1 point is recorded for 20ng. 
For R8 and R52, both BERT-base and BERT-large achieve about the same performance. 
BART-base is a strong single-label classifier, outperforming BERT-base on all single-label datasets and BERT-large on four datasets. 
T5-base generally falls behind BART-base, except for the MR datasets, where it has a $2.69$-point advantage. 
BART and T5 cannot beat DeBERTa, except for MR, where T5 shows a marginal improvement of $0.12$ points over DeBERTa.

The graph-based MHGAT obtains the highest scores on the 20ng dataset among all method families.
However, numbers for GPT-3.5 with CARP and RGPT were not available for this dataset for comparison.
With the help of a pre-trained language model, models like C-BERT and ConTextING-RoBERTa attain higher scores on R8, R52, and MR.
The largest difference is found on the MR sentiment analysis dataset, where ConTextING-RoBERTa reaches $89.43$ compared to $77.08$ of HyperGAT-ind.
The best-performing graph-based model not using a pre-trained language model is TextSSL, closely followed by HyperGAT.
TextGCN is outperformed by a BoW-based logistic regression on four out of five datasets~\cite{DBLP:conf/wsdm/RageshSIBL21}.

The best pure graph-based method TextSSL is outperformed by T5-base by 10 points on the MR dataset.
Comparing DeBERTa with the best graph-based method ConTextING-RoBERTa, there is still superiority of the pure transformer, but the margin is smaller.

\subsection{Sensitivity Analysis to Fine-tuning Learning Rate for Single-label Classification}

Analyzing the results on single-label datasets reported in the literature, we noticed that the performances of BERT and other transformer models on the same datasets differ from paper to paper.
To shed light on the differences between BERT results on the same datasets, we provide an overview in Table~\ref{tab:transformer-comparison} and conduct additional runs with different learning rates.
In our additional runs, we kept preprocessing and other parameters like batch size and tuning effort the same.

We observe that there are substantial differences between the supposedly same BERT models applied on the same datasets reported in the literature.
For BERT-base, the difference is, in many cases, 2 and 3 points on the 20ng and ohsumed datasets, respectively.
For RoBERTa, we even observe deviations of more than 3 points on 20ng and 5 points on ohsumed despite using the same learning rate.

Some of the reported numbers for fine-tuned BERT models are even far behind the others.
For example, Yin et al.~\cite{gltc2023} report BERT-base results that are more than ten points behind the other papers on MR and ohsumed.
We hypothesize that this discrepancy is caused by a suboptimal choice of hyperparameters, \eg a too-high learning rate, which are unfortunately not provided.

On the R8, R52, and MR datasets, the results differ by not more than 1 point.
Remarkably, DistilBERT is quite sensitive to a small change in the learning rate by only $0.5 \cdot 10^{-5}$.
The difference is more than 1 point on R52 and even 2 points on ohsumed.

\renewcommand{\myheader}{
  \caption{Comparison of different transformer models and hyperparameter settings. We report mean accuracy and standard deviation over five runs on the single-label text classification datasets (inductive). Column ``Provenance'' reports the source.
    N/P refers to the case where the paper (or potential supplementary materials) did not provide information about the learning rate.
    }
    
    \label{tab:transformer-comparison}
}

\begin{table*}[!th]
    \mytablefontsize
    \centering
    \myheader{}
    \begin{tabular}{llllllr}
  \toprule
  \textbf{Inductive Setting}  & \textbf{20ng} & \textbf{R8} & \textbf{R52} & \textbf{ohsumed} & \textbf{MR} & \textbf{Provenance} \\
    \midrule
  
    BERT-base ($\mathrm{lr}=5 \cdot 10^{-5}$) & 87.21\mytextsubscript{0.18} & 98.03\mytextsubscript{0.24} & 96.17\mytextsubscript{0.33} & 71.46\mytextsubscript{0.54} & 86.61\mytextsubscript{0.38} & \myflag{}\\

BERT-base ($\mathrm{lr}=5 \cdot 10^{-5} $) & --- & --- & 
    95.55\mytextsubscript{0.36} & 
    65.71\mytextsubscript{0.30} & 
    --- & \cite{DBLP:journals/corr/abs-2405-11524} \\

BERT-base ($\mathrm{lr}=3.5 \cdot 10^{-5}$) & 87.31\mytextsubscript{0.21} & 98.19\mytextsubscript{0.12} & 97.13\mytextsubscript{0.16} & 73.54\mytextsubscript{0.45} & 86.86\mytextsubscript{0.10} & \myflag{} \\

BERT-base ($\mathrm{lr}= 2 \cdot 10^{-5}$)\textsuperscript{1)} & 85.20 & 97.73 & 96.22 & 70.53 & 85.71 & \mycite{am-bert} \\

BERT-base ($\mathrm{lr}=1 \cdot 10^{-5}$) & 84.54 & 97.26 & 96.26 & 68.74 & 85.88 &  \mycite{DBLP:conf/coling/HuangCC22} \\

BERT-base ($\mathrm{lr}=1 \cdot 10^{-5}$) & 85.3 & 97.8 & 96.4 & 70.5 & 85.7 & \mycite{DBLP:conf/acl/LinMSHKLW21} \\
    
BERT-base ($\mathrm{lr}=$ N/P)\textsuperscript{1)}
    & --- & 96.78 & 91.35 & 60.46 & 76.13 & \mycite{gltc2023}\\
    
BERT-base ($\mathrm{lr}=$ N/P) & --- & 98.2 & --- & --- & 85.7 & \mycite{stgcn} \\
        
BERT-base ($\mathrm{lr}=$ N/P) & 83.1 & --- & --- & --- & --- & \mycite{DBLP:conf/icml/ChaiWHWL20} \\

DistilBERT-base ($\mathrm{lr}=5 \cdot 10^{-5}$) & 86.24\mytextsubscript{0.26} & 97.89\mytextsubscript{0.15} & 95.34\mytextsubscript{0.08} & 69.08\mytextsubscript{0.60} & 85.10\mytextsubscript{0.33} & \myflag{}\\

DistilBERT-base ($\mathrm{lr}=4.5 \cdot 10^{-5}$) & 86.90\mytextsubscript{0.04} & 97.93\mytextsubscript{0.11} & 96.89\mytextsubscript{0.12} & 71.65\mytextsubscript{0.38} & 85.11\mytextsubscript{0.25} & \myflag{}\\

    RoBERTa-base ($\mathrm{lr}=4 \cdot 10^{-5}$) & 86.80\mytextsubscript{0.51} & 98.19\mytextsubscript{0.18} & 97.13\mytextsubscript{0.10} & 75.08\mytextsubscript{0.42} & 88.68\mytextsubscript{0.29} & \myflag{} \\
        
    RoBERTa-base ($\mathrm{lr}=4 \cdot 10^{-5}$) & 83.8 & 97.8 & 96.2 & 70.7 & 89.4 & \mycite{DBLP:conf/acl/LinMSHKLW21} \\
    RoBERTa-base ($\mathrm{lr}=1 \cdot 10^{-5}$) & 84.07 & 97.35 & 95.48 & 69.86 & 87.08 & \mycite{DBLP:conf/coling/HuangCC22}\\ 

RoBERTa-base ($\mathrm{lr}=$ N/P)\textsuperscript{1)} & 83.80 &  97.80 & 96.20 & 70.70 & 89.40  & \mycite{am-bert} \\

\bottomrule
\end{tabular}

\begin{tablenotes} 
\item \textsuperscript{1)} Authors applied special preprocessing of the input text. 
\end{tablenotes}

\end{table*}

\subsection{Multi-label Text Classification}

\renewcommand{\myheader}{
\caption{Results for the inductive multi-label text classification datasets. 
    We report the sample-based F1 metric to reflect how well the classifier performs on average per a set of new documents.
    Standard deviation across runs is denoted in braces.}\label{tab:results_multi-label_classification}
}

\begin{table*}[ht]
    \mytablefontsize
    \centering
    \myheader{}
    \begin{tabular}{llllllll}
  \toprule
  \textbf{Inductive Setting} & 
  \textbf{R21578} 
  & \textbf{RCV1-V2} & \textbf{EconBiz} & \textbf{Amaz.} & \textbf{DBPedia} & \textbf{NYT} & \textbf{GoEmo.}\\

\midrule
    \textit{BoW-based methods} \\
    WideMLP & \secondbest{80.41} & \secondbest{69.92}\mytextsubscript{0.11} & \secondbest{23.15} & \secondbest{59.92} & \secondbest{89.47} & \secondbest{62.38}\mytextsubscript{0.27} & \secondbest{37.13} \\
    TF-IDF WideMLP  & \best{88.15} & \best{81.51}\mytextsubscript{0.03} & \best{45.38} & \best{80.32} & \best{94.91} & \best{75.58}\mytextsubscript{0.09} & \best{40.07} \\

    \midrule
    \multicolumn{8}{l}{\textit{Sequence-based methods}} \\

    BERT-base 
    & 92.21 & 88.16\mytextsubscript{0.16} & \secondbest{42.08} & 86.69 & 97.66 & 79.11\mytextsubscript{0.22} & 54.18\\
    BERT-large & \secondbest{92.23} & \secondbest{88.83}\mytextsubscript{0.17} & 33.62 & 88.34 & \best{97.69} & \secondbest{80.32}\mytextsubscript{0.39}& 54.02\\
    DistilBERT 
    & 92.11 & 87.50\mytextsubscript{0.11} & 39.41 & 87.47 & 97.58 & 79.18\mytextsubscript{0.17} & \secondbest{55.95} \\

    RoBERTa-base  
    & 90.85 & 88.62\mytextsubscript{0.21} & 40.56 & 86.21 & 97.26 & 79.14\mytextsubscript{0.55} & 54.64\\

    DeBERTa-base 
    & 91.24 & 88.45\mytextsubscript{0.19} & 41.43 & \best{89.21} & 97.65 & 79.95\mytextsubscript{0.40} & \best{56.51}\\

    BART-base\mytextsubscript{discriminative} 
    & 90.71\mytextsubscript{0.35} & 86.62\mytextsubscript{0.16} & 31.60 &
    85.02\mytextsubscript{0.57}&
    96.44\mytextsubscript{0.35} & 74.73\mytextsubscript{0.60} &  
    52.64\mytextsubscript{1.60}\\
    
    T5-base\mytextsubscript{discriminative} 
    & \best{92.72}\mytextsubscript{0.24} & \best{89.13}\mytextsubscript{0.11} & \best{45.63} & \secondbest{88.93}\mytextsubscript{0.33} & 
    \secondbest{97.68}\mytextsubscript{0.02} & \best{80.79}\mytextsubscript{0.15} &  55.01\mytextsubscript{0.62} \\

\bottomrule
    \end{tabular}
\end{table*}

\renewcommand{\myheader}{
    \caption{Micro and Macro F1 scores with standard deviation (where available) across five runs for hierarchical multi-label classification on three common benchmark datasets using Micro-F1 and Macro-F1 scores.}\label{tab:results_multi-label_classification-extra}
}

\begin{table*}[ht]
    \mytablefontsize
    \centering
    \myheader{}

    \begin{tabular}{lcccr}
    \toprule
    \textbf{Model} & \textbf{WOS}           & \textbf{NYT}           & \textbf{RCV1-V2} & \textbf{Provenance}\\
          & (Micro/Macro) & (Micro/Macro) & (Micro/Macro) & \\
         \midrule
    \textit{BoW-based methods} & & & & \\

WideMLP & \best{73.36}\mytextsubscript{0.14} / \best{62.61}\mytextsubscript{0.23} & \secondbest{57.18}\mytextsubscript{0.28} / \secondbest{21.96}\mytextsubscript{0.19}  & \secondbest{68.31}\mytextsubscript{0.12} / \secondbest{27.88}\mytextsubscript{0.49}  & \myflag{} \\

TF-IDF WideMLP & \secondbest{72.25}\mytextsubscript{0.08} / \secondbest{59.85}\mytextsubscript{0.19} & \best{74.53}\mytextsubscript{0.07} / \best{56.11}\mytextsubscript{0.16} & \best{80.45}\mytextsubscript{0.02} / \best{53.27}\mytextsubscript{0.09} & \myflag{} \\

\midrule
         
    \multicolumn{5}{l}{\textit{Sequence-based methods}}\\


BERT-base 
&  
86.19\mytextsubscript{0.11} / 80.23\mytextsubscript{0.20} &
79.07\mytextsubscript{0.22} / 67.63\mytextsubscript{0.42} & 
86.38\mytextsubscript{0.23} / 67.89\mytextsubscript{1.42} &
\cite{hydra} \\

BERT-base 
& 85.63 / 79.07 & 78.24 / 65.62 & 85.65 / 67.02 & \mycite{DBLP:conf/acl/WangWH0W22} \\ 

BERT-base  
& 86.26 / 80.58 & ---  & 86.26 / 67.35 & \mycite{DBLP:conf/acl/ChenMLY20} \\ 

BERT-large 
& 
86.66\mytextsubscript{0.31} / 80.80\mytextsubscript{0.38} &
80.69\mytextsubscript{0.08} / 70.27\mytextsubscript{0.39} & 
\secondbest{87.40}\mytextsubscript{0.11} / 70.15\mytextsubscript{0.39} &
\cite{hydra} \\
         
DeBERTa-base 
& 
86.82\mytextsubscript{0.11} / 80.85\mytextsubscript{0.48} &
81.21\mytextsubscript{0.17} / 71.33\mytextsubscript{0.26} & 
87.24\mytextsubscript{0.18} / 69.66\mytextsubscript{0.52} &
\cite{hydra} \\
   
DeBERTaV3-base 
& 
86.58\mytextsubscript{0.24} / 80.41\mytextsubscript{0.69} &
79.96\mytextsubscript{0.20} / 67.30\mytextsubscript{0.23} & 
86.77\mytextsubscript{0.18} / 67.49\mytextsubscript{1.87} &
\cite{hydra} \\
        
RoBERTa-base 
& 
86.35\mytextsubscript{0.15} / 80.16\mytextsubscript{0.13} & 
\secondbest{81.47}\mytextsubscript{0.15} / 71.45\mytextsubscript{0.45} &
87.38\mytextsubscript{0.17} / 68.60\mytextsubscript{1.40} &
\cite{hydra} \\


ModernBERT-base \new &
86.15\mytextsubscript{0.33} / 80.34\mytextsubscript{0.22} &	
79.83\mytextsubscript{0.53} / 68.99\mytextsubscript{0.54} & 86.20\mytextsubscript{0.42} / 67.43\mytextsubscript{0.66} & 
\cite{hydra}
\\

HYDRA\mytextsubscript{RoBERTa-base} &   86.91\mytextsubscript{0.18} / 81.22\mytextsubscript{0.40}  &  \best{81.96}\mytextsubscript{0.06} / 72.35\mytextsubscript{0.22} & \best{87.79}\mytextsubscript{0.11} / 70.08\mytextsubscript{0.76}   & \mycite{hydra} \\

BART-base\mytextsubscript{discriminative}
& 85.25\mytextsubscript{0.30} / 77.22\mytextsubscript{0.82} & 72.39\mytextsubscript{0.61} / 52.04\mytextsubscript{1.90} & 84.77\mytextsubscript{0.21} / 63.63\mytextsubscript{0.91} & our experiment \\
T5-base\mytextsubscript{discriminative}
& 79.21\mytextsubscript{1.09} / 60.98\mytextsubscript{3.41} & 79.06\mytextsubscript{0.16} / 66.04\mytextsubscript{0.10} & 87.41\mytextsubscript{0.14} / 69.54\mytextsubscript{0.56} & our experiment \\

BART-base\mytextsubscript{generative}
& 84.08 / 77.43 & 19.21 / 6.49 & 86.20 / 65.11 & \mycite{radar} \\
T5-base\mytextsubscript{generative}
& 82.03 / 74.62 & 46.71 / 20.06 & 84.9 / 57.01 & \mycite{radar} \\

HiGen\mytextsubscript{BART} & 87.39 / 81.45 & 80.89 / \secondbest{72.41} & --- & \mycite{higen} \\

SGM & 67.74 / 74.01 & 64.68 / \best{72.78} & 71.85 / 35.29 & \mycite{radar} \\
SGM w. T5-base & 85.83 / 80.79 & --- & 84.39 /  65.09 & \mycite{yu2022constrained}\\
Seq2Tree\mytextsubscript{T5-base} & \secondbest{87.20} / \best{82.50} & --- & 86.88 / 70.01 & \mycite{yu2022constrained} \\

RADAr\mytextsubscript{RoBERTa-base}  & $87.17\mytextsubscript{0.04}$ / $81.84\mytextsubscript{0.08}$ & $79.84\mytextsubscript{0.07}$ / $68.64\mytextsubscript{0.28}$ & $87.23\mytextsubscript{0.05}$ / $69.64\mytextsubscript{0.12}$ & \cite{radar} \\

HBGL\mytextsubscript{BERT-base} & \best{87.36} / \secondbest{82.00} & 80.47 / 70.19 & 87.23 / 71.07 & \mycite{hbgl} \\  

HLPD\mytextsubscript{T5-base} & 87.05\textsuperscript{1)} / 81.84\textsuperscript{1)}  &  --- / ---  &  88.22\mytextsubscript{0.12} \secondbest{71.19}\mytextsubscript{0.55}  & \mycite{hlpd} \\

LH-Mix\mytextsubscript{BERT-base} & 87.15\mytextsubscript{0.14} / 81.82\mytextsubscript{0.18} & 80.51\mytextsubscript{0.15} / 71.03\mytextsubscript{0.24} & 87.39\mytextsubscript{0.12} / \best{71.49}\mytextsubscript{0.50} & \mycite{lh-mix} \\

GPT 3.5 Turbo\mytextsubscript{zero-shot}  & ---  & --- &  51.35\mytextsubscript{0.18} / 32.20\mytextsubscript{0.30} & \cite{DBLP:conf/emnlp/UHGP23} \\

Retrieval-style ICL\mytextsubscript{16-shot Vicuna-7b + BERT-base}  \new &
81.12\mytextsubscript{0.26} / 73.72\mytextsubscript{0.17} & 
--- & --- & \cite{DBLP:journals/corr/abs-2406-17534-retrieval-style} \\

unLlama-7B\mytextsubscript{LLama 2 7B, LoRA-rank 12} &   84.13 / 74.32   & 77.20 / 61.68 & 84.45 / 61.42  & \mycite{hydra} \\

\midrule

\multicolumn{5}{l}{\textit{Hierarchy-based methods}} \\

BERT+HiMatch\mytextsubscript{BERT-base} & 86.70 / 81.06 & ---  & 86.33 / 68.66 & \mycite{DBLP:conf/acl/ChenMLY20} \\ 

HiAGM-TP+GCN & 85.82 / 80.28 & 74.97 / 60.83  &  83.96 / 63.35 & \mycite{hiagm}  \\


HGCLR\mytextsubscript{BERT-base} & 87.11 / 81.20 & 78.86 / 67.96 & 86.49 / 68.31 & \mycite{DBLP:conf/acl/WangWH0W22} \\ 

HGBL\mytextsubscript{BERT-base} \new & \secondbest{87.07} / 81.10 & 78.55 / 67.08 & \best{87.55} / 68.10 & \mycite{DBLP:journals/npl/ZhangDLZ25-hgbl} \\

HALB\mytextsubscript{BERT-base} & 87.45 / \best{82.04} & 79.56 / 69.28 & 86.94 / 69.32 & \mycite{DBLP:journals/kbs/ZhangLSXTH24-halb} \\

HILL\mytextsubscript{BERT-base} & 87.28 / 81.77 & \secondbest{80.47} / \secondbest{69.96} & 87.31 / \secondbest{70.12} & \mycite{hill2024} \\

HE-AGCRCNN & --- & --- & 77.8 / 51.3 & \mycite{pengHierarchicalTaxonomyAwareAttentional2021}\\

K-HTC \new & 87.29 / 81.69 & --- & --- & \cite{DBLP:conf/acl/LiuZHWZ0C23-k-htc} \\

HyILR\mytextsubscript{BERT-base} &  \best{87.48}\mytextsubscript{0.19} / \secondbest{81.96}\mytextsubscript{0.22}  & \best{81.26}\mytextsubscript{0.23} / \best{70.71}\mytextsubscript{0.28}   &  \secondbest{87.41}\mytextsubscript{0.23} / \best{71.20}\mytextsubscript{0.30} &   \mycite{hyilr} \\

TaxMorph\mytextsubscript{Sonnet-3.5} & --- / 80.2\textsuperscript{2)} & --- / --- & --- / --- & \mycite{taxmorph} \\

\bottomrule

\end{tabular}

\begin{tablenotes} 
\item \textsuperscript{1)} We report numbers directly obtained by the authors after some personal conversation.
\item \textsuperscript{2)} Use a 32.9k train, 7k val, and 7k test split.
\end{tablenotes} 
            
\end{table*}

For multi-label classification, we show the sample-based F1 scores in Table~\ref{tab:results_multi-label_classification}.
We argue that the sample-based F1 represents real-world applications where each document needs to be annotated one document after the other, such as in subject indexing in digital libraries~\cite{DBLP:conf/kcap/GalkeMSBS17,DBLP:conf/jcdl/MaiGS18}.  

\
The T5-base model is the strongest or second-strongest of the sequence-based models for multi-label datasets, except GoEmotions.
This differs from the results on the single-label datasets, where T5-base was behind encoder-only models, except for MR.

A WideMLP with TF-IDF is falling behind the sequence-based models. 
For some datasets, like DBpedia, the differences among the sequence-based methods are very small, while for others, a difference of up to three points can be observed. 

Many of the multi-label datasets are hierarchical, \ie have hierarchically organized labels.
We report the use of hierarchy-based methods for three multi-label classification datasets with hierarchy in the next section.
Furthermore, we report results involving \LLMs in the next section, too. 

\subsection{Hierarchical Text Classification}
Table~\ref{tab:results_multi-label_classification-extra} shows results for Micro-F1 and Macro-F1 metrics for the three hierarchical datasets: Web of Science (WOS)~\cite{DBLP:conf/icmla/KowsariBHMGB17}, NYT, and RCV1-V2 (see Section~\ref{sec:datasets}).

Sequence-based models perform better than the hierarchy-based methods, except for Micro-F1 on WOS. 
The BoW-based methods clearly fall behind, most notably on the more difficult-to-improve F1-Macro measure.

As with single-label and multi-label classification, TF-IDF with BoW is stronger than pure BoW.
In terms of the sequence-based methods, the best performer depends on the dataset and measure.
But all of them are based on an encoder-only model like HBGL or an encoder-decoder like Seq2Tree.
The \LLM-based Retrieval-style ICL uses a 16-shot in-context learning, clearly falling behind.
A fine-tuned and unmasked Llama-7B cannot reach the performance of smaller language models like encoder-only variants.
The difference is particularly notable on F1-Macro.

Regarding the hierarchy-based methods, the strongest method is HyILR.
It is followed by other hierarchy-based methods as close runners-up, such as HILL, and is barely outperformed in two cases by HALB.

\section{Discussion}\label{sec:findings}

\subsection{Key Findings} 

The results highlight that, overall, the strongest model variants are sequence-based language models. 
For single-label classification, encoder-only SLMs and encoder-decoder SLMs remain highly competitive to dedicated LLM pipelines.
Data augmentation and carefully designed ensembles using LLMs yield the strongest overall results, while graph-based approaches provide only dataset-specific advantages.
For multi-label classification, the encoder-decoder T5 model is the strongest approach (except for GoEmotions).
In hierarchical datasets, one observes benefits from hierarchy-aware modeling in sequence-based models, which obtain the highest scores of all model families.
Dedicated hierarchical methods using the given taxonomy generally fail to consistently outperform strong fine-tuned language models.
Below we discuss the results in detail along the three tasks and per model families.

\paragraph{Single-label Classification}
As shown in Table~\ref{tab:results_textclf_micro}, encoder-only \SLMs show strong results. 
The graph-based MHGAT is the best method on the 20ng dataset.
The \LLM-based 16-shot ensemble GPT-3.5+CARP+voting using a fine-tuned RoBERTa-Large is the best on the MR dataset, while RGPT with an ensemble of $k$=7 Llama 2 models is the strongest model on ohsumed.
Both \LLM-based models were not evaluated on 20ng. 
We were not able to independently verify the results of MHGAT 
due to the lack of source code and the authors' accessibility, we can only infer from the reported ablations that the hypergraph-based positional embedding may be a beneficial factor for the 20ng dataset~\cite{mhgat}.
Regarding GPT-3.5+CARP+voting, the use of an ensemble and supervised fine-tuned RoBERTa-Large to select the 16-shot provides the advantages as shown in the results of the different GPT-3.5 model variants in Table~\ref{tab:results_textclf_micro}.
For RGPT, it requires an ensemble of Llama 2 models to outperform the second-best model, a pure DeBERTa-base model, by $1.47$ points.

Regarding the encoder-only models, BERT-large does not improve more than 1 point over BERT-base, despite having three times more parameters.
On 20ng, the performance even drops by one point.
Presumably, the capacity of BERT-base is already sufficient to tackle the single-label classification tasks, especially for the R8 and R52 datasets, and there are reports that BERT-large is harder to fine-tune on small datasets~\cite{DBLP:conf/naacl/DevlinCLT19}.

Augmentation improves the results and makes the \SLM AM-RoBERTa model the overall strongest single-label classifier.
A similar observation is made for the BoW methods, where the use of concept-based TF-ICF features with SVM in CFE-IterativeAdditive~\cite{ATTIEH2023110215} improves the results in three datasets. 
Such an observation was also made by Galke~\etal\cite{DBLP:conf/kcap/GalkeMSBS17}, who used CTF-IDF features for text classification.

\paragraph{Multi-label Classification}

The strongest multi-label classifier is T5-base, outperforming the encoder-only RoBERTa and DeBERTa, in contrast to the general observation for the single-label datasets.
An interesting observation can be made on the multi-label sentiment prediction dataset GoEmotions, where \mlp achieves the worst performance across all models and the highest margin compared to BERT regarding all datasets. 
Similarly, BERT outperforms \mlp on the single-label MR dataset with the largest margin compared to other datasets. 
However, \mlp is a surprisingly strong model on the EconBiz dataset with the most number of labels. 
This may be due to the highly imbalanced (long tail) label distribution of the thousands of labels in the EconBiz dataset~\cite{DBLP:conf/jcdl/MaiGS18}, which may be easier to reflect in a model trained from scratch than in a pre-trained model, such as BERT.

We assume that BoW-based MLP models are at a disadvantage in sentiment prediction compared to sequence-based models, where word order is relevant for the task.
Note that most graph-based methods also discard word order when setting up the graph~\cite{galkescherp-acl2022}, except for models that combine the GNN with a sequence-based transformer model.

Since there are no sample-based F1 results in the literature for hierarchical multi-label datasets, we conducted an additional analysis (not shown in the table) and ran the best-performing HiAGM-TP variant with GCN as graph encoder~\cite{hiagm} and the sequence-based HBGL~\cite{hbgl} on the  RCV1-V2 and NYT datasets.
We use the optimized hyperparameters from the papers.
The results are, HiAGM with a graph-encoder GCN obtained strong results (RCV1-V2: 85.51, Amazon: 89.05, DBPedia: 97.17, NYT: 76.57) but is outperformed by the best encoder-only models.
It ran out of memory on EconBiz on a 40 GB RAM NVIDIA A100 HGX GPU.
HBGL with BERT-base achieves on RCV1-V2 a sample-based F1 score of $88.76$ 
and on NYT a score of $82.01$. 
The former is outperformed by a T5-base, as shown in Table~\ref{tab:results_multi-label_classification}, while the latter surpasses all encoder-only and encoder-decoder models considered, demonstrating that hierarchy-aware methods can be beneficial for hierarchical datasets.
Note that HBGL and HiAGM cannot be applied to the flat multi-label datasets R21578 and GoEmotions as they do not have a label hierarchy.

Existing studies using \LLMs focus on single-label datasets only, \eg \cite{pushing-the-limit-11346014-rgpt,DBLP:conf/emnlp/PecherSB25-100-samples}.
But we were able to include \LLMs for the hierarchical multi-label datasets, which we discuss below.

\paragraph{Hierarchical Classification}

For hierarchical multi-label text classification, we conclude that sequence-based methods are best.
Except for a slim margin of $0.12$ for the hierarchy-based HyILR method regarding Micro-F1 on WOS, the best-performing models are sequence-based methods.
Which sequence-based method performs best depends on the dataset and metric, but generally they are methods using fine-tuned \SLMs. 
The hierarchy-based methods incorporating graph neural networks, \eg to use a GNN to encode the class hierarchy as in HiAGM, fail to outperform the best of our tested language models.

A fine-tuned \LLM unLlama with 7B parameters cannot outperform the \SLMs. 
However, it should be noted that \LLM-based methods with in-context learning are effective with small amounts of labeled data, which we further discuss below.

\subsection{Fine-tuning \SLMs vs. In-Context Learning in \LLMs}

An important discussion regarding model performance is the distinction between fine-tuning and in-context learning. 
When a sufficiently large training dataset is available, it is reported for single-label datasets that \SLMs outperform the use of \LLMs~\cite{DBLP:conf/emnlp/PecherSB25-100-samples}.
While SLMs \emph{need} fine-tuning to obtain their good results, LLMs can also do in-context learning with few examples~\cite{DBLP:conf/emnlp/PecherSB25-100-samples}, which gives them an advantage when little training data is available.  
The \LLM-based method  CARP~\cite{carp}, \ie the variant of GPT-3.5+CARP+vote (16-shot, RoBERTa sampler) requires a fine-tuned \SLM (the RoBERTa model) for sampling the 16 most representative examples per class.
This implies the need for contextualized embeddings from a fine-tuned encoder-only language model prior to prompting the decoder-only language model. 
While analyzing the importance of example selection is of scientific interest, practitioners should take into account that one could use the fine-tuned SLM model directly for the classification task.
One would get results close to the fine-tuned SLM, while avoiding running an \LLM at inference time.

The best-performing method on the single-label ohsumed dataset, RGPT, uses an ensemble of $k$=7 Llama 2 models, each of which is fine-tuned and applied sequentially.  
Each model has about 2.1~M effective trainable parameters (with LoRA rank = 8)~\cite{pushing-the-limit-11346014-rgpt}, which shows that fine-tuning LLMs can be promising. 
Recent work supports the notion that fine-tuned models are generally advantageous over few-shot prompting when a minimum of labeled data is available~\cite{edwards2024language,bucher2024finetunedsmallllmsstill,lepagnol2024smalllanguagemodelsgood,yehudai2024fastfit}.
Pecher et al.~\cite{DBLP:conf/emnlp/PecherSB25-100-samples} find for single-label datasets that in some cases already 100 labeled examples per class are sufficient for fine-tuned models to outperform in-context learning. However, the case for multi-label must be treated with more nuance as the sample complexity grows when a single example may be annotated with more than one label.

A key trick for effectively fine-tuning a \LLM directly for a classification task may be to remove the constraint of the left-to-right attention mask~\cite{li2023label}.
However, experiments on fine-tuning unLlama with 7B parameters and LoRA on hierarchical text classification datasets~\cite{hydra} do not confirm this assumption.
Rather, fine-tuning an \SLM like the BERT-base model already outperforms it.

Regarding reasoning capabilities in \LLMs, Guo, Zhang, and Qin~\cite{guo2026textreasoningbenchdoesreasoningreally} conclude that reasoning generally does not improve classification performance. 
Simpler strategies like Chain of Thought (CoT) yield consistent but limited gains of 1 to 3 percentage points on large models. 
More complex reasoning approaches often fail to outperform simpler baselines, and adding reasoning to small models may even degrade performance~\cite{guo2026textreasoningbenchdoesreasoningreally}. 
This comes at additional costs of using reasoning strategies, which consume up to two orders of magnitude more tokens for only marginal performance improvements. 
It seems that while sophisticated prompting strategies can be to some extent beneficial, longer reasoning does not necessarily lead to better classification performance and may even degrade performance. 

\subsection{Synthetic Text-graphs Hardly Bring an Advantage}

Methods that generate a synthetic text graph and apply graph-based neural networks (GNNs) for text classification incur high training costs.
First, the graph has to be computed.
Second, a GNN has to be trained. 
For standard GNN methods, the whole graph has to fit into the GPU memory, and mini-batching is non-trivial but possible with dedicated sampling techniques for GNNs~\cite{DBLP:conf/icml/FeyLWL21}. 
Computing the word-document graphs requires $\mathcal{O}(N^2)$ space, where $N$ is the number of documents plus the vocabulary size, which is a hurdle for large-scale applications.

Our experiments show that simple BoW-based models like WideMLP and SVM already outperform graph-based models like TextGCN, HeteGCN, and Induct-GCN in the inductive text classification setting.
An exception is 20ng, where MHGAT performs better.
An explanation may be that transforming a text corpus into a graph cannot add any new information according to the data processing inequality~\cite{cover_elements_1991}.

The seminal paper on graph convolution by Kipf \& Welling~\cite{DBLP:conf/iclr/KipfW17} argued that graph neural networks are most effective when the edges provide additional information that cannot be modeled otherwise.
Therefore, it is important to distinguish between text-induced graphs for text classification, which seem to provide little to no gain, and tasks where the \emph{natural} structure of the graph data provides more information than the mere text, \eg citation networks. 
When extra information is encoded in the graph, graph neural networks are the state of the art~\cite{DBLP:conf/iclr/KipfW17,velickovic2018graph} and superior to MLPs that use only the node features and not the graph structure~\cite{DBLP:journals/corr/abs-1811-05868}.

Our survey suggests that a graph induced from pure text generally
does \emph{not} improve text classification. 
Recently, Bugue\~{n}o and de Melo~\cite{buguenoConnectingDotsWhat2023} compared different document representations (Word2vec, GloVe, and frozen BERT) for graph neural networks.
Using different datasets, they confirm that on most datasets, graph neural networks did \emph{not} outperform a fine-tuned BERT regardless of the choice of input representation. 
The finding was also confirmed in a study on short text classification~\cite{DBLP:conf/cdmake/KarlS23},
where several graph-based methods have been compared to \SLMs on six benchmark datasets (including R8 and MR) and four new datasets.

\subsection{Using a Graph Encoder for the Hierarchy Hardly Brings an Advantage for Hierarchical Text Classification}

In hierarchical text classification, we make similar observations as with single-label classification. 
Encoder-only models like DeBERTa and RoBERTa, the HBGL method (which incorporates the hierarchy into a standard BERT model), RADAr (which uses an autoregressive decoder instead of a classifier head), and HYDRA (which represents the hierarchy in level-wise losses), are overall best-performing depending on the dataset and metric. 
HiAGM uses a GNN to encode the class hierarchy but fails to outperform the hierarchy-agnostic sequence-based DeBERTa model.
More modern models like HALB and HILL outperform DeBERTa-base, but fall behind other encoder-only models like HYDRA.

Ablations by
Wang~\etal\cite{DBLP:conf/acl/WangWH0W22} have shown that removing the graph encoder reduces the performance only by about 1 point
(Micro/Macro-F1)~\cite{DBLP:conf/acl/WangWH0W22}.  
Similarly, RADAr~\cite{radar} and HYDRA~\cite{hydra} also provide empirical results that an explicit graph encoder is not needed for hierarchical text classification.

\subsection{BERT Baselines are Often Undertuned}

We found that BERT baselines are often undertuned in the literature. This can be declared as ``baseline nerfing'' and may be accidental~\cite{leech2024questionablepracticesmachinelearning}. 
We hope that our comparison sheds new light on the various proposed methods with solid BERT baselines. What we argue to be particularly problematic is omitting encoder-only baselines as soon as some prior work has a marginal advantage over BERT, as this practice prohibits readers from properly contextualizing the results, \eg when the new method is also only marginally better than BERT. Based on the results of our quantitative comparison, we argue that simple baselines such as BERT, an MLP, or an SVM should not be omitted in text classification.

\section{Limitations}\label{sec:limitations}

A limitation of our comparison may be the selection of datasets on which we base this comparative survey. 
Although the selection of specific datasets could potentially bias the comparison, we chose the most common datasets for maximum coverage of the methods.
The current dataset collection was made to ensure broad coverage of approaches for single-label text classification, multi-label text classification, and hierarchical text classification. 
But we observe that the choice of multi-label datasets in the literature is more scattered than in the single-label case, which harms comparability. 
In order to balance this, we have included the most prominent multi-label datasets that provide a hierarchy, namely, WOS, NYT, and RCV1-V2.
We also include datasets that go beyond the news domain, such as EconBiz, DBPedia, and even non-topical classification tasks, such as GoEmotions. 
For comparability, we have reported three variants of the F measure, namely sample-based, micro, and macro F1.
Generally, we expect similar observations to be made on other text classification datasets because we already covered a wide range of text classification settings from long, medium, to short texts, topic and sentiment classification, and from single-label, multi-label, to hierarchical classification in the domains of forum postings, news, movie reviews, scholarly articles, and product reviews. 

We further emphasize that the experimental datasets are limited to English.
While word order is important in the English language, 
it is notable that methods that discard word order still work very well for topical text classification.
We assume that BoW-based models perform even better for languages with a richer morphology, where word order is less important.
It would be interesting to see to what extent our results of comparing BoW-based vs. sequence-based vs. graph-based vs. hierarchy-based methods for text classification transfer to other languages.
Towards this direction, Gonz\'{a}lez-Carvajal and Garrido-Merch\'{a}n~\cite{gonzales-2020-comparing-bert} show that BERT outperforms classic TF-IDF BoW approaches on English, Chinese, and Portuguese text classification datasets.
But other methods are yet to be considered.
Another direction is to consider specifically designed text classifiers for a single language, \eg in Chinese~\cite{9837023-chinese} where methods are tailored to the characteristics of Chinese characters, words, and radical information.
Furthermore, there is room for an even larger coverage of datasets in future work~\cite{Bhatia16}. 

Lastly, a major problem with evaluating \GenLMs is that benchmark datasets may have leaked into their training data~\cite{balloccuLeakCheatRepeat2024}.
It is not (always) clear if certain test sets are included in the language models' pre-training data.

\section{Future Directions}\label{sec:future}

As argued in the previous section, \SLMs benefit from fine-tuning on a training set, while \GenLMs can do in-context learning with few examples. 
However, a large language model like Llama can also be fine-tuned for text classification~\cite{li2023label}.
This strengthens the view that the main distinction is whether the model is fine-tuned or not and hints at fine-tuning of \GenLMs as a direction of future work, \eg via group relative policy optimization~\cite{shao2024deepseekmath}.

A different direction is to increase the model size of masked language models. That is, train larger versions of BERT on modern dataset collections. However, masked language models have lower ``throughput'' than causal language models. This is because causal language models obtain a training signal from every token, whereas masked language models only receive a training signal from masked tokens. Still, encoding left and right context has been shown to be important for text classification performance~\cite{li2023label}.
It remains unclear which strategy is better: training a large masked language model or unmasking a large causal language model during fine-tuning.

Existing studies on using \LLMs are mostly considering single-label datasets, \eg~\cite{pushing-the-limit-11346014-rgpt,carp,DBLP:conf/emnlp/PecherSB25-100-samples}. 
Expanding large language models for multi-label and hierarchical text classification raises interesting questions: how to feed the multi-label hierarchy into the input context of a language model, \eg as in \cite{fatemi2024talk}.
Existing approaches so far underperform, such as the 16-shot Retrieval-style ICL~\cite{DBLP:journals/corr/abs-2406-17534-retrieval-style} or TaxMorph~\cite{taxmorph}, of which the latter even modifies the provided taxonomy for the \LLM.
Thus, an open question is how to effectively use \LLMs for multi-label classification and provide the hierarchy as input.

Another challenging setting is few-shot classification as in prompt-based large language models~\cite{DBLP:conf/nips/BrownMRSKDNSSAA20}.
Here, generative \LLMs can exploit their advantage by accessing large numbers of parameters trained on large amounts of data.
As a study on single-label datasets showed~\cite{DBLP:conf/emnlp/PecherSB25-100-samples}, sufficient training data is needed for \SLMs to outperform \LLMs.
Thus, an interesting direction may be to investigate approaches that allow \SLMs to perform better in a few-shot regime, \eg end-task-aware pre-training vs. fine-tuning after pre-training~\cite{dery2021should}.

\section{Conclusion}
The introduction of sequence-based transformer models, starting with the seminal BERT paper by Devlin \etal~\cite{DBLP:conf/naacl/DevlinCLT19} in late 2018, has, as with other NLP tasks, strongly advanced the field. 
From our survey, we see that among the single-label, multi-label, and hierarchical text classification tasks, the top performers are fine-tuned language models.
We argue that discriminative \SLMs and generative \LLMs have individual strengths, influenced by factors such as the availability of sufficient training data.

Reflecting back on our three central questions from Section~\ref{sec:questions}, we see that beyond the introduction of transformer models, today's methods operating on graphs generally provide no benefit over pre-trained language models.
This is regardless of whether the graph is synthetically induced from text or if a hierarchy is provided with a multi-label dataset.
Even worse, many new approaches fail to outperform straightforward baselines, such as an SVM or a multilayer perceptron.

Moreover, despite the astounding performance of \LLMs in few-shot prompting, the best or close-to-best performance in single-label classification can be achieved through fine-tuning.
This requires that sufficient training data is available~\cite{DBLP:conf/emnlp/PecherSB25-100-samples}.
The amount of training data needed depends on the task. 
It shows that despite advanced techniques for \LLMs, like prompting and selecting in-context examples, text classification with fine-tuning \SLMs is competitive.
We argue that research in text classification should employ a pre-trained, encoder-only \SLM such as a fine-tuned BERT as baseline. 

Regarding the question of comparability, we found that, particularly for the strong baselines BERT and RoBERTa, the reported numbers vary across the literature.
This is despite using benchmark datasets and apparently identical experimental settings.
While it is not clear in all cases what caused these differences, our experiments have shown that already small changes in the learning rate can have a comparatively large effect on downstream performance.
Thus, one has to carefully study a new method and check if experimental setups are comparable in order to derive conclusions about the advancements claimed.

Our study immediately impacts practitioners seeking to employ robust text classification models in research projects and industrial operational environments. 
Our recommendation to practitioners is to use a pre-trained \SLM when sufficient training data are available.
If this condition is not met, a frozen \LLM with zero-shot or in-context classification may be used via few-shot learning. 
The trade-off lies between computational cost for fine-tuning vs\@. continued token cost for in-context learning.

\subsubsection*{Acknowledgments}
We thank Yousef Younes and Gregor Donabauer for discussions and experiments on hierarchical and graph-based models.
This work was co-funded by the DFG as part of the SmartER project (No. 515537520). L.G.P.\ acknowledges support from the Novo Nordisk Foundation under grant reference number NNF25OC0103204.

\bibliographystyle{IEEEtran}
\bibliography{main}

@inproceedings{DBLP:conf/icmla/KowsariBHMGB17,
  author       = {Kamran Kowsari and
                  Donald E. Brown and
                  Mojtaba Heidarysafa and
                  Kiana Jafari Meimandi and
                  Matthew S. Gerber and
                  Laura E. Barnes},
  no_editor       = {Xue{-}wen Chen and
                  Bo Luo and
                  Feng Luo and
                  Vasile Palade and
                  M. Arif Wani},
  title        = {{HDLTex}: Hierarchical Deep Learning for Text Classification},
  booktitle    = {ICMLA},
  no_pages        = {364--371},
  publisher    = {{IEEE}},
  year         = {2017},
  no_url          = {https://doi.org/10.1109/ICMLA.2017.0-134},
  doi          = {10.1109/ICMLA.2017.0-134},
  bibsource    = {dblp computer science bibliography, https://dblp.org}
}

@article{llama3,
  author       = {Llama Team},
  title        = {The {Llama} 3 Herd of Models},
  journal      = {CoRR},
  volume       = {abs/2407.21783},
  year         = {2024},
  url          = {https://doi.org/10.48550/arXiv.2407.21783},
  doi          = {10.48550/ARXIV.2407.21783},
  eprinttype   = {arXiv},
  eprint       = {2407.21783},
  bibsource    = {dblp computer science bibliography, https://dblp.org}
}

@inproceedings{hyilr,
  author       = {Ashish Kumar and
                  Durga Toshniwal},
  no_editor       = {Jin Zhao and
                  Mingyang Wang and
                  Zhu Liu},
  title        = {{HyILR}: Hyperbolic Instance-Specific Local Relationships for Hierarchical
                  Text Classification},
  booktitle    = {Student Research Workshop at ACL},
  no_pages        = {872--883},
  publisher    = {ACL},
  year         = {2025},
  no_url          = {https://doi.org/10.18653/v1/2025.acl-srw.63},
  doi          = {10.18653/V1/2025.ACL-SRW.63},
  bibsource    = {dblp computer science bibliography, https://dblp.org}
}

@misc{guo2026textreasoningbenchdoesreasoningreally,
      title={{TextReasoningBench}: Does Reasoning Really Improve Text Classification in Large Language Models?}, 
      author={Xinyu Guo and Yazhou Zhang and Jing Qin},
      year={2026},
      eprint={2603.19558},
      archivePrefix={arXiv},
      primaryClass={cs.CL},
      url={https://arxiv.org/abs/2603.19558}, 
}

@inproceedings{hlpd,
  author       = {Fatos Torba and
                  Christophe Gravier and
                  Charlotte Laclau and
                  Abderrhammen Kammoun and
                  Julien Subercaze},
  no_editor       = {Claudia Hauff and
                  Craig Macdonald and
                  Dietmar Jannach and
                  Gabriella Kazai and
                  Franco Maria Nardini and
                  Fabio Pinelli and
                  Fabrizio Silvestri and
                  Nicola Tonellotto},
  title        = {Decoding the Hierarchy: {A} Hybrid Approach to Hierarchical Multi-label
                  Text Classification},
  booktitle    = {ECIR},
  no_pages        = {405--420},
  publisher    = {Springer},
  year         = {2025},
  no_url          = {https://doi.org/10.1007/978-3-031-88708-6\_26},
  doi          = {10.1007/978-3-031-88708-6\_26},
  bibsource    = {dblp computer science bibliography, https://dblp.org}
}

@inproceedings{lh-mix,
  author       = {Fanshuang Kong and
                  Richong Zhang and
                  Ziqiao Wang},
  no_editor       = {Yizhou Sun and
                  Flavio Chierichetti and
                  Hady W. Lauw and
                  Claudia Perlich and
                  Wee Hyong Tok and
                  Andrew Tomkins},
  title        = {{LH-Mix}: Local Hierarchy Correlation Guided Mixup over Hierarchical
                  Prompt Tuning},
  booktitle    = {SIGKDD},
  no_pages        = {636--646},
  no_publisher    = {{ACM}},
  year         = {2025},
  no_url          = {https://doi.org/10.1145/3690624.3709326},
  doi          = {10.1145/3690624.3709326},
  bibsource    = {dblp computer science bibliography, https://dblp.org}
}

@inproceedings{PAAM-HiA-T5,
  author       = {Wei Huang and
                  Chen Liu and
                  Bo Xiao and
                  Yihua Zhao and
                  Zhaoming Pan and
                  Zhimin Zhang and
                  Xinyun Yang and
                  Guiquan Liu},
  no_editor       = {Nicoletta Calzolari and
                  Chu{-}Ren Huang and
                  Hansaem Kim and
                  James Pustejovsky and
                  Leo Wanner and
                  Key{-}Sun Choi and
                  Pum{-}Mo Ryu and
                  Hsin{-}Hsi Chen and
                  Lucia Donatelli and
                  Heng Ji and
                  Sadao Kurohashi and
                  Patrizia Paggio and
                  Nianwen Xue and
                  Seokhwan Kim and
                  Younggyun Hahm and
                  Zhong He and
                  Tony Kyungil Lee and
                  Enrico Santus and
                  Francis Bond and
                  Seung{-}Hoon Na},
  title        = {Exploring Label Hierarchy in a Generative Way for Hierarchical Text
                  Classification},
  booktitle    = {COLING},
  no_pages        = {1116--1127},
  publisher    = {Computational Linguistics},
  year         = {2022},
  no_url          = {https://aclanthology.org/2022.coling-1.95},
  bibsource    = {dblp computer science bibliography, https://dblp.org}
}

@inproceedings{taxmorph,
  author       = {Jonas Golde and
                  Nicolaas Paul Jedema and
                  RaviKiran Krishnan and
                  Phong Le},
  no_editor       = {Vera Demberg and
                  Kentaro Inui and
                  Llu{\'{\i}}s Marquez},
  title        = {Hierarchical Text Classification with {LLM}-Refined Taxonomies},
  booktitle    = {EACL},
  no_pages        = {214--228},
  publisher    = {ACL},
  year         = {2026},
  no_url          = {https://aclanthology.org/2026.eacl-long.10/},
  bibsource    = {dblp computer science bibliography, https://dblp.org}
}

@inproceedings{hydra,
  author       = {Fabian Karl and
                  Ansgar Scherp},
  no_editor       = {Christos Christodoulopoulos and
                  Tanmoy Chakraborty and
                  Carolyn Rose and
                  Violet Peng},
  title        = {{HYDRA:} {A} Multi-Head Encoder-only Architecture for Hierarchical
                  Text Classification},
  booktitle    = {EMNLP},
  no_pages        = {9292--9303},
  publisher    = {ACL},
  year         = {2025},
  no_url          = {https://doi.org/10.18653/v1/2025.emnlp-main.472},
  doi          = {10.18653/V1/2025.EMNLP-MAIN.472},
  bibsource    = {dblp computer science bibliography, https://dblp.org}
}

@article{sharma2025readreinforcementbasedadversariallearning,
      title={READ: Reinforcement-based Adversarial Learning for Text Classification with Limited Labeled Data}, 
      author={Rohit Sharma and Shanu Kumar and Avinash Kumar},
    journal={arXiv:2501.08035},
      year={2025},
}

@inproceedings{DBLP:conf/iclr/HeGC23,
  author       = {Pengcheng He and
                  Jianfeng Gao and
                  Weizhu Chen},
  title        = {{DeBERTaV3}: Improving DeBERTa using ELECTRA-Style Pre-Training with
                  Gradient-Disentangled Embedding Sharing},
  booktitle    = {ICLR},
  publisher    = {OpenReview.net},
  year         = {2023},no_url          = {https://openreview.net/forum?id=sE7-XhLxHA},
  bibsource    = {dblp computer science bibliography, https://dblp.org}
}

@inproceedings{DBLP:conf/iclr/ClarkLLM20-electra,
  author       = {Kevin Clark and
                  Minh{-}Thang Luong and
                  Quoc V. Le and
                  Christopher D. Manning},
  title        = {{ELECTRA:} Pre-training Text Encoders as Discriminators Rather Than
                  Generators},
  booktitle    = {ICLR},
  publisher    = {OpenReview.net},
  year         = {2020},no_url          = {https://openreview.net/forum?id=r1xMH1BtvB},
  bibsource    = {dblp computer science bibliography, https://dblp.org}
}

@article{warner2024smarterbetterfasterlonger,
      title={Smarter, Better, Faster, Longer: A Modern Bidirectional Encoder for Fast, Memory Efficient, and Long Context Finetuning and Inference}, 
      author={Benjamin Warner and Antoine Chaffin and Benjamin Clavié and Orion Weller and Oskar Hallström and Said Taghadouini and Alexis Gallagher and Raja Biswas and others},
      year={2024},
      journal={arXiv:2412.13663},
}

@article{donabauer2024tokenlevelgraphsshorttext,
      title={Token-Level Graphs for Short Text Classification}, 
      author={Gregor Donabauer and Udo Kruschwitz},
      year={2024},
      journal={arXiv:2412.12754},
}

@article{peng2025text,
  author    = {Peng, Qiaojuan and Luo, Xiong and Yuan, Yuqi and Gu, Fengbo and Shen, Hailun and Huang, Ziyang},
  title     = {A text classification method combining in-domain pre-training and prompt learning for the steel e-commerce industry},
  journal   = {Int. Journal of Web Information Systems},
  year      = {2025},
  volume    = {21},
  number    = {1},
 no_pages =  {96--119},
  doi       = {10.1108/IJWIS-09-2024-0277},no_url       = {https://doi.org/10.1108/IJWIS-09-2024-0277},
  publisher = {Emerald Publishing Limited}
}

@article{petridis2024textclassificationneuralnetworks,
      title={Text Classification: Neural Networks VS Machine Learning Models VS Pre-trained Models}, 
      author={Christos Petridis},
      year={2024},
      journal={arXiv:2412.21022},
}

@inproceedings{mao2024lowresourcefasttextclassification,
  author       = {Yanxu Mao and
                  Peipei Liu and
                  Tiehan Cui and
                  Congying Liu and
                  Datao You},
  title        = {Low-Resource Fast Text Classification Based on Intra-Class and Inter-Class
                  Distance Calculation},
  booktitle    = {{COLING}},
  no_pages        = {1045--1056},
  no_publisher    = {Association for Computational Linguistics},
  year         = {2025}
}

@article{DBLP:journals/npl/ZhangDLZ25-hgbl,
  author       = {Chaoqun Zhang and
                  Linlin Dai and
                  Chengxing Liu and
                  Longhao Zhang},
  title        = {{HGBL:} {A} Fine Granular Hierarchical Multi-Label Text Classification
                  Model},
  journal      = {Neural Process. Lett.},
  volume       = {57},
  number       = {1},
 no_pages =  {1},
  year         = {2025},no_url          = {https://doi.org/10.1007/s11063-024-11713-x},
  doi          = {10.1007/S11063-024-11713-X},
  bibsource    = {dblp computer science bibliography, https://dblp.org}
}

@inproceedings{DBLP:conf/acl/LiuZHWZ0C23-k-htc,
  author       = {Ye Liu and
                  Kai Zhang and
                  Zhenya Huang and
                  Kehang Wang and
                  Yanghai Zhang and
                  Qi Liu and
                  Enhong Chen},
  no_editor       = {Anna Rogers and
                  Jordan L. Boyd{-}Graber and
                  Naoaki Okazaki},
  title        = {Enhancing Hierarchical Text Classification through Knowledge Graph
                  Integration},
  booktitle    = {ACL Findings},
 no_pages =  {5797--5810},
  no_publisher    = {ACL},
  year         = {2023},no_url          = {https://doi.org/10.18653/v1/2023.findings-acl.358},
  doi          = {10.18653/V1/2023.FINDINGS-ACL.358},
  bibsource    = {dblp computer science bibliography, https://dblp.org}
}

@article{DBLP:journals/nca/JiaJDZLXC23-mhgat,
  author       = {Xiangen Jia and
                  Min Jiang and
                  Yihong Dong and
                  Feng Zhu and
                  Haocai Lin and
                  Yu Xin and
                  Huahui Chen},
  title        = {Multimodal heterogeneous graph attention network},
  journal      = {Neural Comput. Appl.},
  volume       = {35},
  number       = {4},
 no_pages =  {3357--3372},
  year         = {2023},no_url          = {https://doi.org/10.1007/s00521-022-07862-6},
  doi          = {10.1007/S00521-022-07862-6},
  bibsource    = {dblp computer science bibliography, https://dblp.org}
}

@inproceedings{DBLP:conf/acl/QoribMN24,
  author       = {Muhammad Reza Qorib and
                  Geonsik Moon and
                  Hwee Tou Ng},
  no_editor       = {Lun{-}Wei Ku and
                  Andre Martins and
                  Vivek Srikumar},
  title        = {Are Decoder-Only Language Models Better than Encoder-Only Language
                  Models in Understanding Word Meaning?},
  booktitle    = {ACL Findings},
 no_pages =  {16339--16347},
  no_publisher    = {ACL},
  year         = {2024},no_url          = {https://doi.org/10.18653/v1/2024.findings-acl.967},
  doi          = {10.18653/V1/2024.FINDINGS-ACL.967},
  bibsource    = {dblp computer science bibliography, https://dblp.org}
}

@article{DBLP:journals/kbs/ZhangLSXTH24-halb,
  author       = {Jun Zhang and
                  Yubin Li and
                  Fanfan Shen and
                  Chenxi Xia and
                  Hai Tan and
                  Yanxiang He},
  title        = {Hierarchy-Aware and Label Balanced Model for Hierarchical Text Classification},
  journal      = {Knowl. Based Syst.},
  volume       = {300},
 no_pages =  {112153},
  year         = {2024},no_url          = {https://doi.org/10.1016/j.knosys.2024.112153},
  doi          = {10.1016/J.KNOSYS.2024.112153},
  bibsource    = {dblp computer science bibliography, https://dblp.org}
}

@INPROCEEDINGS{10748883-boing-aviation-paper,
  author={Niraula, Nobal and Ayhan, Samet and Chidambaram, Balaguruna and Whyatt, Daniel},
  booktitle={2024 AIAA DATC/IEEE 43rd Digital Avionics Systems Conf. (DASC)}, 
  title={Multi-Label Classification with Generative Large Language Models}, 
  year={2024},
  volume={},
  number={},
 no_pages = {1-7},
  doi={10.1109/DASC62030.2024.10748883}}

@article{DBLP:journals/eswa/ReusensSTSVBB24,
  author       = {Manon Reusens and
                  Alexander Stevens and
                  Jonathan Tonglet and
                  Johannes De Smedt and
                  Wouter Verbeke and
                  Seppe vanden Broucke and
                  Bart Baesens},
  title        = {Evaluating text classification: {A} benchmark study},
  journal      = {Expert Syst. Appl.},
  volume       = {254},
 no_pages =  {124302},
  year         = {2024},no_url          = {https://doi.org/10.1016/j.eswa.2024.124302},
  doi          = {10.1016/J.ESWA.2024.124302},
  bibsource    = {dblp computer science bibliography, https://dblp.org}
}

@inproceedings{sgm,
  author       = {Pengcheng Yang and
                  Xu Sun and
                  Wei Li and
                  Shuming Ma and
                  Wei Wu and
                  Houfeng Wang},
  no_editor       = {Emily M. Bender and
                  Leon Derczynski and
                  Pierre Isabelle},
  title        = {{SGM:} Sequence Generation Model for Multi-label Classification},
  booktitle    = {COLING},
 no_pages =  {3915--3926},
  no_publisher    = {ACL},
  year         = {2018},no_url          = {https://aclanthology.org/C18-1330/},
  bibsource    = {dblp computer science bibliography, https://dblp.org}
}

@inproceedings{seq2tree,
  author       = {Chao Yu and
                  Yi Shen and
                  Yue Mao},
  no_editor       = {Enrique Amig{\'{o}} and
                  Pablo Castells and
                  Julio Gonzalo and
                  Ben Carterette and
                  J. Shane Culpepper and
                  Gabriella Kazai},
  title        = {Constrained Sequence-to-Tree Generation for Hierarchical Text Classification},
  booktitle    = {{SIGIR}},
 no_pages =  {1865--1869},
  no_publisher    = {{ACM}},
  year         = {2022},no_url          = {https://doi.org/10.1145/3477495.3531765},
  doi          = {10.1145/3477495.3531765},
  bibsource    = {dblp computer science bibliography, https://dblp.org}
}

@article{mhgat,
  author       = {Yilun Jin and
                  Wei Yin and
                  Haoseng Wang and
                  Fang He},
  title        = {Capturing word positions does help: {A} multi-element hypergraph gated
                  attention network for document classification},
  journal      = {Expert Syst. Appl.},
  volume       = {251},
 no_pages =  {124002},
  year         = {2024},no_url          = {https://doi.org/10.1016/j.eswa.2024.124002},
  doi          = {10.1016/J.ESWA.2024.124002},
  bibsource    = {dblp computer science bibliography, https://dblp.org}
}

@ARTICLE{LMTCSG-10705790,
  author={Sun, Guoying and Li, Jie and Cheng, Yanan and Zhang, Zhaoxin},
  journal={IEEE Trans. on Industrial Informatics}, 
  title={LMTCSG: Multilabel Text Classification Combining Sequence-Based and GNN-Based Features}, 
  year={2024},
  volume={},
  number={},
 no_pages = {1-9},
  doi={10.1109/TII.2024.3465596}
}

@article{AGBESI2024e38515,
title = {MuTCELM: An optimal multi-TextCNN-based ensemble learning for text classification},
journal = {Heliyon},
volume = {10},
number = {19},no_pages =  {e38515},
year = {2024},
issn = {2405-8440},
doi = {https://doi.org/10.1016/j.heliyon.2024.e38515},no_url = {https://www.sciencedirect.com/science/article/pii/S240584402414546X},
author = {Victor Kwaku Agbesi and Wenyu Chen and Sophyani Banaamwini Yussif and Chiagoziem C. Ukwuoma and Yeong Hyeon Gu and Mugahed A. Al-antari}}

@InProceedings{10.1007/978-3-031-72350-6_20,
author="Xu, Xiantao
and Hu, Minghao
and Wang, Yongjie
and Luo, Wei
and Liu, Shilong
and Luo, ZhunChen
and Tan, Yushan",
no_editor="Wand, Michael
and Malinovsk{\'a}, Krist{\'i}na
and Schmidhuber, J{\"u}rgen
and Tetko, Igor V.",
title="PLIClass: Weakly Supervised Text Classification with Iterative Training and Denoisy Inference",
booktitle="ICANN 2024",
year="2024",
publisher="Springer",
no_address="Cham",no_pages = "292--305",
isbn="978-3-031-72350-6"
}

@article{DBLP:journals/corr/abs-2408-15650,
  author       = {Lingyu Gao},
  title        = {Harnessing the Intrinsic Knowledge of Pretrained Language Models for
                  Challenging Text Classification Settings},
  journal      = {arXiv:2408.15650},
  year         = {2024}
}

@article{Wei_Sun_2024, title={Study on text classification model combining BERT and convolutional neural network}, volume={2},no_url={https://drpress.org/ojs/index.php/mmaa/article/view/25642}, DOI={10.54097/7h5bs772}, abstractNote={
Text classification has been widely used in many practical scenarios such as sentiment analysis and topic classification, and is an important task in the field of natural language processing. In recent years, research on pre-training model based on Bidirectional Encoder Representations from Transformers (BERT) has made remarkable progress. However, at present, most studies are based on open source databases. In this study, a text classification model combining BERT model and convolutional neural network (CNN) is proposed to complete the text emotion classification task through the self-built dialect text database. This paper demonstrates the effectiveness and superiority of this model in Chinese text classification tasks by comparing experiments and performance evaluation.
}, number={3}, journal={Mathematical Modeling and Algorithm Application}, author={Wei , Jia and Sun, Xiangguo}, year={2024}, no_month={Sep.},no_pages = {10–12} }

@inproceedings{DBLP:conf/kdd/Yu00S24,
  author       = {Linzhu Yu and
                  Huan Li and
                  Ke Chen and
                  Lidan Shou},
  no_editor       = {Ricardo Baeza{-}Yates and
                  Francesco Bonchi},
  title        = {BoKA: Bayesian Optimization based Knowledge Amalgamation for Multi-unknown-domain
                  Text Classification},
  booktitle    = {SIGKDD},
 no_pages =  {4035--4046},
  publisher    = {{ACM}},
  year         = {2024},no_url          = {https://doi.org/10.1145/3637528.3671963},
  doi          = {10.1145/3637528.3671963},
  bibsource    = {dblp computer science bibliography, https://dblp.org}
}

@inproceedings{DBLP:conf/acl/LiangZ0Z24,
  author       = {Wenxin Liang and
                  Tingyu Zhang and
                  Han Liu and
                  Feng Zhang},
  no_editor       = {Lun{-}Wei Ku and
                  Andre Martins and
                  Vivek Srikumar},
  title        = {{SELP:} {A} Semantically-Driven Approach for Separated and Accurate
                  Class Prototypes in Few-Shot Text Classification},
  booktitle    = {ACL Findings},
 no_pages =  {9732--9741},
  no_publisher    = {ACL},
  year         = {2024},no_url          = {https://aclanthology.org/2024.findings-acl.579},
  bibsource    = {dblp computer science bibliography, https://dblp.org}
}

@article{DBLP:journals/corr/abs-2405-11524,
  author       = {Mengyu Li and
                  Yonghao Liu and
                  Fausto Giunchiglia and
                  Xiaoyue Feng and
                  Renchu Guan},
  title        = {Simple-Sampling and Hard-Mixup with Prototypes to Rebalance Contrastive
                  Learning for Text Classification},
  journal      = {arXiv:2405.11524},
  year         = {2024},
}

@inproceedings{higen,
  author       = {Vidit Jain and
                  Mukund Rungta and
                  Yuchen Zhuang and
                  Yue Yu and
                  Zeyu Wang and
                  Mu Gao and
                  Jeffrey Skolnick and
                  Chao Zhang},
  no_editor       = {Yvette Graham and
                  Matthew Purver},
  title        = {{HiGen}: Hierarchy-Aware Sequence Generation for Hierarchical Text Classification},
  booktitle    = {EACL},
  no_pages        = {1354--1368},
  publisher    = {ACL},
  year         = {2024},
  no_url          = {https://aclanthology.org/2024.eacl-long.82},
  bibsource    = {dblp computer science bibliography, https://dblp.org}
}

@inproceedings{DBLP:conf/emnlp/UHGP23,
  author       = {Simon Chi Lok U and
                  Jie He and
                  V{\'{\i}}ctor Guti{\'{e}}rrez{-}Basulto and
                  Jeff Z. Pan},
  no_editor       = {Houda Bouamor and
                  Juan Pino and
                  Kalika Bali},
  title        = {Instances and Labels: Hierarchy-aware Joint Supervised Contrastive
                  Learning for Hierarchical Multi-Label Text Classification},
  booktitle    = {EMNLP Findings},
  no_pages        = {8858--8875},
  no_publisher    = {ACL},
  year         = {2023},
  no_url          = {https://doi.org/10.18653/v1/2023.findings-emnlp.594},
  doi          = {10.18653/V1/2023.FINDINGS-EMNLP.594},
  bibsource    = {dblp computer science bibliography, https://dblp.org}
}

@article{liu2016recurrent,
  title={Recurrent neural network for text classification with multi-task learning},
  author={Liu, Pengfei and Qiu, Xipeng and Huang, Xuanjing},
  journal={arXiv:1605.05101},
  year={2016}
}

@article{bahdanau2014neural,
  title={Neural machine translation by jointly learning to align and translate},
  author={Bahdanau, Dzmitry and Cho, Kyunghyun and Bengio, Yoshua},
  journal={arXiv:1409.0473},
  year={2014}
}

@inproceedings{DBLP:conf/sigir/DaiYCX24,
  author       = {Le Dai and
                  Yu Yin and
                  Enhong Chen and
                  Hui Xiong},
  no_editor       = {Grace Hui Yang and
                  Hongning Wang and
                  Sam Han and
                  Claudia Hauff and
                  Guido Zuccon and
                  Yi Zhang},
  title        = {Unifying Graph Retrieval and Prompt Tuning for Graph-Grounded Text
                  Classification},
  booktitle    = {SIGIR},
 no_pages =  {2682--2686},
  no_publisher    = {{ACM}},
  year         = {2024},no_url          = {https://doi.org/10.1145/3626772.3657934},
  doi          = {10.1145/3626772.3657934},
  bibsource    = {dblp computer science bibliography, https://dblp.org}
}

@article{DBLP:journals/corr/abs-2406-17534-retrieval-style,
  author       = {Huiyao Chen and
                  Yu Zhao and
                  Zulong Chen and
                  Mengjia Wang and
                  Liangyue Li and
                  Meishan Zhang and
                  Min Zhang},
  title        = {Retrieval-style In-context Learning for Few-shot Hierarchical Text
                  Classification},
  journal      = {Trans. Assoc. Comput. Linguistics},
  volume       = {12},
  no_pages        = {1214--1231},
  year         = {2024},
  no_url          = {https://doi.org/10.1162/tacl\_a\_00697},
  doi          = {10.1162/TACL\_A\_00697},
  bibsource    = {dblp computer science bibliography, https://dblp.org}
}

@article{DBLP:journals/corr/abs-2401-03158,
  author       = {Hui Wu and
                  Yuanben Zhang and
                  Zhonghe Han and
                  Yingyan Hou and
                  Lei Wang and
                  Siye Liu and
                  Qihang Gong and
                  Yunping Ge},
  title        = {Quartet Logic: {A} Four-Step Reasoning {(QLFR)} framework for advancing
                  Short Text Classification},
  journal      = {arXiv:2401.03158},
  year         = {2024},
}

@article{DBLP:journals/access/ThaminkaewLV24,
  author       = {Thanakorn Thaminkaew and
                  Piyawat Lertvittayakumjorn and
                  Peerapon Vateekul},
  title        = {Prompt-Based Label-Aware Framework for Few-Shot Multi-Label Text Classification},
  journal      = {{IEEE} Access},
  volume       = {12},
 no_pages =  {28310--28322},
  year         = {2024},no_url          = {https://doi.org/10.1109/ACCESS.2024.3367994},
  doi          = {10.1109/ACCESS.2024.3367994},
  bibsource    = {dblp computer science bibliography, https://dblp.org}
}

@article{DBLP:journals/tkde/HuLZHNL24,
  author       = {Linmei Hu and
                  Zeyi Liu and
                  Ziwang Zhao and
                  Lei Hou and
                  Liqiang Nie and
                  Juanzi Li},
  title        = {A Survey of Knowledge Enhanced Pre-Trained Language Models},
  journal      = {{IEEE} Trans. Knowl. Data Eng.},
  volume       = {36},
  number       = {4},
 no_pages =  {1413--1430},
  year         = {2024},no_url          = {https://doi.org/10.1109/TKDE.2023.3310002},
  doi          = {10.1109/TKDE.2023.3310002},
  bibsource    = {dblp computer science bibliography, https://dblp.org}
}

@inproceedings{radar,
  title="{RADAr}: A Transformer-based Autoregressive Decoder Architecture for Hierarchical Text Classification",
  author="Yousef Younes and Lukas Galke and Ansgar Scherp",
  booktitle="ECAI",
  no_publisher="AAAI",
  year="2024"
}

@ARTICLE{thaminkaew2024ieeeaccess,
  author={Thaminkaew, Thanakorn and Lertvittayakumjorn, Piyawat and Vateekul, Peerapon},
  journal={IEEE Access}, 
  title={Prompt-Based Label-Aware Framework for Few-Shot Multi-Label Text Classification}, 
  year={2024},
  volume={12},
  number={},
 no_pages = {28310-28322},
  doi={10.1109/ACCESS.2024.3367994}
}

@ARTICLE{pushing-the-limit-11346014-rgpt,
  author={Wang, Mengyao and Zhang, Yazhou and Ren, Chenyu and Li, Qiuchi and Tiwari, Prayag and Wang, Benyou and Qin, Jing},
  journal={IEEE Transactions on Neural Networks and Learning Systems}, 
  title={Adaptive Boosting {LLMs} for Text Classification}, 
  year={2026},
  volume={},
  number={},
  pages={1-12},
  doi={10.1109/TNNLS.2025.3639613}}

@article{DBLP:journals/corr/abs-2401-01667-MLP-Compass,
  author       = {Li Zhou and
                  Wenyu Chen and
                  Yong Cao and
                  Dingyi Zeng and
                  Wanlong Liu and
                  Hong Qu},
  title        = {{MLP}s Compass: What is learned when {MLP}s are combined with {PLM}s?},
  journal      = {arXiv:2401.01667},
  year         = {2024},
}

@inproceedings{DBLP:conf/sigir/Nallapati04,
  author       = {Ramesh Nallapati},
  no_editor       = {Mark Sanderson and
                  Kalervo J{\"{a}}rvelin and
                  James Allan and
                  Peter Bruza},
  title        = {Discriminative models for information retrieval},
  booktitle    = {{SIGIR}},
 no_pages =  {64--71},
  publisher    = {{ACM}},
  year         = {2004},no_url          = {https://doi.org/10.1145/1008992.1009006},
  doi          = {10.1145/1008992.1009006},
  bibsource    = {dblp computer science bibliography, https://dblp.org}
}

@inproceedings{DBLP:conf/iclr/PatelLRCRC23,
  author       = {Ajay Patel and
                  Bryan Li and
                  Mohammad Sadegh Rasooli and
                  Noah Constant and
                  Colin Raffel and
                  Chris Callison{-}Burch},
  title        = {Bidirectional Language Models Are Also Few-shot Learners},
  booktitle    = {ICLR},
  publisher    = {OpenReview.net},
  year         = {2023},no_url          = {https://openreview.net/pdf?id=wCFB37bzud4},
  bibsource    = {dblp computer science bibliography, https://dblp.org}
}

@inproceedings{DBLP:conf/aaai/Hu0CSLW024,
  author       = {Beizhe Hu and
                  Qiang Sheng and
                  Juan Cao and
                  Yuhui Shi and
                  Yang Li and
                  Danding Wang and
                  Peng Qi},
  no_editor       = {Michael J. Wooldridge and
                  Jennifer G. Dy and
                  Sriraam Natarajan},
  title        = {Bad Actor, Good Advisor: Exploring the Role of Large Language Models
                  in Fake News Detection},
  booktitle    = {AAAI},
 no_pages =  {22105--22113},
  no_publisher    = {{AAAI} Press},
  year         = {2024},no_url          = {https://doi.org/10.1609/aaai.v38i20.30214},
  doi          = {10.1609/AAAI.V38I20.30214},
  bibsource    = {dblp computer science bibliography, https://dblp.org}
}

@inproceedings{yu2022constrained,
  author       = {Chao Yu and
                  Yi Shen and
                  Yue Mao},
  no_editor       = {Enrique Amig{\'{o}} and
                  Pablo Castells and
                  Julio Gonzalo and
                  Ben Carterette and
                  J. Shane Culpepper and
                  Gabriella Kazai},
  title        = {Constrained Sequence-to-Tree Generation for Hierarchical Text Classification},
  booktitle    = {{SIGIR}},
 no_pages =  {1865--1869},
  publisher    = {{ACM}},
  year         = {2022},no_url          = {https://doi.org/10.1145/3477495.3531765},
  doi          = {10.1145/3477495.3531765},
  bibsource    = {dblp computer science bibliography, https://dblp.org}
}

@article{DBLP:journals/corr/abs-2402-16844,
  author       = {Benjamin Bergner and
                  Andrii Skliar and
                  Amelie Royer and
                  Tijmen Blankevoort and
                  Yuki M. Asano and
                  Babak Ehteshami Bejnordi},
  title        = {Think Big, Generate Quick: {LLM-to-SLM} for Fast Autoregressive Decoding},
  journal      = {arXiv:2402.16844},
  year         = {2024},
}

@inproceedings{DBLP:conf/emnlp/PecherSB25-100-samples,
  author       = {Branislav Pecher and
                  Ivan Srba and
                  M{\'{a}}ria Bielikov{\'{a}}},
  no_editor       = {Christos Christodoulopoulos and
                  Tanmoy Chakraborty and
                  Carolyn Rose and
                  Violet Peng},
  title        = {Comparing Specialised Small and General Large Language Models on Text
                  Classification: 100 Labelled Samples to Achieve Break-Even Performance},
  booktitle    = {EMNLP},
  no_pages        = {165--184},
  publisher    = {ACL},
  year         = {2025},
  no_url          = {https://doi.org/10.18653/v1/2025.emnlp-main.9},
  doi          = {10.18653/V1/2025.EMNLP-MAIN.9},
  bibsource    = {dblp computer science bibliography, https://dblp.org}
}

@article{DBLP:journals/corr/abs-2303-08774,
  author       = {OpenAI},
  title        = {{GPT-4} Technical Report},
  journal      = {arXiv:2303.08774},
  year         = {2023},
}

@article{DBLP:journals/tkde/MoreoES20,
  author       = {Alejandro Moreo and
                  Andrea Esuli and
                  Fabrizio Sebastiani},
  title        = {Learning to Weight for Text Classification},
  journal      = {{IEEE} Trans. Knowl. Data Eng.},
  volume       = {32},
  number       = {2},
 no_pages =  {302--316},
  year         = {2020},no_url          = {https://doi.org/10.1109/TKDE.2018.2883446},
  doi          = {10.1109/TKDE.2018.2883446},
  bibsource    = {dblp computer science bibliography, https://dblp.org}
}

@article{DBLP:journals/csur/Sebastiani02,
  author       = {Fabrizio Sebastiani},
  title        = {Machine learning in automated text categorization},
  journal      = {{ACM} Comput. Surv.},
  volume       = {34},
  number       = {1},
 no_pages =  {1--47},
  year         = {2002},no_url          = {https://doi.org/10.1145/505282.505283},
  doi          = {10.1145/505282.505283},
  bibsource    = {dblp computer science bibliography, https://dblp.org}
}

@inproceedings{chain-of-thought,
 author = {Wei, Jason and Wang, Xuezhi and Schuurmans, Dale and Bosma, Maarten and ichter, brian and Xia, Fei and Chi, Ed and Le, Quoc V and Zhou, Denny},
 booktitle = {NeurIPS},
 no_editor = {S. Koyejo and S. Mohamed and A. Agarwal and D. Belgrave and K. Cho and A. Oh},
no_pages =  {24824--24837},
 publisher = {Curran Associates, Inc.},
 title = {Chain-of-Thought Prompting Elicits Reasoning in Large Language Models},no_url = {https://proceedings.neurips.cc/paper_files/paper/2022/file/9d5609613524ecf4f15af0f7b31abca4-Paper-Conference.pdf},
 no_volume = {35},
 year = {2022}
}

@article{yang2023harnessing,
  author       = {Jingfeng Yang and
                  Hongye Jin and
                  Ruixiang Tang and
                  Xiaotian Han and
                  Qizhang Feng and
                  Haoming Jiang and
                  Shaochen Zhong and
                  Bing Yin and
                  Xia Ben Hu},
  title        = {Harnessing the Power of LLMs in Practice: {A} Survey on ChatGPT and
                  Beyond},
  journal      = {{ACM} Trans. Knowl. Discov. Data},
  volume       = {18},
  no_number       = {6},
  no_pages        = {160:1--160:32},
  year         = {2024}
}

@inproceedings{lester-etal-2021-power,
    title = "The Power of Scale for Parameter-Efficient Prompt Tuning",
    author = "Lester, Brian  and
      Al-Rfou, Rami  and
      Constant, Noah",
    no_editor = "Moens, Marie-Francine  and
      Huang, Xuanjing  and
      Specia, Lucia  and
      Yih, Scott Wen-tau",
    booktitle = "EMNLP",
    no_month = nov,
    year = "2021",
    no_address = "Online and Punta Cana, Dominican Republic",
    no_publisher = "ACL",
no_url = "https://aclanthology.org/2021.emnlp-main.243",
    doi = "10.18653/v1/2021.emnlp-main.243",
   no_pages =  "3045--3059",
}

@inproceedings{liu-etal-2022-p,
    title = "{P}-Tuning: Prompt Tuning Can Be Comparable to Fine-tuning Across Scales and Tasks",
    author = "Liu, Xiao  and
      Ji, Kaixuan  and
      Fu, Yicheng  and
      Tam, Weng  and
      Du, Zhengxiao  and
      Yang, Zhilin  and
      Tang, Jie",
    no_editor = "Muresan, Smaranda  and
      Nakov, Preslav  and
      Villavicencio, Aline",
    booktitle = "ACL (Volume 2: Short Papers)",
    no_month = may,
    year = "2022",
    no_address = "Dublin, Ireland",
    no_publisher = "ACL",
no_url = "https://aclanthology.org/2022.acl-short.8",
    doi = "10.18653/v1/2022.acl-short.8",
   no_pages =  "61--68",
}

@inproceedings{
wei2022finetuned,
title={Finetuned Language Models are Zero-Shot Learners},
author={Jason Wei and Maarten Bosma and Vincent Zhao and Kelvin Guu and Adams Wei Yu and Brian Lester and Nan Du and Andrew M. Dai and Quoc V Le},
booktitle={ICLR},
year={2022},no_url={https://openreview.net/forum?id=gEZrGCozdqR}
}

@article{fieldsSurveyTextClassification2024,
  title = {A {{Survey}} of {{Text Classification}} with {{Transformers}}: {{How}} Wide? {{How}} Large? {{How}} Long? {{How}} Accurate? {{How}} Expensive? {{How}} Safe?},
  shorttitle = {A {{Survey}} of {{Text Classification}} with {{Transformers}}},
  author = {Fields, John and Chovanec, Kevin and Madiraju, Praveen},
  year = {2024},
  journal = {IEEE Access},
 no_pages =  {1--1},
  issn = {2169-3536},
  doi = {10.1109/ACCESS.2024.3349952},
}

@article{canutoThoroughEvaluationDistanceBased2018,
  title = {A {{Thorough Evaluation}} of {{Distance-Based Meta-Features}} for {{Automated Text Classification}}},
  author = {Canuto, S{\'e}rgio and Sousa, Daniel Xavier and Gon{\c c}alves, Marcos Andr{\'e} and Rosa, Thierson Couto},
  year = {2018},
  no_month = dec,
  journal = {IEEE Trans. Knowl. Data Eng.},
  volume = {30},
  no_number = {12},
 no_pages =  {2242--2256},
  issn = {1558-2191},
  doi = {10.1109/TKDE.2018.2820051},
}

@article{pengHierarchicalTaxonomyAwareAttentional2021,
  title = {Hierarchical {{Taxonomy-Aware}} and {{Attentional Graph Capsule RCNNs}} for {{Large-Scale Multi-Label Text Classification}}},
  author = {Peng, Hao and Li, Jianxin and Wang, Senzhang and Wang, Lihong and Gong, Qiran and Yang, Renyu and Li, Bo and Yu, Philip S. and He, Lifang},
  year = {2021},
  journal = {IEEE Trans. on Knowl. and Data Eng.},
  volume = {33},
  number = {6},
 no_pages =  {2505--2519},
  issn = {1558-2191},
  doi = {10.1109/TKDE.2019.2959991},
}

@article{tan2022,
  title = {Dynamic {{Embedding Projection-Gated Convolutional Neural Networks}} for {{Text Classification}}},
  author = {Tan, Zhipeng and Chen, Jing and Kang, Qi and Zhou, Mengchu and Abusorrah, Abdullah and Sedraoui, Khaled},
  year = {2022},
  journal = {IEEE Trans. on Neural Networks and Learning Systems},
  volume = {33},
  number = {3},
 no_pages =  {973--982},
  issn = {2162-2388},
  doi = {10.1109/TNNLS.2020.3036192},
}

@article{moreo2020,
  title = {Learning to Weight for Text Classification},
  author = {Moreo, Alejandro and Esuli, Andrea and Sebastiani, Fabrizio},
  year = {2020},
  journal = {IEEE Trans. on Knowl. and Data Eng.},
  volume = {32},
  no_number = {2},
 no_pages =  {302--316},
  issn = {1558-2191},
  doi = {10.1109/TKDE.2018.2883446},
}

@article{zhu2023,
  title = {Prompt-Learning for Short Text Classification},
  author = {Zhu, Yi and Wang, Ye and Qiang, Jipeng and Wu, Xindong},
  year = {2023},
  journal = {IEEE Trans. on Knowl. and Data Eng.},
 no_pages =  {1--13},
  issn = {1558-2191},
  doi = {10.1109/TKDE.2023.3332787},
}

@inproceedings{zhang2023pretraining,
  author       = {Yu Zhang and
                  Hao Cheng and
                  Zhihong Shen and
                  Xiaodong Liu and
                  Ye{-}Yi Wang and
                  Jianfeng Gao},
  title        = {Pre-training Multi-task Contrastive Learning Models for Scientific
                  Literature Understanding},
  booktitle    = {{EMNLP} Findings},
  no_pages        = {12259--12275},
  no_publisher    = {Association for Computational Linguistics},
  year         = {2023}
}

@article{am-bert,
  title        = {Enhancing text classification with attention matrices based on {BERT}},
  author       = {Yu, Zhiyi and Li, Hong and Feng, Jialin},
  year         = 2023,
  no_month        = 11,
  journal      = {Expert Systems},
 no_pages =  {},
  doi          = {10.1111/exsy.13512}
}

@inproceedings{StylianouEtAl2023,
  title        = {Domain-Aligned Data Augmentation for Low-Resource and Imbalanced Text Classification},
  author       = {Nikolaos Stylianou and Despoina Chatzakou and Theodora Tsikrika and Stefanos Vrochidis and Ioannis Kompatsiaris},
  year         = 2023,
  booktitle    = {ECIR},
  publisher    = {Springer},
  no_series       = {Lecture Notes in Computer Science},
  no_volume       = 13981,
 no_pages =  {172--187},
  doi          = {10.1007/978-3-031-28238-6\_12},
  no_editor    = {Jaap Kamps and Lorraine Goeuriot and Fabio Crestani and Maria Maistro and Hideo Joho and Brian Davis and Cathal Gurrin and Udo Kruschwitz and Annalina Caputo},
  no_url       = {https://doi.org/10.1007/978-3-031-28238-6\_12},
  bibsource    = {dblp computer science bibliography, https://dblp.org}
}

@article{DBLP:journals/tist/LiPLXYSYH22,
  title        = {A Survey on Text Classification: From Traditional to Deep Learning},
  author       = {Qian Li and Hao Peng and Jianxin Li and Congying Xia and Renyu Yang and Lichao Sun and Philip S. Yu and Lifang He},
  year         = 2022,
  journal      = {{ACM} Trans. Intell. Syst. Technol.},
  volume       = 13,
  number       = 2,
 no_pages =  {31:1--31:41},
  doi          = {10.1145/3495162},
  no_url       = {https://doi.org/10.1145/3495162},
  bibsource    = {dblp computer science bibliography, https://dblp.org}
}

@article{wang2023graph,
  author       = {Kunze Wang and
                  Yihao Ding and
                  Soyeon Caren Han},
  title        = {Graph neural networks for text classification: a survey},
  journal      = {Artif. Intell. Rev.},
  volume       = {57},
  number       = {8},
  no_pages        = {190},
  year         = {2024}
}

@article{Yang-ClimateChangeClassifiert-2023,
  title        = {Research on the Automatic Subject-Indexing Method of Academic Papers Based on Climate Change Domain Ontology},
  author       = {Yang, Heng and Wang, Nan and Yang, Lina and Liu, Wei and Wang, Sili},
  year         = 2023,
  no_month        = {Feb},
  journal      = {Sustainability},
  publisher    = {MDPI AG},
  volume       = 15,
  number       = 5,
 no_pages =  3919,
  doi          = {10.3390/su15053919},
  issn         = {2071-1050},
  no_url       = {http://dx.doi.org/10.3390/su15053919}
}

@inproceedings{DBLP:conf/cdmake/KarlS23,
  title        = {Transformers are Short-Text Classifiers},
  author       = {Fabian Karl and Ansgar Scherp},
  year         = 2023,
  booktitle    = {Int. Cross-Domain Conf., {CD-MAKE}},
  publisher    = {Springer},
  no_series       = {Lecture Notes in Computer Science},
  no_volume       = 14065,
 no_pages =  {103--122},
  doi          = {10.1007/978-3-031-40837-3\_7},no_url          = {https://doi.org/10.1007/978-3-031-40837-3\_7},
  no_editor       = {Andreas Holzinger and Peter Kieseberg and Federico Cabitza and Andrea Campagner and A Min Tjoa and Edgar R. Weippl},
  bibsource    = {dblp computer science bibliography, https://dblp.org}
}

@article{DBLP:journals/csur/MinaeeKCNCG21,
  title        = {Deep Learning-based Text Classification: {A} Comprehensive Review},
  author       = {Shervin Minaee and Nal Kalchbrenner and Erik Cambria and Narjes Nikzad and Meysam Chenaghlu and Jianfeng Gao},
  year         = 2022,
  journal      = {{ACM} Comput. Surv.},
  volume       = 54,
  number       = 3,
 no_pages =  {62:1--62:40},
  doi          = {10.1145/3439726},
  no_url       = {https://doi.org/10.1145/3439726},
  bibsource    = {dblp computer science bibliography, https://dblp.org}
}

@misc{raihan2021-survey,
  title        = {Deep Learning Techniques for Text Classification},
  author       = {Diardano Raihan},
  year         = 2021,no_url          = {https://towardsdatascience.com/deep-learning-techniques-for-text-classification-78d9dc40bf7c}
}

@inproceedings{DBLP:conf/ijcnn/TranSZPB22,
  title        = {Comparing the Robustness of Classical and Deep Learning Techniques for Text Classification},
  author       = {Quynh Tran and Krystsina Shpileuskaya and Elaine Zaunseder and Larissa Putzar and Sven Blankenburg},
  year         = 2022,
  booktitle    = {IJCNN},
  publisher    = {{IEEE}},
 no_pages =  {1--10},
  doi          = {10.1109/IJCNN55064.2022.9892242},
  no_url       = {https://doi.org/10.1109/IJCNN55064.2022.9892242},
  bibsource    = {dblp computer science bibliography, https://dblp.org}
}

@inproceedings{DBLP:conf/eacl/SchwenkBCL17,
  title        = {Very Deep Convolutional Networks for Text Classification},
  author       = {Alexis Conneau and Holger Schwenk and Lo{\"{\i}}c Barrault and Yann LeCun},
  year         = 2017,
  booktitle    = {EACL},
  no_publisher    = {ACL},
 no_pages =  {1107--1116},
  doi          = {10.18653/v1/e17-1104},
  no_editor    = {Mirella Lapata and Phil Blunsom and Alexander Koller},
  no_url       = {https://doi.org/10.18653/v1/e17-1104},
  bibsource    = {dblp computer science bibliography, https://dblp.org}
}

@inproceedings{DBLP:journals/corr/abs-2211-02563,
  author       = {Yasmen Wahba and
                  Nazim H. Madhavji and
                  John Steinbacher},
  title        = {A Comparison of {{SVM}} Against Pre-trained Language Models ({PLMs}) for
                  Text Classification Tasks},
  booktitle    = {{LOD} {(2)}},
  no_series       = {Lecture Notes in Computer Science},
  no_volume       = {13811},
  no_pages        = {304--313},
  publisher    = {Springer},
  year         = {2022}
}

@inproceedings{DBLP:conf/emnlp/Kim14,
  title        = {Convolutional Neural Networks for Sentence Classification},
  author       = {Yoon Kim},
  year         = 2014,
  booktitle    = {{ACL}},
  no_publisher    = {{ACL}},
 no_pages =  {1746--1751},
  doi          = {10.3115/v1/d14-1181},
  no_editor    = {Alessandro Moschitti and Bo Pang and Walter Daelemans},
  no_url       = {https://doi.org/10.3115/v1/d14-1181},
  bibsource    = {dblp computer science bibliography, https://dblp.org}
}

@inproceedings{DBLP:conf/ijcnlp/ZhangW17,
  title        = {A Sensitivity Analysis of (and Practitioners' Guide to) Convolutional Neural Networks for Sentence Classification},
  author       = {Ye Zhang and Byron C. Wallace},
  year         = 2017,
  booktitle    = {{IJCNLP}},
  no_publisher    = {Asian Federation of Natural Language Processing},
 no_pages =  {253--263},no_url          = {https://aclanthology.org/I17-1026/},
  no_editor    = {Greg Kondrak and Taro Watanabe},
  bibsource    = {dblp computer science bibliography, https://dblp.org}
}

@inproceedings{DBLP:conf/aiia/BreazzanoC021a,
  title        = {Multi-task and Generative Adversarial Learning for Robust and Sustainable Text Classification},
  author       = {Claudia Breazzano and Danilo Croce and Roberto Basili},
  year         = 2021,
  booktitle    = {AIxIA 2021},
  publisher    = {Springer},
  no_series       = {Lecture Notes in Computer Science},
  no_volume       = 13196,
 no_pages =  {228--244},
  doi          = {10.1007/978-3-031-08421-8\_16},
  no_editor    = {Stefania Bandini and Francesca Gasparini and Viviana Mascardi and Matteo Palmonari and Giuseppe Vizzari},
  no_url       = {https://doi.org/10.1007/978-3-031-08421-8\_16},
  bibsource    = {dblp computer science bibliography, https://dblp.org}
}

@article{LiuEtAl2022-LongText,
  title        = {Hierarchical Graph Convolutional Networks for Structured Long Document Classification},
  author       = {Liu, Tengfei and Hu, Yongli and Wang, Boyue and Sun, Yanfeng and Gao, Junbin and Yin, Baocai},
  year         = 2022,
  journal      = {IEEE Trans. on Neural Networks and Learning Systems},
  volume       = {},
  number       = {},
 no_pages =  {1--15},
  doi          = {10.1109/TNNLS.2022.3185295}
}

@article{DBLP:journals/corr/abs-2112-11389,
  title        = {Supervised Graph Contrastive Pretraining for Text Classification},
  author       = {Samujjwal Ghosh and Subhadeep Maji and Maunendra Sankar Desarkar},
  year         = 2021,
  journal      = {arXiv:2112.11389},
}

@article{DBLP:journals/corr/abs-2206-07253,
  author       = {Zhizhi Yu and
                  Di Jin and
                  Jianguo Wei and
                  Yawen Li and
                  Ziyang Liu and
                  Yue Shang and
                  Jiawei Han and
                  Lingfei Wu},
  title        = {TeKo: Text-Rich Graph Neural Networks With External Knowledge},
  journal      = {{IEEE} Trans. Neural Networks Learn. Syst.},
  volume       = {35},
  number       = {10},
  no_pages        = {14699--14711},
  year         = {2024}
}

@inproceedings{DBLP:conf/ijcnn/SunHCYSM22,
  title        = {Contrastive Learning with Heterogeneous Graph Attention Networks on Short Text Classification},
  author       = {Zhongtian Sun and Anoushka Harit and Alexandra I. Cristea and Jialin Yu and Lei Shi and Noura Al Moubayed},
  year         = 2022,
  booktitle    = {{IJCNN}},
  publisher    = {{IEEE}},
 no_pages =  {1--6},
  doi          = {10.1109/IJCNN55064.2022.9892257},
  no_url       = {https://doi.org/10.1109/IJCNN55064.2022.9892257},
  bibsource    = {dblp computer science bibliography, https://dblp.org}
}

@inproceedings{YuanEtAl-MSVM-kNN-2008,
  title        = {{MSVM-kNN}: {C}ombining {SVM} and {k-NN} for Multi-class Text Classification},
  author       = {Yuan, Pingpeng and Chen, Yuqin and Jin, Hai and Huang, Li},
  year         = 2008,
  booktitle    = {IEEE Int. Workshop on Semantic Computing and Systems},
  volume       = {},
  number       = {},
 no_pages =  {133--140},
  doi          = {10.1109/WSCS.2008.36}
}

@article{sf-cnn-iasc.2023.027429,
  title        = {{SF-CNN}: {D}eep Text Classification and Retrieval for Text Documents},
  author       = {R. Sarasu and K. K. Thyagharajan and N. R. Shanker},
  year         = 2023,
  journal      = {Intelligent Automation \& Soft Computing},
  volume       = 35,
  number       = 2,
 no_pages =  {1799--1813},
  doi          = {10.32604/iasc.2023.027429},
  issn         = {2326-005X},
  no_url       = {http://www.techscience.com/iasc/v35n2/48876}
}

@inproceedings{DBLP:conf/coling/HuangCC22,
  title        = {ConTextING: Granting Document-Wise Contextual Embeddings to Graph Neural Networks for Inductive Text Classification},
  author       = {Yen{-}Hao Huang and Yi{-}Hsin Chen and Yi{-}Shin Chen},
  year         = 2022,
  booktitle    = {COLING},
  no_publisher    = {Int. Committee on Computational Linguistics},
 no_pages =  {1163--1168},no_url          = {https://aclanthology.org/2022.coling-1.100},
  no_editor    = {Nicoletta Calzolari and Chu{-}Ren Huang and Hansaem Kim and James Pustejovsky and Leo Wanner and Key{-}Sun Choi and Pum{-}Mo Ryu and Hsin{-}Hsi Chen and Lucia Donatelli and Heng Ji and Sadao Kurohashi and Patrizia Paggio and Nianwen Xue and Seokhwan Kim and Younggyun Hahm and Zhong He and Tony Kyungil Lee and Enrico Santus and Francis Bond and Seung{-}Hoon Na},
  bibsource    = {dblp computer science bibliography, https://dblp.org}
}

@inproceedings{galkescherp-acl2022,
  title        = {Bag-of-Words vs. Graph vs. Sequence in Text Classification: Questioning the Necessity of Text-Graphs and the Surprising Strength of a Wide {MLP}},
  author       = {Lukas Galke and Ansgar Scherp},
  year         = 2022,
  booktitle    = {{ACL}},
  no_publisher    = {ACL},
 no_pages =  {4038--4051},
  doi          = {10.18653/v1/2022.acl-long.279},
  no_editor    = {Smaranda Muresan and Preslav Nakov and Aline Villavicencio},
  no_url       = {https://doi.org/10.18653/v1/2022.acl-long.279},
  bibsource    = {dblp computer science bibliography, https://dblp.org}
}

@inproceedings{DBLP:conf/emnlp/ZhengWYD22,
  title        = {Simplified Graph Learning for Inductive Short Text Classification},
  author       = {Kaixin Zheng and Yaqing Wang and Quanming Yao and Dejing Dou},
  year         = 2022,
  booktitle    = {{EMNLP}},
  no_publisher    = {ACL},
 no_pages =  {10717--10724},no_url          = {https://aclanthology.org/2022.emnlp-main.735},
  no_editor    = {Yoav Goldberg and Zornitsa Kozareva and Yue Zhang},
  bibsource    = {dblp computer science bibliography, https://dblp.org}
}

@article{DBLP:journals/apin/TanRW23,
  title        = {An adaptive convolution with label embedding for text classification},
  author       = {Changgeng Tan and Yuan Ren and Chen Wang},
  year         = 2023,
  journal      = {Appl. Intell.},
  volume       = 53,
  number       = 1,
 no_pages =  {804--812},
  doi          = {10.1007/s10489-021-02702-x},
  no_url       = {https://doi.org/10.1007/s10489-021-02702-x},
  bibsource    = {dblp computer science bibliography, https://dblp.org}
}

@article{ATTIEH2023110215,
  title        = {Supervised term-category feature weighting for improved text classification},
  author       = {Joseph Attieh and Joe Tekli},
  year         = 2023,
  journal      = {Knowl.-Based Systems},
  volume       = 261,
 no_pages =  110215,
  doi          = {https://doi.org/10.1016/j.knosys.2022.110215},
  issn         = {0950-7051},
  no_url       = {https://www.sciencedirect.com/science/article/pii/S0950705122013119}
}

@inproceedings{DBLP:conf/acl/ChenMLY20,
  title        = {Hierarchy-aware Label Semantics Matching Network for Hierarchical Text Classification},
  author       = {Haibin Chen and Qianli Ma and Zhenxi Lin and Jiangyue Yan},
  year         = 2021,
  booktitle    = {{ACL/IJCNLP}},
  no_publisher    = {ACL},
 no_pages =  {4370--4379},
  doi          = {10.18653/v1/2021.acl-long.337},
  no_editor    = {Chengqing Zong and Fei Xia and Wenjie Li and Roberto Navigli},
  no_url       = {https://doi.org/10.18653/v1/2021.acl-long.337},
  bibsource    = {dblp computer science bibliography, https://dblp.org}
}

@inproceedings{DBLP:conf/acl/LinMSHKLW21,
  title        = {{BertGCN}: Transductive Text Classification by Combining {GNN} and {BERT}},
  author       = {Yuxiao Lin and Yuxian Meng and Xiaofei Sun and Qinghong Han and Kun Kuang and Jiwei Li and Fei Wu},
  year         = 2021,
  booktitle    = {{ACL/IJCNLP} Findings},
  no_publisher    = {ACL},
  no_series       = {Findings of {ACL}},
  no_volume       = {{ACL/IJCNLP} 2021},
 no_pages =  {1456--1462},
  doi          = {10.18653/v1/2021.findings-acl.126},
  no_editor    = {Chengqing Zong and Fei Xia and Wenjie Li and Roberto Navigli},
  no_url       = {https://doi.org/10.18653/v1/2021.findings-acl.126},
  bibsource    = {dblp computer science bibliography, https://dblp.org}
}

@inproceedings{ctgcn,
  title        = {Circulant Tensor Graph Convolutional Network for Text Classification},
  author       = {Xuran Xu and Tong Zhang and Chunyan Xu and Zhen Cui},
  year         = 2021,
  booktitle    = {Pattern Recognition - 6th Asian Conf., {ACPR}},
  publisher    = {Springer},
  no_series       = {Lecture Notes in Computer Science},
  no_volume       = 13188,
 no_pages =  {32--46},
  doi          = {10.1007/978-3-031-02375-0\_3},
  no_url       = {https://doi.org/10.1007/978-3-031-02375-0\_3},
  bibsource    = {dblp computer science bibliography, https://dblp.org}
}

@article{DBLP:journals/pr/TarekegnGM21,
  title        = {A review of methods for imbalanced multi-label classification},
  author       = {Adane Nega Tarekegn and Mario Giacobini and Krzysztof Michalak},
  year         = 2021,
  journal      = {Pattern Recognit.},
  no_volume       = 118,
 no_pages =  107965,
  doi          = {10.1016/j.patcog.2021.107965},
  no_url       = {https://doi.org/10.1016/j.patcog.2021.107965},
  bibsource    = {dblp computer science bibliography, https://dblp.org}
}

@article{DBLP:journals/corr/abs-2107-03158,
  title        = {A Survey on Data Augmentation for Text Classification},
  author       = {Markus Bayer and Marc{-}Andr{\'{e}} Kaufhold and Christian Reuter},
  year         = 2023,
  journal      = {{ACM} Comput. Surv.},
  volume       = 55,
  number       = 7,
 no_pages =  {146:1--146:39}
}

@article{DBLP:journals/corr/abs-2006-15795,
  title        = {Combine Convolution with Recurrent Networks for Text Classification},
  author       = {Shengfei Lyu and Jiaqi Liu},
  year         = 2020,
  journal      = {arXiv:2006.15795},
}

@article{T5,
  title        = {Exploring the Limits of Transfer Learning with a Unified Text-to-Text Transformer},
  author       = {Colin Raffel and Noam Shazeer and Adam Roberts and Katherine Lee and Sharan Narang and Michael Matena and Yanqi Zhou and Wei Li and Peter J. Liu},
  year         = 2020,
  journal      = {JMLR},
  volume       = 21,
 no_pages =  {140:1--140:67}
}

@article{DBLP:journals/jmlr/BleiNJ03,
  title        = {Latent Dirichlet Allocation},
  author       = {David M. Blei and Andrew Y. Ng and Michael I. Jordan},
  year         = 2003,
  journal      = {JMLR},
  volume       = 3,
 no_pages =  {993--1022}
}

@book{book:hamilton:grl,
  title        = {Graph Representation Learning},
  author       = {William L. Hamilton},
  year         = 2020,
  publisher    = {Morgan and Claypool},
  no_url       = {https://www.cs.mcgill.ca/~wlh/grl_book/}
}

@inproceedings{DBLP:conf/emnlp/PenningtonSM14,
  title        = {Glove: Global Vectors for Word Representation},
  author       = {Jeffrey Pennington and Richard Socher and Christopher D. Manning},
  year         = 2014,
  booktitle    = {{EMNLP}},
  no_publisher    = {{ACL}},
 no_pages =  {1532--1543},
  doi          = {10.3115/v1/d14-1162},
  no_editor    = {Alessandro Moschitti and Bo Pang and Walter Daelemans},
  no_url       = {https://doi.org/10.3115/v1/d14-1162},
  bibsource    = {dblp computer science bibliography, https://dblp.org}
}

@inproceedings{DBLP:conf/emnlp/HuangMLZW19,
  title        = {Text Level Graph Neural Network for Text Classification},
  author       = {Lianzhe Huang and Dehong Ma and Sujian Li and Xiaodong Zhang and Houfeng Wang},
  year         = 2019,
  booktitle    = {{EMNLP-IJCNLP}},
  no_publisher    = {ACL},
 no_pages =  {3442--3448},
  doi          = {10.18653/v1/D19-1345},
  no_editor    = {Kentaro Inui and Jing Jiang and Vincent Ng and Xiaojun Wan},
  no_url       = {https://doi.org/10.18653/v1/D19-1345},
  bibsource    = {dblp computer science bibliography, https://dblp.org}
}

@article{9837023-chinese,
  title        = {Supervised Contrastive Learning with Term Weighting for Improving Chinese Text Classification},
  author       = {Guo, Jiabao and Zhao, Bo and Liu, Hui and Liu, Yifan and Zhong, Qian},
  year         = 2023,
  journal      = {Tsinghua Science and Technology},
  volume       = 28,
  number       = 1,
 no_pages =  {59--68},
  doi          = {10.26599/TST.2021.9010079}
}

@inproceedings{DBLP:conf/icml/ChaiWHWL20,
  title        = {Description Based Text Classification with Reinforcement Learning},
  author       = {Duo Chai and Wei Wu and Qinghong Han and Fei Wu and Jiwei Li},
  year         = 2020,
  booktitle    = {{ICML}},
  no_publisher    = {{PMLR}},
  no_series       = {PLMR},
  no_volume       = 119,
 no_pages =  {1371--1382},no_url          = {http://proceedings.mlr.press/v119/chai20a.html},
  bibsource    = {dblp computer science bibliography, https://dblp.org}
}

@misc{Bhatia16,
  title        = {The extreme classification repository: Multi-label datasets and code},
  author       = {Bhatia, K. and Dahiya, K. and Jain, H. and Kar, P. and Mittal, A. and Prabhu, Y. and Varma, M.},
  year         = 2016,
    url          = {http://manikvarma.org/downloads/XC/XMLRepository.html}
}

@inproceedings{DBLP:conf/acl/KalchbrennerGB14,
  title        = {A Convolutional Neural Network for Modelling Sentences},
  author       = {Nal Kalchbrenner and Edward Grefenstette and Phil Blunsom},
  year         = 2014,
  booktitle    = {{ACL}},
  no_publisher    = {The Association for Computer Linguistics},
 no_pages =  {655--665},
  doi          = {10.3115/v1/p14-1062},
  no_url       = {https://doi.org/10.3115/v1/p14-1062},
  bibsource    = {dblp computer science bibliography, https://dblp.org}
}

@inproceedings{DBLP:conf/acl/IyyerMBD15,
  title        = {Deep Unordered Composition Rivals Syntactic Methods for Text Classification},
  author       = {Mohit Iyyer and Varun Manjunatha and Jordan L. Boyd{-}Graber and Hal Daum{\'{e}} III},
  year         = 2015,
  booktitle    = {{ACL}},
  no_publisher    = {ACL},
 no_pages =  {1681--1691},
  doi          = {10.3115/v1/p15-1162},
  no_url       = {https://doi.org/10.3115/v1/p15-1162}
}

@article{chae_davidson_2023,
  title={Large language models for text classification: From zero-shot learning to fine-tuning},
  author={Chae, Youngjin and Davidson, Thomas},
  journal={Open Science Foundation},
  volume={10},
  year={2023}
}

@inproceedings{DBLP:conf/nips/BrownMRSKDNSSAA20,
  title        = {Language Models are Few-Shot Learners},
  author       = {Tom B. Brown and Benjamin Mann and Nick Ryder and Melanie Subbiah and Jared Kaplan and Prafulla Dhariwal and Arvind Neelakantan and Pranav Shyam and Girish Sastry and Amanda Askell and others},
  year         = 2020,
  booktitle    = {{NeurIPS}},
  no_url          = {https://proceedings.neurips.cc/paper/2020/hash/1457c0d6bfcb4967418bfb8ac142f64a-Abstract.html},
  no_editor    = {Hugo Larochelle and Marc'Aurelio Ranzato and Raia Hadsell and Maria{-}Florina Balcan and Hsuan{-}Tien Lin},
  bibsource    = {dblp computer science bibliography, https://dblp.org}
}

@inproceedings{DBLP:conf/aaai/LaiXLZ15,
  title        = {Recurrent Convolutional Neural Networks for Text Classification},
  author       = {Siwei Lai and Liheng Xu and Kang Liu and Jun Zhao},
  year         = 2015,
  booktitle    = {{AAAI}},
  no_publisher    = {{AAAI} Press},
 no_pages =  {2267--2273},no_url          = {http://www.aaai.org/ocs/index.php/AAAI/AAAI15/paper/view/9745},
  no_editor    = {Blai Bonet and Sven Koenig},
  bibsource    = {dblp computer science bibliography, https://dblp.org}
}

@inproceedings{DBLP:conf/coling/ZhouQZXBX16,
  title        = {Text Classification Improved by Integrating Bidirectional {LSTM} with Two-dimensional Max Pooling},
  author       = {Peng Zhou and Zhenyu Qi and Suncong Zheng and Jiaming Xu and Hongyun Bao and Bo Xu},
  year         = 2016,
  booktitle    = {{COLING}},
  no_publisher    = {{ACL}},
 no_pages =  {3485--3495},no_url          = {https://aclanthology.org/C16-1329/},
  no_editor    = {Nicoletta Calzolari and Yuji Matsumoto and Rashmi Prasad},
  bibsource    = {dblp computer science bibliography, https://dblp.org}
}

@article{DBLP:journals/corr/abs-1811-05868,
  title        = {Pitfalls of Graph Neural Network Evaluation},
  author       = {Oleksandr Shchur and Maximilian Mumme and Aleksandar Bojchevski and Stephan G{\"{u}}nnemann},
  year         = 2018,
  journal      = {arXiv:1811.05868},
}

@inproceedings{DBLP:conf/recsys/ShehzadJ23,
  author       = {Faisal Shehzad and
                  Dietmar Jannach},
  no_editor       = {Jie Zhang and
                  Li Chen and
                  Shlomo Berkovsky and
                  Min Zhang and
                  Tommaso Di Noia and
                  Justin Basilico and
                  Luiz Pizzato and
                  Yang Song},
  title        = {Everyone's a Winner! On Hyperparameter Tuning of Recommendation Models},
  booktitle    = {{RecSys}},
 no_pages =  {652--657},
  publisher    = {{ACM}},
  year         = {2023},no_url          = {https://doi.org/10.1145/3604915.3609488},
  doi          = {10.1145/3604915.3609488},
  bibsource    = {dblp computer science bibliography, https://dblp.org}
}

@inproceedings{DBLP:conf/recsys/DacremaCJ19,
  title        = {Are we really making much progress? {A} worrying analysis of recent neural recommendation approaches},
  author       = {Maurizio Ferrari Dacrema and Paolo Cremonesi and Dietmar Jannach},
  year         = 2019,
  booktitle    = {{RecSys}},
  publisher    = {{ACM}},
 no_pages =  {101--109},
  doi          = {10.1145/3298689.3347058},
  no_editor    = {Toine Bogers and Alan Said and Peter Brusilovsky and Domonkos Tikk},
  no_url       = {https://doi.org/10.1145/3298689.3347058},
  bibsource    = {dblp computer science bibliography, https://dblp.org}
}

@article{DBLP:journals/air/Kadhim19,
  title        = {Survey on supervised machine learning techniques for automatic text classification},
  author       = {Ammar Ismael Kadhim},
  year         = 2019,
  journal      = {Artif. Intell. Rev.},
  volume       = 52,
  number       = 1,
 no_pages =  {273--292}
}

@article{DBLP:journals/wias/ZhouGLVTBBK20,
  title        = {A survey on text classification and its applications},
  author       = {Xujuan Zhou and Raj Gururajan and Yuefeng Li and Revathi Venkataraman and Xiaohui Tao and Ghazal Bargshady and Prabal Datta Barua and Srinivas Kondalsamy{-}Chennakesavan},
  year         = 2020,
  journal      = {Web Intell.},
  volume       = 18,
  number       = 3,
 no_pages =  {205--216}
}

@article{DBLP:journals/information/KowsariMHMBB19,
  title        = {Text Classification Algorithms: {A} Survey},
  author       = {Kamran Kowsari and Kiana Jafari Meimandi and Mojtaba Heidarysafa and Sanjana Mendu and Laura E. Barnes and Donald E. Brown},
  year         = 2019,
  journal      = {Inf.},
  volume       = 10,
  number       = 4,
 no_pages =  150
}

@incollection{Rose2010,
  title        = {Automatic Keyword Extraction from Individual Documents},
  author       = {Rose, Stuart and Engel, Dave and Cramer, Nick and Cowley, Wendy},
  year         = 2010,
  booktitle    = {Text Mining. Applications and Theory},
  publisher    = {Wiley},
 no_pages =  {1--20},
  isbn         = 9780470689646,
  no_doi       = {10.1002/9780470689646.ch1},
  no_editor    = {Berry, Michael W. and Kogan, Jacob},
  no_url       = {http://dx.doi.org/10.1002/9780470689646.ch1}
}

@inproceedings{DBLP:conf/adl/OhsawaBY98,
  title        = {KeyGraph: Automatic Indexing by Co-Occurrence Graph Based on Building Construction Metaphor},
  author       = {Yukio Ohsawa and Nels E. Benson and Masahiko Yachida},
  year         = 1998,
  booktitle    = {IEEE Forum on Res. and Techn. Adv. Digit. Libr.},
  no_publisher    = {{IEEE}},
 no_pages =  {12--18}
}

@inproceedings{DBLP:conf/www/XiaoLS19,
  title        = {Efficient Path Prediction for Semi-Supervised and Weakly Supervised Hierarchical Text Classification},
  author       = {Huiru Xiao and Xin Liu and Yangqiu Song},
  year         = 2019,
  booktitle    = {{WWW}},
  publisher    = {{ACM}},
 no_pages =  {3370--3376}
}

@inproceedings{DBLP:conf/emnlp/DingWLLL20,
  title        = {Be More with Less: Hypergraph Attention Networks for Inductive Text Classification},
  author       = {Kaize Ding and Jianling Wang and Jundong Li and Dingcheng Li and Huan Liu},
  year         = 2020,
  booktitle    = {{EMNLP}},
  no_publisher    = {ACL},
 no_pages =  {4927--4936},
  doi          = {10.18653/v1/2020.emnlp-main.399},
  no_editor    = {Bonnie Webber and Trevor Cohn and Yulan He and Yang Liu},
  no_url       = {https://doi.org/10.18653/v1/2020.emnlp-main.399},
  bibsource    = {dblp computer science bibliography, https://dblp.org}
}

@inproceedings{DBLP:conf/acl/HenaoLCSSWWMZ18,
  title        = {Baseline Needs More Love: On Simple Word-Embedding-Based Models and Associated Pooling Mechanisms},
  author       = {Dinghan Shen and Guoyin Wang and Wenlin Wang and Martin Renqiang Min and Qinliang Su and Yizhe Zhang and Chunyuan Li and Ricardo Henao and Lawrence Carin},
  year         = 2018,
  booktitle    = {{ACL}},
  no_publisher    = {ACL},
 no_pages =  {440--450},
  doi          = {10.18653/v1/P18-1041},
  no_editor    = {Iryna Gurevych and Yusuke Miyao},
  no_url       = {https://aclanthology.org/P18-1041/},
  bibsource    = {dblp computer science bibliography, https://dblp.org}
}

@inproceedings{DBLP:conf/eacl/GraveMJB17,
  title        = {Bag of Tricks for Efficient Text Classification},
  author       = {Armand Joulin and Edouard Grave and Piotr Bojanowski and Tom{\'{a}}s Mikolov},
  year         = 2017,
  booktitle    = {{EACL}},
  no_publisher    = {ACL},
 no_pages =  {427--431},
  doi          = {10.18653/v1/e17-2068},
  no_editor    = {Mirella Lapata and Phil Blunsom and Alexander Koller},
  no_url       = {https://doi.org/10.18653/v1/e17-2068},
  bibsource    = {dblp computer science bibliography, https://dblp.org}
}

@article{DBLP:journals/tacl/BojanowskiGJM17,
  title        = {Enriching Word Vectors with Subword Information},
  author       = {Piotr Bojanowski and Edouard Grave and Armand Joulin and Tom{\'{a}}s Mikolov},
  year         = 2017,
  journal      = {Trans. Assoc. Comput. Linguistics},
  volume       = 5,
 no_pages =  {135--146}
}

@inproceedings{DBLP:conf/icml/WuSZFYW19,
  title        = {Simplifying Graph Convolutional Networks},
  author       = {Felix Wu and Amauri H. Souza Jr. and Tianyi Zhang and Christopher Fifty and Tao Yu and Kilian Q. Weinberger},
  year         = 2019,
  booktitle    = {{ICML}},
  no_publisher    = {{PMLR}},
  no_series       = {PLMR},
  no_volume       = 97,
 no_pages =  {6861--6871},no_url          = {http://proceedings.mlr.press/v97/wu19e.html},
  no_editor    = {Kamalika Chaudhuri and Ruslan Salakhutdinov},
  bibsource    = {dblp computer science bibliography, https://dblp.org}
}

@inproceedings{DBLP:conf/aaai/LiuYZWL20,
  title        = {Tensor Graph Convolutional Networks for Text Classification},
  author       = {Xien Liu and Xinxin You and Xiao Zhang and Ji Wu and Ping Lv},
  year         = 2020,
  booktitle    = {{AAAI}/{IAAI}/{EAAI}},
  no_publisher    = {{AAAI} Press},
 no_pages =  {8409--8416},no_url          = {https://ojs.aaai.org/index.php/AAAI/article/view/6359},
  bibsource    = {dblp computer science bibliography, https://dblp.org}
}

@inproceedings{DBLP:conf/iclr/WangSMHLB19,
  title        = {{GLUE:} {A} Multi-Task Benchmark and Analysis Platform for Natural Language Understanding},
  author       = {Alex Wang and Amanpreet Singh and Julian Michael and Felix Hill and Omer Levy and Samuel R. Bowman},
  year         = 2019,
  booktitle    = {{ICLR}},
  publisher    = {OpenReview.net},no_url          = {https://openreview.net/forum?id=rJ4km2R5t7},
  bibsource    = {dblp computer science bibliography, https://dblp.org}
}

@article{distilbert,
  title        = {{DistilBERT}, a distilled version of {BERT:} smaller, faster, cheaper and lighter},
  author       = {Victor Sanh and Lysandre Debut and Julien Chaumond and Thomas Wolf},
  year         = 2019,
  journal      = {arXiv:1910.01108},
}

@inproceedings{DBLP:conf/naacl/DevlinCLT19,
  title        = {{BERT:} Pre-training of Deep Bidirectional Transformers for Language Understanding},
  author       = {Jacob Devlin and Ming{-}Wei Chang and Kenton Lee and Kristina Toutanova},
  year         = 2019,
  booktitle    = {{NAACL-HLT}},
  no_publisher    = {ACL},
 no_pages =  {4171--4186},
  doi          = {10.18653/v1/n19-1423},
  no_editor    = {Jill Burstein and Christy Doran and Thamar Solorio},
  no_url       = {https://doi.org/10.18653/v1/n19-1423},
  bibsource    = {dblp computer science bibliography, https://dblp.org}
}

@inproceedings{DBLP:conf/iclr/KipfW17,
  title        = {Semi-Supervised Classification with Graph Convolutional Networks},
  author       = {Thomas N. Kipf and Max Welling},
  year         = 2017,
  booktitle    = {{ICLR}},
  publisher    = {OpenReview.net},no_url          = {https://openreview.net/forum?id=SJU4ayYgl},
  bibsource    = {dblp computer science bibliography, https://dblp.org}
}

@inproceedings{velickovic2018graph,
  title        = {Graph Attention Networks},
  author       = {Petar Velickovic and Guillem Cucurull and Arantxa Casanova and Adriana Romero and Pietro Li{\`{o}} and Yoshua Bengio},
  year         = 2018,
  booktitle    = {{ICLR}},
  no_publisher    = {OpenReview.net},no_url          = {https://openreview.net/forum?id=rJXMpikCZ},
  bibsource    = {dblp computer science bibliography, https://dblp.org}
}

@article{DBLP:journals/corr/abs-2011-11197,
  author       = {Weiwei Liu and
                  Haobo Wang and
                  Xiaobo Shen and
                  Ivor W. Tsang},
  title        = {The Emerging Trends of Multi-Label Learning},
  journal      = {{IEEE} Trans. Pattern Anal. Mach. Intell.},
  no_volume       = {44},
  no_number       = {11},
  no_pages        = {7955--7974},
  year         = {2022}
}

@inproceedings{DBLP:conf/nips/MikolovSCCD13,
  title        = {Distributed Representations of Words and Phrases and their Compositionality},
  author       = {Mikolov, Tomas and Sutskever, Ilya and Chen, Kai and Corrado, Gregory S. and Dean, Jeffrey},
  year         = 2013,
  booktitle    = {{NeurIPS}},
 no_pages =  {3111--3119},
  bibsource    = {dblp computer science bibliography, https://dblp.org},
  x-color      = {#009966}
}

@inproceedings{DBLP:conf/kcap/GalkeMSBS17,
  title        = {Using Titles vs. Full-text as Source for Automated Semantic Document Annotation},
  author       = {Lukas Galke and Florian Mai and Alan Schelten and Dennis Brunsch and Ansgar Scherp},
  year         = 2017,
  booktitle    = {Knowledge Capture Conf., {K-CAP} 2017},
  publisher    = {{ACM}},
 no_pages =  {20:1--20:4},
  doi          = {10.1145/3148011.3148039},
  no_editor    = {{\'{O}}scar Corcho and Krzysztof Janowicz and Giuseppe Rizzo and Ilaria Tiddi and Daniel Garijo},
  no_url       = {https://doi.org/10.1145/3148011.3148039},
  bibsource    = {dblp computer science bibliography, https://dblp.org}
}

@inproceedings{DBLP:journals/corr/LiTBZ15,
  title        = {Gated Graph Sequence Neural Networks},
  author       = {Yujia Li and Daniel Tarlow and Marc Brockschmidt and Richard S. Zemel},
  year         = 2016,
  booktitle    = {{ICLR}},no_url          = {http://arxiv.org/abs/1511.05493},
  no_editor    = {Yoshua Bengio and Yann LeCun}
}

@inproceedings{DBLP:journals/corr/KingmaB14,
  title        = {Adam: {A} Method for Stochastic Optimization},
  author       = {Diederik P. Kingma and Jimmy Ba},
  year         = 2015,
  booktitle    = {{ICLR}},no_url          = {http://arxiv.org/abs/1412.6980},
  no_editor    = {Yoshua Bengio and Yann LeCun},
  bibsource    = {dblp computer science bibliography, https://dblp.org}
}

@inproceedings{DBLP:conf/aaai/YaoM019,
  title        = {Graph Convolutional Networks for Text Classification},
  author       = {Liang Yao and Chengsheng Mao and Yuan Luo},
  year         = 2019,
  booktitle    = {{AAAI}},
  no_publisher    = {{AAAI} Press},
 no_pages =  {7370--7377},
  doi          = {10.1609/aaai.v33i01.33017370},
  no_url       = {https://doi.org/10.1609/aaai.v33i01.33017370},
  bibsource    = {dblp computer science bibliography, https://dblp.org}
}

@inproceedings{DBLP:conf/wsdm/RageshSIBL21,
  title        = {{HeteGCN}: Heterogeneous Graph Convolutional Networks for Text Classification},
  author       = {Rahul Ragesh and Sundararajan Sellamanickam and Arun Iyer and Ramakrishna Bairi and Vijay Lingam},
  year         = 2021,
  booktitle    = {{WSDM}},
  publisher    = {{ACM}},
 no_pages =  {860--868},
  doi          = {10.1145/3437963.3441746},
  no_editor    = {Liane Lewin{-}Eytan and David Carmel and Elad Yom{-}Tov and Eugene Agichtein and Evgeniy Gabrilovich},
  no_url       = {https://doi.org/10.1145/3437963.3441746},
  bibsource    = {dblp computer science bibliography, https://dblp.org}
}

@inproceedings{pang-lee-2005-seeing,
  title        = {Seeing Stars: Exploiting Class Relationships for Sentiment Categorization with Respect to Rating Scales},
  author       = {Pang, Bo  and Lee, Lillian},
  year         = 2005,
  no_month        = jun,
  booktitle    = {{ACL}},
  no_publisher    = {ACL},
  no_address      = {Ann Arbor, Michigan},
 no_pages =  {115--124},
  doi          = {10.3115/1219840.1219855},
  no_url       = {https://aclanthology.org/P05-1015}
}

@inproceedings{DBLP:conf/kdd/TangQM15,
  title        = {{PTE:} Predictive Text Embedding through Large-scale Heterogeneous Text Networks},
  author       = {Jian Tang and Meng Qu and Qiaozhu Mei},
  year         = 2015,
  booktitle    = {{SIGKDD}},
  publisher    = {{ACM}},
 no_pages =  {1165--1174},
  doi          = {10.1145/2783258.2783307},
  no_editor    = {Longbing Cao and Chengqi Zhang and Thorsten Joachims and Geoffrey I. Webb and Dragos D. Margineantu and Graham Williams},
  no_url       = {https://doi.org/10.1145/2783258.2783307},
  bibsource    = {dblp computer science bibliography, https://dblp.org}
}

@inproceedings{DBLP:conf/esann/QaraeiKB20,
  title        = {Why state-of-the-art deep learning barely works as good as a linear classifier in extreme multi-label text classification},
  author       = {Mohammadreza Qaraei and Sujay Khandagale and Rohit Babbar},
  year         = 2020,
  booktitle    = {ESANN},
 no_pages =  {223--228},no_url          = {https://www.esann.org/sites/default/files/proceedings/2020/ES2020-207.pdf},
  bibsource    = {dblp computer science bibliography, https://dblp.org}
}

@inproceedings{DBLP:conf/emnlp/ZhangDXLZ21,
  title        = {Weakly-supervised Text Classification Based on Keyword Graph},
  author       = {Lu Zhang and Jiandong Ding and Yi Xu and Yingyao Liu and Shuigeng Zhou},
  year         = 2021,
  booktitle    = {{EMNLP}},
  no_publisher    = {ACL},
 no_pages =  {2803--2813}
}

@inproceedings{DBLP:conf/emnlp/LiuGG0F21,
  title        = {Deep Attention Diffusion Graph Neural Networks for Text Classification},
  author       = {Yonghao Liu and Renchu Guan and Fausto Giunchiglia and Yanchun Liang and Xiaoyue Feng},
  year         = 2021,
  booktitle    = {{EMNLP}},
  no_publisher    = {ACL},
 no_pages =  {8142--8152},
  doi          = {10.18653/v1/2021.emnlp-main.642},
  no_editor    = {Marie{-}Francine Moens and Xuanjing Huang and Lucia Specia and Scott Wen{-}tau Yih},
  no_url       = {https://doi.org/10.18653/v1/2021.emnlp-main.642},
  bibsource    = {dblp computer science bibliography, https://dblp.org}
}

@inproceedings{DBLP:conf/icml/FeyLWL21,
  title        = {{GNNAutoScale}: Scalable and Expressive Graph Neural Networks via Historical Embeddings},
  author       = {Matthias Fey and Jan Eric Lenssen and Frank Weichert and Jure Leskovec},
  year         = 2021,
  booktitle    = {{ICML}},
  no_publisher    = {{PMLR}},
  no_series       = {PLMR},
  no_volume       = 139,
 no_pages =  {3294--3304},no_url          = {http://proceedings.mlr.press/v139/fey21a.html},
  no_editor    = {Marina Meila and Tong Zhang},
  bibsource    = {dblp computer science bibliography, https://dblp.org}
}

@article{gonzales-2020-comparing-bert,
  title        = {Comparing {BERT} against traditional machine learning text classification},
  author       = {Santiago Gonz{\'{a}}lez-Carvajal and Eduardo C. Garrido-Merch{\'{a}}n},
  year         = 2020,
  journal      = {arXiv:2005.13012},
}

@inproceedings{DBLP:conf/nips/VaswaniSPUJGKP17,
  title        = {Attention is All you Need},
  author       = {Ashish Vaswani and Noam Shazeer and Niki Parmar and Jakob Uszkoreit and Llion Jones and Aidan N. Gomez and Lukasz Kaiser and Illia Polosukhin},
  year         = 2017,
  booktitle    = {{NeurIPS}},
 no_pages =  {5998--6008},no_url          = {https://proceedings.neurips.cc/paper/2017/hash/3f5ee243547dee91fbd053c1c4a845aa-Abstract.html},
  no_editor    = {Isabelle Guyon and Ulrike von Luxburg and Samy Bengio and Hanna M. Wallach and Rob Fergus and S. V. N. Vishwanathan and Roman Garnett}
}

@inproceedings{DBLP:conf/sigir/ZhangWYWZ16,
  title        = {Bayesian Performance Comparison of Text Classifiers},
  author       = {Dell Zhang and Jun Wang and Emine Yilmaz and Xiaoling Wang and Yuxin Zhou},
  year         = 2016,
  booktitle    = {{SIGIR}},
  publisher    = {{ACM}},
 no_pages =  {15--24},
  doi          = {10.1145/2911451.2911547},
  no_editor    = {Raffaele Perego and Fabrizio Sebastiani and Javed A. Aslam and Ian Ruthven and Justin Zobel},
  no_url       = {https://doi.org/10.1145/2911451.2911547},
  bibsource    = {dblp computer science bibliography, https://dblp.org}
}

@inproceedings{demszky2020goemotions,
  title        = {Go{E}motions: {A} Dataset of Fine-Grained Emotions},
  author       = {Dorottya Demszky and Dana Movshovitz{-}Attias and Jeongwoo Ko and Alan S. Cowen and Gaurav Nemade and Sujith Ravi},
  year         = 2020,
  booktitle    = {{ACL}},
  no_publisher    = {ACL},
 no_pages =  {4040--4054},
  doi          = {10.18653/v1/2020.acl-main.372},
  no_editor    = {Dan Jurafsky and Joyce Chai and Natalie Schluter and Joel R. Tetreault},
  no_url       = {https://doi.org/10.18653/v1/2020.acl-main.372}
}

@inproceedings{shen-etal-2021-taxoclass,
  title        = {{T}axo{C}lass: Hierarchical Multi-Label Text Classification Using Only Class Names},
  author       = {Shen, Jiaming  and Qiu, Wenda  and Meng, Yu  and Shang, Jingbo  and Ren, Xiang  and Han, Jiawei},
  year         = 2021,
  no_month        = jun,
  booktitle    = {{NAACL}},
  no_publisher    = {ACL},
  no_address      = {Online},
 no_pages =  {4239--4249},
  doi          = {10.18653/v1/2021.naacl-main.335},
  no_url       = {https://aclanthology.org/2021.naacl-main.335}
}

@inproceedings{DBLP:conf/jcdl/MaiGS18,
  title        = {Using Deep Learning for Title-Based Semantic Subject Indexing to Reach Competitive Performance to Full-Text},
  author       = {Florian Mai and Lukas Galke and Ansgar Scherp},
  year         = 2018,
  booktitle    = {JCDL},
  publisher    = {{ACM}},
 no_pages =  {169--178},
  doi          = {10.1145/3197026.3197039},
  no_editor    = {Jiangping Chen and Marcos Andr{\'{e}} Gon{\c{c}}alves and Jeff M. Allen and Edward A. Fox and Min{-}Yen Kan and Vivien Petras},
  no_url       = {https://doi.org/10.1145/3197026.3197039},
  bibsource    = {dblp computer science bibliography, https://dblp.org}
}

@inproceedings{hiddenfactors,
  title        = {Hidden Factors and Hidden Topics: Understanding Rating Dimensions with Review Text},
  author       = {McAuley, Julian and Leskovec, Jure},
  year         = 2013,
  booktitle    = {RecSys},
  no_location     = {Hong Kong, China},
  no_publisher    = {ACM},
  no_address      = {New York, NY, USA},
  no_series       = {RecSys '13},
 no_pages =  {165–172},
  doi          = {10.1145/2507157.2507163},
  isbn         = 9781450324090,
  no_url       = {https://doi.org/10.1145/2507157.2507163},
  numpages     = 8
}

@article{Lehmann2015DBpediaA,
  title        = {{DBpedia} - A large-scale, multilingual knowledge base extracted from {W}ikipedia},
  author       = {Jens Lehmann and Robert Isele and Max Jakob and Anja Jentzsch and Dimitris Kontokostas and Pablo N. Mendes and Sebastian Hellmann and Mohamed Morsey and Patrick van Kleef and S{\"o}ren Auer and Christian Bizer},
  year         = 2015,
  journal      = {Semantic Web},
  volume       = 6,
 no_pages =  {167--195}
}

@misc{reuters,
  title        = {{UCI} Machine Learning Repository},
  author       = {Bache, K. and Lichman, M.},
  year         = 2013,
url          = {http://archive.ics.uci.edu/ml},
  note         = {{U}niversity of California, Irvine, School of Information and Computer Sciences}
}

@inproceedings{hiagm,
  title        = {Hierarchy-Aware Global Model for Hierarchical Text Classification},
  author       = {Zhou, Jie  and Ma, Chunping  and Long, Dingkun  and Xu, Guangwei  and Ding, Ning  and Zhang, Haoyu  and Xie, Pengjun  and Liu, Gongshen},
  year         = 2020,
  no_month        = jul,
  booktitle    = {{ACL}},
  no_publisher    = {ACL},
  no_address      = {Online},
 no_pages =  {1106--1117},
  doi          = {10.18653/v1/2020.acl-main.104},
  no_url       = {https://aclanthology.org/2020.acl-main.104}
}

@misc{sandhaus,
  title        = {{The New York Times Annotated Corpus}},
  author       = {Evan Sandhaus},
  year         = 2008,
  note         = {Linguistic Data Consortium, Philadelphia, 6(12):e26752.}
}

@article{rcv1-v2,
  title        = {{RCV1}: A New Benchmark Collection for Text Categorization Research},
  author       = {Lewis, David D. and Yang, Yiming and Rose, Tony G. and Li, Fan},
  year         = 2004,
  no_month        = {dec},
  journal      = {JMLR},
  no_publisher    = {JMLR.org},
  volume       = 5,
 no_pages =  {361–397},
  issn         = {1532-4435},
  issue_date   = {12/1/2004},
  numpages     = 37
}

@article{zhao2021sequential,
  title        = {A Sequential Graph Neural Network for Short Text Classification},
  author       = {Zhao, Ke and Huang, Lan and Song, Rui and Shen, Qiang and Xu, Hao},
  year         = 2021,
  journal      = {Algorithms},
  publisher    = {MDPI},
  volume       = 14,
  no_number       = 12,
 no_pages =  352
}

@inproceedings{DBLP:conf/nlpcc/WangWYZSJWZ22,
  title        = {{KGAT:} {A}n Enhanced Graph-Based Model for Text Classification},
  author       = {Xin Wang and Chao Wang and Haiyang Yang and Xingpeng Zhang and Qi Shen and Kan Ji and Yuhong Wu and Huayi Zhan},
  year         = 2022,
  booktitle    = {Nat. Lang. Proc. Chinese Comput.},
  publisher    = {Springer},
  no_series       = {Lecture Notes in Computer Science},
  no_volume       = 13551,
 no_pages =  {656--668},
  doi          = {10.1007/978-3-031-17120-8\_51},
  no_editor    = {Wei Lu and Shujian Huang and Yu Hong and Xiabing Zhou},
  no_url       = {https://doi.org/10.1007/978-3-031-17120-8\_51},
  bibsource    = {dblp computer science bibliography, https://dblp.org}
}

@article{zhang,
  title        = {A Review on Multi-Label Learning Algorithms},
  author       = {Zhang, Min-Ling and Zhou, Zhi-Hua},
  year         = 2014,
  journal      = {IEEE Trans. Knowl. Data Eng.},
  volume       = 26,
  number       = 8,
 no_pages =  {1819--1837},
  doi          = {10.1109/TKDE.2013.39}
}

@article{Tsoumakas,
  title        = {Multi-Label Classification: An Overview},
  author       = {Tsoumakas, Grigorios and Katakis, Ioannis},
  year         = 2009,
  no_month        = {09},
  journal      = {Int. Journal of Data Warehousing and Mining},
  volume       = 3,
 no_pages =  {1--13},
  doi          = {10.4018/jdwm.2007070101}
}

@inproceedings{dery2021should,
  author       = {Lucio M. Dery and
                  Paul Michel and
                  Ameet Talwalkar and
                  Graham Neubig},
  title        = {Should We Be Pre-training? {A}n Argument for End-task Aware Training
                  as an Alternative},
  booktitle    = {{ICLR}},
  no_publisher    = {OpenReview.net},
  year         = {2022}
}

@article{JAMSHIDI2024102306,
title = {Effective text classification using {BERT}, {MTM} {LSTM}, and {DT}},
journal = {Data and Knowl. Eng.},
volume = {151},no_pages =  {102306},
year = {2024},
author = {Saman Jamshidi and Mahin Mohammadi and Saeed Bagheri and Hamid Esmaeili Najafabadi and Alireza Rezvanian and Mehdi Gheisari and Mustafa Ghaderzadeh and Amir Shahab Shahabi and Zongda Wu},
}

@inproceedings{DBLP:conf/aaai/LiuZ0ZWMCYZ24,
  author       = {Han Liu and
                  Siyang Zhao and
                  Xiaotong Zhang and
                  Feng Zhang and
                  Wei Wang and
                  Fenglong Ma and
                  Hongyang Chen and
                  Hong Yu and
                  Xianchao Zhang},
  no_editor       = {Michael J. Wooldridge and
                  Jennifer G. Dy and
                  Sriraam Natarajan},
  title        = {Liberating Seen Classes: Boosting Few-Shot and Zero-Shot Text Classification
                  via Anchor Generation and Classification Reframing},
  booktitle    = {{AAAI}},
 no_pages =  {18644--18652},
  no_publisher    = {{AAAI} Press},
  year         = {2024},no_url          = {https://doi.org/10.1609/aaai.v38i17.29827},
  doi          = {10.1609/AAAI.V38I17.29827},
  bibsource    = {dblp computer science bibliography, https://dblp.org}
}

@article{dong2021explainable,
  title        = {Explainable automated coding of clinical notes using hierarchical label-wise attention networks and label embedding initialisation},
  author       = {Hang Dong and V{\'{\i}}ctor Su{\'{a}}rez{-}Paniagua and William Whiteley and Honghan Wu},
  year         = 2021,
  journal      = {J. Biomed. Informatics},
  volume       = 116,
 no_pages =  103728,
  doi          = {10.1016/j.jbi.2021.103728},
  no_url       = {https://doi.org/10.1016/j.jbi.2021.103728},
  bibsource    = {dblp computer science bibliography, https://dblp.org}
}

@article{DBLP:journals/corr/abs-2009-10938,
  author       = {Xinyi Zhang and
                  Jiahao Xu and
                  Charlie Soh and
                  Lihui Chen},
  title        = {{LA-HCN:} Label-based Attention for Hierarchical Multi-label Text
                  Classification Neural Network},
  journal      = {Expert Syst. Appl.},
  volume       = {187},
  no_pages        = {115922},
  year         = {2022}
}

@inproceedings{liu2021improving,
  author       = {Hui Liu and
                  Danqing Zhang and
                  Bing Yin and
                  Xiaodan Zhu},
  title        = {Improving Pretrained Models for Zero-shot Multi-label Text Classification
                  through Reinforced Label Hierarchy Reasoning},
  booktitle    = {{NAACL-HLT}},
  no_pages        = {1051--1062},
  no_publisher    = {Association for Computational Linguistics},
  year         = {2021}
}

@inproceedings{DBLP:journals/corr/abs-1812-11270,
  title        = {Weakly-supervised Text Classification Based on Keyword Graph},
  author       = {Lu Zhang and Jiandong Ding and Yi Xu and Yingyao Liu and Shuigeng Zhou},
  year         = 2021,
  booktitle    = {{EMNLP}},
  no_publisher    = {ACL},
 no_pages =  {2803--2813},
  doi          = {10.18653/v1/2021.emnlp-main.222},
  no_editor    = {Marie{-}Francine Moens and Xuanjing Huang and Lucia Specia and Scott Wen{-}tau Yih},
  no_url       = {https://doi.org/10.18653/v1/2021.emnlp-main.222},
  bibsource    = {dblp computer science bibliography, https://dblp.org}
}

@inproceedings{DBLP:journals/corr/abs-1909-00161,
  author       = {Wenpeng Yin and
                  Jamaal Hay and
                  Dan Roth},
  title        = {Benchmarking Zero-shot Text Classification: Datasets, Evaluation and
                  Entailment Approach},
  booktitle    = {{EMNLP/IJCNLP}},
  no_pages        = {3912--3921},
  no_publisher    = {Association for Computational Linguistics},
  year         = {2019}
}

@article{lstm,
  title        = {Long Short-Term Memory},
  author       = {Sepp Hochreiter and J{\"{u}}rgen Schmidhuber},
  year         = 1997,
  journal      = {Neural Comput.},
  volume       = 9,
  number       = 8,
 no_pages =  {1735--1780},
  doi          = {10.1162/neco.1997.9.8.1735},
  no_url       = {https://doi.org/10.1162/neco.1997.9.8.1735},
  bibsource    = {dblp computer science bibliography, https://dblp.org}
}

@article{DBLP:journals/tois/YangHSJLN21,
  title        = {{HGAT:} Heterogeneous Graph Attention Networks for Semi-supervised Short Text Classification},
  author       = {Tianchi Yang and Linmei Hu and Chuan Shi and Houye Ji and Xiaoli Li and Liqiang Nie},
  year         = 2021,
  journal      = {{ACM} Trans. Inf. Syst.},
  volume       = 39,
  number       = 3,
 no_pages =  {32:1--32:29},
  doi          = {10.1145/3450352},
  no_url       = {https://doi.org/10.1145/3450352},
  bibsource    = {dblp computer science bibliography, https://dblp.org}
}

@inproceedings{hbgl,
  title        = {Exploiting Global and Local Hierarchies for Hierarchical Text Classification},
  author       = {Jiang, Ting  and Wang, Deqing  and Sun, Leilei  and Chen, Zhongzhi  and Zhuang, Fuzhen  and Yang, Qinghong},
  year         = 2022,
  no_month        = dec,
  booktitle    = {{EMNLP}},
  no_publisher    = {ACL},
  no_address      = {Abu Dhabi, United Arab Emirates},
 no_pages =  {4030--4039},no_url          = {https://aclanthology.org/2022.emnlp-main.268}
}

@inproceedings{DBLP:conf/acl/WangWH0W22,
  title        = {Incorporating Hierarchy into Text Encoder: a Contrastive Learning Approach for Hierarchical Text Classification},
  author       = {Zihan Wang and Peiyi Wang and Lianzhe Huang and Xin Sun and Houfeng Wang},
  year         = 2022,
  booktitle    = {{ACL}},
  no_publisher    = {ACL},
 no_pages =  {7109--7119},
  doi          = {10.18653/v1/2022.acl-long.491},
  no_editor    = {Smaranda Muresan and Preslav Nakov and Aline Villavicencio},
  no_url       = {https://doi.org/10.18653/v1/2022.acl-long.491},
  bibsource    = {dblp computer science bibliography, https://dblp.org}
}

@inproceedings{DBLP:conf/emnlp/WangWYD21,
  title        = {Hierarchical Heterogeneous Graph Representation Learning for Short Text Classification},
  author       = {Yaqing Wang and Song Wang and Quanming Yao and Dejing Dou},
  year         = 2021,
  booktitle    = {{EMNLP}},
  no_publisher    = {ACL},
 no_pages =  {3091--3101},
  doi          = {10.18653/v1/2021.emnlp-main.247},
  no_editor    = {Marie{-}Francine Moens and Xuanjing Huang and Lucia Specia and Scott Wen{-}tau Yih},
  no_url       = {https://doi.org/10.18653/v1/2021.emnlp-main.247},
  bibsource    = {dblp computer science bibliography, https://dblp.org}
}

@inproceedings{stgcn,
  title        = {Document and Word Representations Generated by Graph Convolutional Network and {BERT} for Short Text Classification},
  author       = {Zhihao Ye and Gongyao Jiang and Ye Liu and Zhiyong Li and Jin Yuan},
  year         = 2020,
  booktitle    = {{ECAI}},
  publisher    = {{IOS} Press},
  no_series       = {Frontiers in Artificial Intelligence and Applications},
  no_volume       = 325,
 no_pages =  {2275--2281},
  doi          = {10.3233/FAIA200355},
  no_editor    = {Giuseppe De Giacomo and Alejandro Catal{\'{a}} and Bistra Dilkina and Michela Milano and Sen{\'{e}}n Barro and Alberto Bugar{\'{\i}}n and J{\'{e}}r{\^{o}}me Lang},
  no_url       = {https://doi.org/10.3233/FAIA200355},
  bibsource    = {dblp computer science bibliography, https://dblp.org}
}

@article{aggnn,
  title        = {Text {Classification} with {Attention} {Gated} {Graph} {Neural} {Network}},
  author       = {Deng, Zhaoyang and Sun, Chenxiang and Zhong, Guoqiang and Mao, Yuxu},
  year         = 2022,
  no_month        = jul,
  journal      = {Cogn. Comput.},
  volume       = 14,
  no_number       = 4,
 no_pages =  {1464--1473},
  doi          = {10.1007/s12559-022-10017-3},
  issn         = {1866-9964},no_urldate      = {2022-09-08},
  no_url       = {https://doi.org/10.1007/s12559-022-10017-3},
  language     = {en}
}

@inproceedings{induct-gcn,
  title        = {{InducT-GCN}: Inductive Graph Convolutional Networks for Text Classification},
  author       = {Kunze Wang and Soyeon Caren Han and Josiah Poon},
  year         = 2022,
  booktitle    = {26th Int. Conf. on Pattern Recognition, {ICPR}},
  publisher    = {{IEEE}},
 no_pages =  {1243--1249},
  doi          = {10.1109/ICPR56361.2022.9956075},no_url          = {https://doi.org/10.1109/ICPR56361.2022.9956075},
  bibsource    = {dblp computer science bibliography, https://dblp.org}
}

@inproceedings{ilgcn,
  title        = {Inductive {Light} {Graph} {Convolution} {Network} for {Text} {Classification} {Based} on {Word}-{Label} {Graph}},
  author       = {Shi, Jinze and Wu, Xiaoming and Liu, Xiangzhi and Lu, Wenpeng and Li, Shu},
  year         = 2022,
  booktitle    = {Intelligent {Information} {Processing} {XI}},
  publisher    = {Springer},
  no_address      = {Cham},
  no_series       = {{IFIP} {Advances} in {Information} and {Communication} {Technology}},
 no_pages =  {42--55},
  no_editor    = {Shi, Zhongzhi and Zucker, Jean-Daniel and An, Bo}
}

@article{tsw-gnn,
  title        = {Integration of global and local information for text classification},
  author       = {Li, Xianghua and Wu, Xinyu and Luo, Zheng and Du, Zhanwei and Wang, Zhen and Gao, Chao},
  year         = 2022,
  no_month        = aug,
  journal      = {Neural Computing and Applications},
  doi          = {10.1007/s00521-022-07727-y},
  issn         = {0941-0643, 1433-3058},no_urldate      = {2022-09-19},
  no_url       = {https://link.springer.com/10.1007/s00521-022-07727-y},
  language     = {en}
}

@article{liu_roberta:_2019,
  title        = {{RoBERTa}: {A} {Robustly} {Optimized} {BERT} {Pretraining} {Approach}},
  shorttitle   = {{RoBERTa}},
  author       = {Liu, Yinhan and Ott, Myle and Goyal, Naman and Du, Jingfei and Joshi, Mandar and Chen, Danqi and Levy, Omer and Lewis, Mike and Zettlemoyer, Luke and Stoyanov, Veselin},
  year         = 2019,
  journal      = {arXiv:1907.11692},
}

@inproceedings{lan_albert_2020,
  author       = {Zhenzhong Lan and
                  Mingda Chen and
                  Sebastian Goodman and
                  Kevin Gimpel and
                  Piyush Sharma and
                  Radu Soricut},
  title        = {{ALBERT:} {A} Lite {BERT} for Self-supervised Learning of Language
                  Representations},
  booktitle    = {{ICLR}},
  publisher    = {OpenReview.net},
  year         = {2020}
}

@inproceedings{sun_ernie_2019,
  author       = {Yu Sun and
                  Shuohuan Wang and
                  Yu{-}Kun Li and
                  Shikun Feng and
                  Hao Tian and
                  Hua Wu and
                  Haifeng Wang},
  title        = {{ERNIE} 2.0: {A} Continual Pre-Training Framework for Language Understanding},
  booktitle    = {{AAAI}},
  no_pages        = {8968--8975},
  publisher    = {{AAAI} Press},
  year         = {2020}
}

@inproceedings{he_deberta_2021,
  author       = {Pengcheng He and
                  Xiaodong Liu and
                  Jianfeng Gao and
                  Weizhu Chen},
  title        = {Deberta: decoding-Enhanced Bert with Disentangled Attention},
  booktitle    = {{ICLR}},
  no_publisher    = {OpenReview.net},
  year         = {2021}
}

@inproceedings{DBLP:conf/icaart/PalSS20,
  title        = {{MAGNET:} Multi-Label Text Classification using Attention-based Graph Neural Network},
  author       = {Ankit Pal and Muru Selvakumar and Malaikannan Sankarasubbu},
  year         = 2020,
  booktitle    = {ICAART},
  publisher    = {{SCITEPRESS}},
 no_pages =  {494--505},
  doi          = {10.5220/0008940304940505},
  no_editor    = {Ana Paula Rocha and Luc Steels and H. Jaap van den Herik},
  no_url       = {https://doi.org/10.5220/0008940304940505},
  bibsource    = {dblp computer science bibliography, https://dblp.org}
}

@book{cover_elements_1991,
  title        = {Elements of {Information} {Theory}},
  author       = {Cover, Thomas M and Thomas, Joy A},
  year         = 1991,
  publisher    = {Wiley},
 no_pages =  413,
  language     = {en}
}

@inproceedings{texting_acl2020,
  title        = {Every Document Owns Its Structure: Inductive Text Classification via Graph Neural Networks},
  author       = {Yufeng Zhang and Xueli Yu and Zeyu Cui and Shu Wu and Zhongzhen Wen and Liang Wang},
  year         = 2020,
  booktitle    = {{ACL}},
  no_publisher    = {ACL},
 no_pages =  {334--339},
  doi          = {10.18653/v1/2020.acl-main.31},
  no_editor    = {Dan Jurafsky and Joyce Chai and Natalie Schluter and Joel R. Tetreault},
  no_url       = {https://doi.org/10.18653/v1/2020.acl-main.31},
  bibsource    = {dblp computer science bibliography, https://dblp.org}
}

@article{bgnn-xml,
  title        = {{BGNN-XML}: Bilateral Graph Neural Networks for Extreme Multi-label Text Classification},
  author       = {Zong, Daoming and Sun, Shiliang},
  year         = 2022,
  journal      = {IEEE Trans. on Knowl. and Data Eng.},
  volume       = {},
  number       = {},
 no_pages =  {1--12},
  doi          = {10.1109/TKDE.2022.3193657}
}

@article{duarte2023review,
  title        = {A review of semi-supervised learning for text classification},
  author       = {Duarte, Jos{\'e} Marcio and Berton, Lilian},
  year         = 2023,
  journal      = {Artif. Intel.l. Rev.},
  publisher    = {Springer},
 no_pages =  {1--69}
}

@inproceedings{chalkidis-etal-2020-empirical,
  title        = {An Empirical Study on Large-Scale Multi-Label Text Classification Including Few and Zero-Shot Labels},
  author       = {Chalkidis, Ilias  and Fergadiotis, Manos  and Kotitsas, Sotiris  and Malakasiotis, Prodromos  and Aletras, Nikolaos  and Androutsopoulos, Ion},
  year         = 2020,
  no_month        = nov,
  booktitle    = {{EMNLP}},
  no_publisher    = {ACL},
  no_address      = {Online},
 no_pages =  {7503--7515},
  doi          = {10.18653/v1/2020.emnlp-main.607},
  no_url       = {https://aclanthology.org/2020.emnlp-main.607}
}

@inproceedings{gururangan-etal-2019-variational,
  title        = {Variational Pretraining for Semi-supervised Text Classification},
  author       = {Gururangan, Suchin  and Dang, Tam  and Card, Dallas  and Smith, Noah A.},
  year         = 2019,
  no_month        = jul,
  booktitle    = {{ACL}},
  no_publisher    = {ACL},
  no_address      = {Florence, Italy},
 no_pages =  {5880--5894},
  doi          = {10.18653/v1/P19-1590},
  no_url       = {https://aclanthology.org/P19-1590}
}

@inproceedings{DBLP:conf/emnlp/HuangGKOO21,
  title        = {Balancing Methods for Multi-label Text Classification with Long-Tailed Class Distribution},
  author       = {Yi Huang and Buse Giledereli and Abdullatif K{\"{o}}ksal and Arzucan {\"{O}}zg{\"{u}}r and Elif Ozkirimli},
  year         = 2021,
  booktitle    = {{EMNLP}},
  no_publisher    = {ACL},
 no_pages =  {8153--8161},
  doi          = {10.18653/v1/2021.emnlp-main.643},
  no_editor    = {Marie{-}Francine Moens and Xuanjing Huang and Lucia Specia and Scott Wen{-}tau Yih},
  no_url       = {https://doi.org/10.18653/v1/2021.emnlp-main.643},
  bibsource    = {dblp computer science bibliography, https://dblp.org}
}

@article{galke2023really,
  title        = {Are We Really Making Much Progress? Bag-of-Words vs. Sequence vs. Graph vs. Hierarchy for Single- and Multi-Label Text Classification},
  author       = {Lukas Galke and Andor Diera and Bao Xin Lin and Bhakti Khera and Tim Meuser and Tushar Singhal and Fabian Karl and Ansgar Scherp},
  year         = 2023,
  journal      = {arXiv:2204.03954},
}

@inproceedings{textssl2022,
  title        = {Sparse Structure Learning via Graph Neural Networks for Inductive Document Classification},
  author       = {Piao, Yinhua and Lee, Sangseon and Lee, Dohoon and Kim, Sun},
  year         = 2022,
  no_month        = jun,
  booktitle      = {AAAI},
  no_volume       = 36,
  no_number       = 10,
 no_pages =  {11165--11173},
  doi          = {10.1609/aaai.v36i10.21366},
  issn         = {2374-3468}
}

@article{gltc2023,
  title        = {Integrating Information by {{Kullback}}\textendash{{Leibler}} Constraint for Text Classification},
  author       = {Yin, Shu and Zhu, Peican and Wu, Xinyu and Huang, Jiajin and Li, Xianghua and Wang, Zhen and Gao, Chao},
  year         = 2023,
  no_month        = may,
  journal      = {Neural Computing and Applications},
  doi          = {10.1007/s00521-023-08602-0},
  issn         = {1433-3058},no_urldate      = {2023-05-11},
  langid       = {english}
}

@inproceedings{carp,
  author       = {Xiaofei Sun and
                  Xiaoya Li and
                  Jiwei Li and
                  Fei Wu and
                  Shangwei Guo and
                  Tianwei Zhang and
                  Guoyin Wang},
  title        = {Text Classification via Large Language Models},
  booktitle    = {{EMNLP} Findings},
  no_pages        = {8990--9005},
  no_publisher    = {Association for Computational Linguistics},
  year         = {2023}
}

@inproceedings{namMaximizingSubsetAccuracy2017,
  title        = {Maximizing {{Subset Accuracy}} with {{Recurrent Neural Networks}} in {{Multi-label Classification}}},
  author       = {Nam, Jinseok and Loza Menc{\'i}a, Eneldo and Kim, Hyunwoo J and F{\"u}rnkranz, Johannes},
  year         = 2017,
  booktitle    = {{NeurIPS}},
  publisher    = {{Curran Associates, Inc.}},
 no_pages =  {5413--5423},
 no_editor       = {Guyon, I. and Luxburg, U. V. and Bengio, S. and Wallach, H. and Fergus, R. and Vishwanathan, S. and Garnett, R.}
}

@article{Xie2024DataLT,
  title={Data labeling through the centralities of co-reference networks improves the classification accuracy of scientific papers},
  author={Zheng Xie and Yiqin Lv and Yiping Song and Qi Wang},
  journal={Journal of Informetrics},
  year={2024},no_url={https://api.semanticscholar.org/CorpusID:267116237}
}

@inproceedings{DBLP:conf/coling/LyKCK24,
  author       = {Khang Ly and
                  Yury Kashnitsky and
                  Savvas Chamezopoulos and
                  Valeria V. Krzhizhanovskaya},
  no_editor       = {Nicoletta Calzolari and
                  Min{-}Yen Kan and
                  V{\'{e}}ronique Hoste and
                  Alessandro Lenci and
                  Sakriani Sakti and
                  Nianwen Xue},
  title        = {Article Classification with Graph Neural Networks and Multigraphs},
  booktitle    = {{LREC/COLING}},
 no_pages =  {1539--1547},
  no_publisher    = {{ELRA} and {ICCL}},
  year         = {2024},no_url          = {https://aclanthology.org/2024.lrec-main.136},
  bibsource    = {dblp computer science bibliography, https://dblp.org}
}

@article{MTAPparlakNovelFeatureClassbased2023,
  title        = {A Novel Feature and Class-Based Globalization Technique for Text Classification},
  author       = {Parlak, Bekir},
  year         = 2023,
  no_month        = apr,
  journal      = {Multimedia Tools and Applications},
  doi          = {10.1007/s11042-023-15459-x},
  issn         = {1380-7501, 1573-7721}
}

@article{radfordLanguageModelsAre,
  title        = {Language Models Are Unsupervised Multitask Learners},
  author       = {Radford, Alec and Wu, Jeffrey and Child, Rewon and Luan, David and Amodei, Dario and Sutskever, Ilya},
  journal={OpenAI blog},
    year={2019},
  volume={1},
  number={8},
}

@article{buguenoConnectingDotsWhat2023,
  title        = {Connecting the {{Dots}}: {{What Graph-Based Text Representations Work Best}} for {{Text Classification}} Using {{Graph Neural Networks}}?},
  shorttitle   = {Connecting the {{Dots}}},
  author       = {Bugue{\~n}o, Margarita and {de Melo}, Gerard},
  year         = 2023,
  journal       = {{arXiv}:2305.14578},
}

@inproceedings{yuan2023revisiting,
  author       = {Lifan Yuan and
                  Yangyi Chen and
                  Ganqu Cui and
                  Hongcheng Gao and
                  Fangyuan Zou and
                  Xingyi Cheng and
                  Heng Ji and
                  Zhiyuan Liu and
                  Maosong Sun},
  title        = {Revisiting Out-of-distribution Robustness in {NLP:} {B}enchmarks, Analysis,
                  and {LLMs} Evaluations},
  booktitle    = {NeurIPS},
  year         = {2023}
}

@article{li2023chatgpt,
  title        = {Are {ChatGPT} and {GPT-4} General-Purpose Solvers for Financial Text Analytics? An Examination on Several Typical Tasks},
  author       = {Li, Xianzhi and Zhu, Xiaodan and Ma, Zhiqiang and Liu, Xiaomo and Shah, Sameena},
  year         = 2023,
  journal      = {{arXiv}:2305.05862}
}

@article{yu2023open,
  title        = {Open, Closed, or Small Language Models for Text Classification?},
  author       = {Yu, Hao and Yang, Zachary and Pelrine, Kellin and Godbout, Jean Francois and Rabbany, Reihaneh},
  year         = 2023,
  journal      = {{arXiv}:2308.10092}
}

@inproceedings{xml2,
  title = {{{CascadeXML}}: {{Rethinking Transformers}} for {{End-to-end Multi-resolution Training}} in {{Extreme Multi-label Classification}}},
  shorttitle = {{{CascadeXML}}},
  booktitle = {{NeurIPS}},
  author = {Kharbanda, Siddhant and Banerjee, Atmadeep and Schultheis, Erik and Babbar, Rohit},
  year = {2022},
  no_month = dec,
  no_volume = {35},
 no_pages =  {2074--2087},no_urldate = {2023-09-20},
  langid = {english}
}

@inproceedings{xml3,
  title = {{{InceptionXML}}: {{A Lightweight Framework}} with {{Synchronized Negative Sampling}} for {{Short Text Extreme Classification}}},
  shorttitle = {{{InceptionXML}}},
  booktitle = {{SIGIR}},
  author = {Kharbanda, Siddhant and Banerjee, Atmadeep and Gupta, Devaansh and Palrecha, Akash and Babbar, Rohit},
  year = {2023},
  no_month = jul,
  no_series = {{{SIGIR}} '23},
 no_pages =  {760--769},
  no_publisher = {ACM},
  no_address = {New York, NY, USA},
  doi = {10.1145/3539618.3591699},no_urldate = {2023-09-20},
  isbn = {978-1-4503-9408-6}
}

@inproceedings{xml0,
  title = {{{AttentionXML}}: {{Label Tree-based Attention-Aware Deep Model}} for {{High-Performance Extreme Multi-Label Text Classification}}},
  shorttitle = {{{AttentionXML}}},
  booktitle = {{NeurIPS}},
  author = {You, Ronghui and Zhang, Zihan and Wang, Ziye and Dai, Suyang and Mamitsuka, Hiroshi and Zhu, Shanfeng},
  year = {2019},
  volume = {32},
  publisher = {Curran Associates, Inc.},no_urldate = {2023-09-20}
}

@inproceedings{xml4,
  title = {Fast {{Multi-Resolution Transformer Fine-tuning}} for {{Extreme Multi-label Text Classification}}},
  booktitle = {{NeurIPS}},
  author = {Zhang, Jiong and Chang, Wei-Cheng and Yu, Hsiang-Fu and Dhillon, Inderjit},
  year = {2021},
  volume = {34},
 no_pages =  {7267--7280},
  publisher = {Curran Associates, Inc.},no_urldate = {2023-09-20}
}

@inproceedings{
fatemi2024talk,
title={Talk like a Graph: Encoding Graphs for Large Language Models},
author={Bahare Fatemi and Jonathan Halcrow and Bryan Perozzi},
booktitle={{ICLR}},
year={2024},no_url={https://openreview.net/forum?id=IuXR1CCrSi}
}

@article{touvronLlamaOpenFoundation2023,
  title = {Llama 2: {{Open Foundation}} and {{Fine-Tuned Chat Models}}},
  shorttitle = {Llama 2},
  author = {Touvron, Hugo and Martin, Louis and Stone, Kevin and Albert, Peter and Almahairi, Amjad and Babaei, Yasmine and Bashlykov, Nikolay and Batra, Soumya and Bhargava, Prajjwal and others},
  year = {2023},
  no_month = jul,
  journal = {{arXiv}:2307.09288},
}

@inproceedings{balloccuLeakCheatRepeat2024,
  title = {Leak, {{Cheat}}, {{Repeat}}: {{Data Contamination}} and {{Evaluation Malpractices}} in {{Closed-Source LLMs}}},
  shorttitle = {Leak, {{Cheat}}, {{Repeat}}},
  booktitle = {{EACL}},
  author = {Balloccu, Simone and Schmidtov{\'a}, Patr{\'i}cia and Lango, Mateusz and Dusek, Ondrej},
  no_editor = {Graham, Yvette and Purver, Matthew},
  year = {2024},
  no_month = mar,
 no_pages =  {67--93},
  no_publisher = {ACL},
  no_no_address = {St. Julian's, Malta},
}

@inproceedings{edwards2024language,
    title = "Language Models for Text Classification: Is In-Context Learning Enough?",
    author = "Edwards, Aleksandra  and
      Camacho-Collados, Jose",
    no_editor = "Calzolari, Nicoletta  and
      Kan, Min-Yen  and
      Hoste, Veronique  and
      Lenci, Alessandro  and
      Sakti, Sakriani  and
      Xue, Nianwen",
    booktitle = {{LREC-COLING}},
    no_month = may,
    year = "2024",
    no_address = "Torino, Italia",
    no_publisher = "ELRA and ICCL",no_url = "https://aclanthology.org/2024.lrec-main.879",
   no_pages =  "10058--10072",
}

@InProceedings{pmlr-v202-hou23b,
  title = 	 {{P}rompt{B}oosting: Black-Box Text Classification with Ten Forward Passes},
  author =       {Hou, Bairu and O'Connor, Joe and Andreas, Jacob and Chang, Shiyu and Zhang, Yang},
  booktitle = 	 {{ICML}},
 no_pages =  	 {13309--13324},
  year = 	 {2023},
  no_editor = 	 {Krause, Andreas and Brunskill, Emma and Cho, Kyunghyun and Engelhardt, Barbara and Sabato, Sivan and Scarlett, Jonathan},
  no_volume = 	 {202},
  no_series = 	 {PLMR},
  no_month = 	 {23--29 Jul},
  no_publisher =    {PMLR},no_url = 	 {https://proceedings.mlr.press/v202/hou23b.html}
}

@article{li2023label,
      title={Label Supervised LLaMA Finetuning}, 
      author={Zongxi Li and Xianming Li and Yuzhang Liu and Haoran Xie and Jing Li and Fu-lee Wang and Qing Li and Xiaoqin Zhong},
      year={2023},
      journal={{arXiv}:2310.01208},
}

@article{yehudai2024fastfit,
      title={When {LLMs} are Unfit Use {FastFit}: Fast and Effective Text Classification with Many Classes}, 
      author={Asaf Yehudai and Elron Bendel},
      year={2024},
      journal={{arXiv}:2404.12365},
}

@article{leech2024questionablepracticesmachinelearning,
      title={Questionable practices in machine learning}, 
      author={Gavin Leech and Juan J. Vazquez and Misha Yagudin and Niclas Kupper and Laurence Aitchison},
      year={2024},
      journal={{arXiv}:2407.12220},
}

@article{bucher2024finetunedsmallllmsstill,
      title={Fine-Tuned 'Small' {LLMs} (Still) Significantly Outperform Zero-Shot Generative AI Models in Text Classification}, 
      author={Martin Juan José Bucher and Marco Martini},
      year={2024},
      journal={{arXiv}:2406.08660},
}

@inproceedings{lepagnol2024smalllanguagemodelsgood,
  author       = {Pierre Lepagnol and
                  Thomas Gerald and
                  Sahar Ghannay and
                  Christophe Servan and
                  Sophie Rosset},
  title        = {Small Language Models Are Good Too: An Empirical Study of Zero-Shot
                  Classification},
  booktitle    = {{LREC/COLING}},
  no_pages        = {14923--14936},
  publisher    = {{ELRA} and {ICCL}},
  year         = {2024}
}

@inproceedings{liu2024llmembedrethinkinglightweightllms,
  author       = {ChunLiu ChunLiu and
                  Hongguang Zhang and
                  Kainan Zhao and
                  Xinghai Ju and
                  Lin Yang},
  title        = {LLMEmbed: Rethinking Lightweight LLM's Genuine Function in Text Classification},
  booktitle    = {{ACL}},
  no_pages        = {7994--8004},
  no_publisher    = {Association for Computational Linguistics},
  year         = {2024}
}

@article{gnns-for-nlp-survey,no_url = {http://dx.doi.org/10.1561/2200000096},
year = {2023},
volume = {16},
journal = {Foundations and Trends in Machine Learning},
title = {Graph neural networks for natural language processing: a survey},
doi = {10.1561/2200000096},
issn = {1935-8237},
number = {2},no_pages =  {119-328},
author = {Lingfei Wu and Yu Chen and Kai Shen and Xiaojie Guo and Hanning Gao and Shucheng Li and Jian Pei and Bo Long}
}

@inproceedings{hill2024,
  author       = {He Zhu and
                  Junran Wu and
                  Ruomei Liu and
                  Yue Hou and
                  Ze Yuan and
                  Shangzhe Li and
                  Yicheng Pan and
                  Ke Xu},
  no_editor       = {Kevin Duh and
                  Helena G{\'{o}}mez{-}Adorno and
                  Steven Bethard},
  title        = {{HILL:} {H}ierarchy-aware Information Lossless Contrastive Learning for Hierarchical Text Classification},
  booktitle    = {NAACL},
 no_pages =  {4731--4745},
  no_publisher    = {ACL},
  year         = {2024},no_url          = {https://doi.org/10.18653/v1/2024.naacl-long.265},
  doi          = {10.18653/V1/2024.NAACL-LONG.265},
  bibsource    = {dblp computer science bibliography, https://dblp.org}
}

@Article{electronics13071199,
AUTHOR = {Zangari, Alessandro and Marcuzzo, Matteo and Rizzo, Matteo and Giudice, Lorenzo and Albarelli, Andrea and Gasparetto, Andrea},
TITLE = {Hierarchical Text Classification and Its Foundations: A Review of Current Research},
JOURNAL = {Electronics},
VOLUME = {13},
YEAR = {2024},
NUMBER = {7},
ARTICLE-NUMBER = {1199},no_url = {https://www.mdpi.com/2079-9292/13/7/1199},
ISSN = {2079-9292},
DOI = {10.3390/electronics13071199}
}

@article{shao2024deepseekmath,
  title={Deep{S}eek{M}ath: {P}ushing the limits of mathematical reasoning in open language models},
  author={Shao, Zhihong and Wang, Peiyi and Zhu, Qihao and Xu, Runxin and Song, Junxiao and Bi, Xiao and Zhang, Haowei and Zhang, Mingchuan and Li, YK and Wu, Y and others},
  journal={arXiv:2402.03300},
  year={2024}
}

@inproceedings{lewis2020bart,
  title={{BART}: Denoising sequence-to-sequence pre-training for natural language generation, translation, and comprehension},
  author={Lewis, Mike and Liu, Yinhan and Goyal, Naman and Ghazvininejad, Marjan and Mohamed, Abdelrahman and Levy, Omer and Stoyanov, Veselin and Zettlemoyer, Luke},
  booktitle={ACL},
  no_pages={7871--7880},
  year={2020}
}

\begin{IEEEbiographynophoto}{Ansgar Scherp} is Full Professor for Data Science and Big Data Analytics with Ulm University, Germany, since 2020. He has long-standing experience in interdisciplinary projects in Germany as well as the USA and UK. His main interests are in graph machine learning and natural language processing, often with applications in information retrieval and extraction.

\end{IEEEbiographynophoto}

\begin{IEEEbiographynophoto}{Andor Diera, Fabian Karl, Bao Xin Lin, Bhakti Khera, Tim Meuser, and Tushar Singhal} contributed as students in computer science and cognitive systems at Ulm University. Bao passed away during the preparation of this manuscript; her colleagues and co-authors dedicate this work to her memory.
\end{IEEEbiographynophoto}

\begin{IEEEbiographynophoto}{Lukas Galke Poech} 
(formerly published as L. Galke) received the Ph.D. degree in computer science from Kiel University, Germany in 2023.
He was a Doctoral Researcher with Kiel University, Germany and with the Leibniz Information Centre for Economics, and from 2022 to 2024 a Postdoctoral Researcher at the Max Planck Institute for Psycholinguistics, Nijmegen. Since 2024 he has been with the University of Southern Denmark where he is currently
an Associate Professor. His research interests include AI safety, interpretability, and representation learning in language models.
For more information, see \url{https://lgalke.github.io}.
\end{IEEEbiographynophoto}

\end{document}